%% file: egpaper_for_review_final.tex
\documentclass[10pt,twocolumn,letterpaper]{article}

\usepackage{iccv}
\usepackage{times}
\usepackage{epsfig}
\usepackage{graphicx}
\usepackage{amsmath}
\usepackage{amssymb}
\usepackage{setspace}

\usepackage[pagebackref=true,breaklinks=true,letterpaper=true,colorlinks,bookmarks=false]{hyperref}

\input{dg_def}

 \iccvfinalcopy 

\def\iccvPaperID{****} 

\begin{document}

\title{An Effective Two-Branch Model-Based Deep Network for Single \\Image Deraining}

\author{Yinglong Wang{$^{\dag}$}, Dong Gong{$^{\ddag}$}, Jie Yang{$^{\ddag}$}, Qinfeng Shi{$^{\ddag}$}, Anton van den Hengel{$^{\ddag}$}, \\ Dehua Xie{$^{\dag}$}, Bing Zeng{$^{\dag}$}\\
$^{\dag}$School of Information and Communication Engineering,\\ University of Electronic Science and Technology of China\\
$^{\ddag}$The University of Adelaide\\}

\maketitle

\begin{abstract}

Removing rain effects from an image is of importance for various applications such as autonomous driving, drone piloting, and photo editing. Conventional methods rely on some heuristics to handcraft various priors to remove or separate the rain effects from an image. Recent deep learning models are proposed to learn end-to-end methods to complete this task.
However, they often fail to obtain satisfactory results in many realistic scenarios, especially when the observed images suffer from heavy rain.
Heavy rain brings not only rain streaks but also haze-like effect caused by the accumulation of tiny raindrops.
Different from the existing deep learning deraining methods that mainly focus on handling the rain streaks, we design a deep neural network by incorporating a physical raining image model. Specifically, in the proposed model, two branches are designed to handle both the rain streaks and haze-like effects. An additional submodule is jointly trained to finally refine the results, which give the model flexibility to control the strength of removing the mist.
Extensive experiments on several datasets show that our method outperforms the state-of-the-art in both objective assessments and visual quality.

\end{abstract}

\section{Introduction}

\begin{figure}[!t]
\begin{center}
\begin{minipage}{0.24\linewidth}
\centering{\includegraphics[width=1\linewidth]{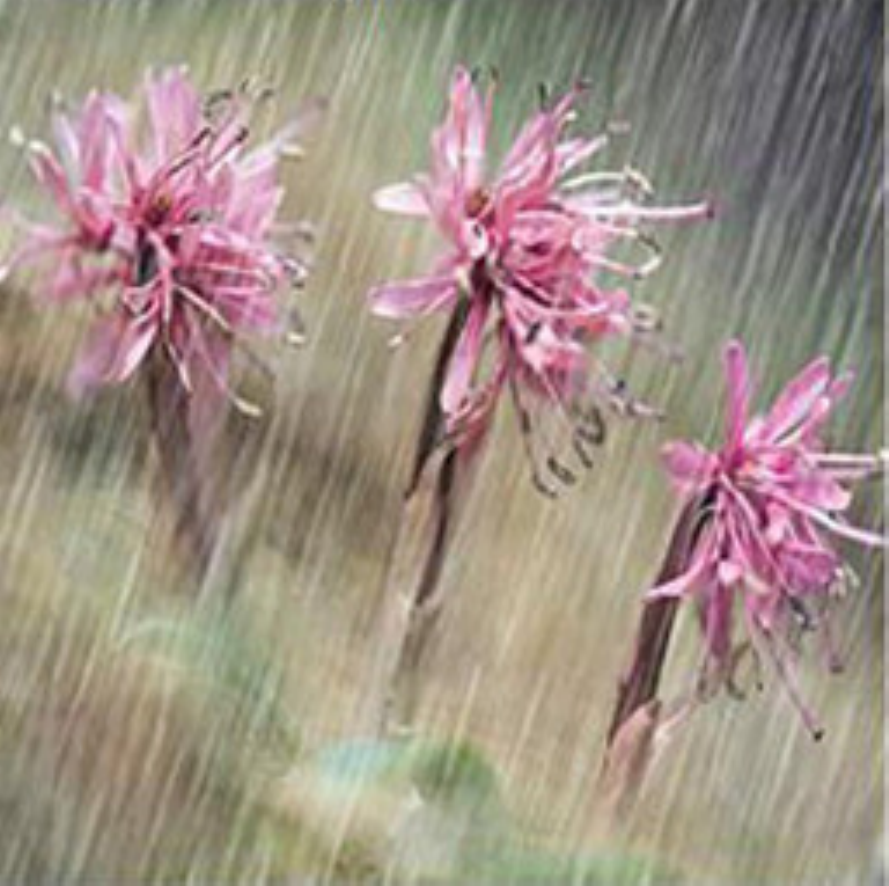}}
\centerline{(a)}
\end{minipage}
\hfill
\begin{minipage}{0.24\linewidth}
\centering{\includegraphics[width=1\linewidth]{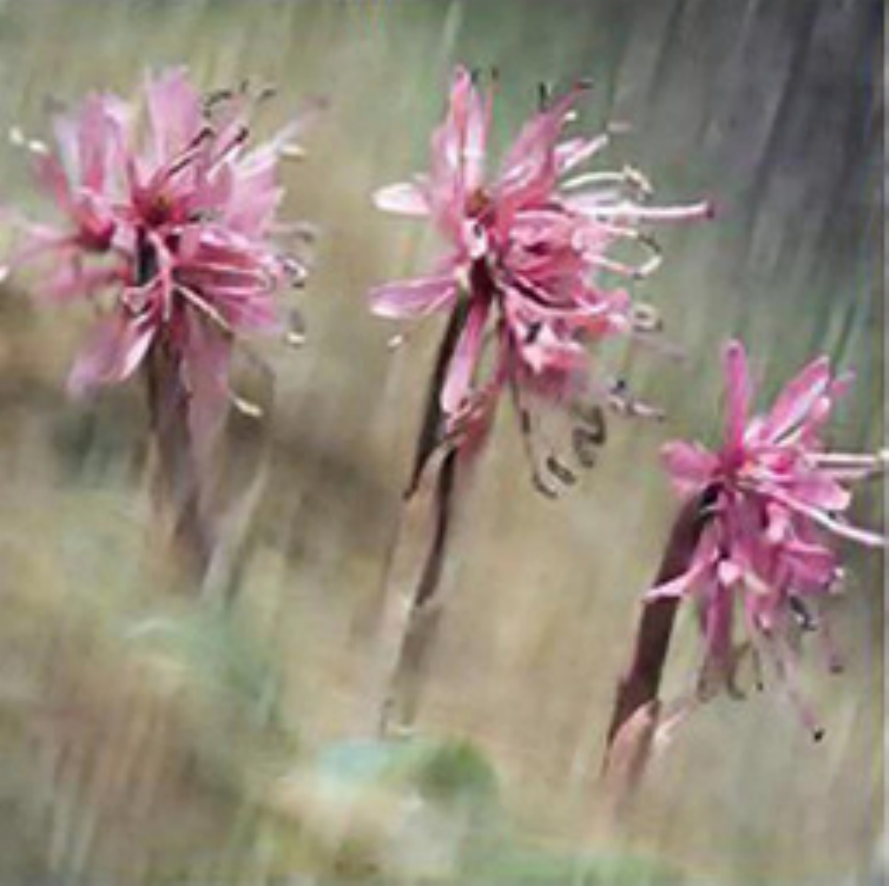}}
\centerline{(b)}
\end{minipage}
\hfill
\begin{minipage}{0.24\linewidth}
\centering{\includegraphics[width=1\linewidth]{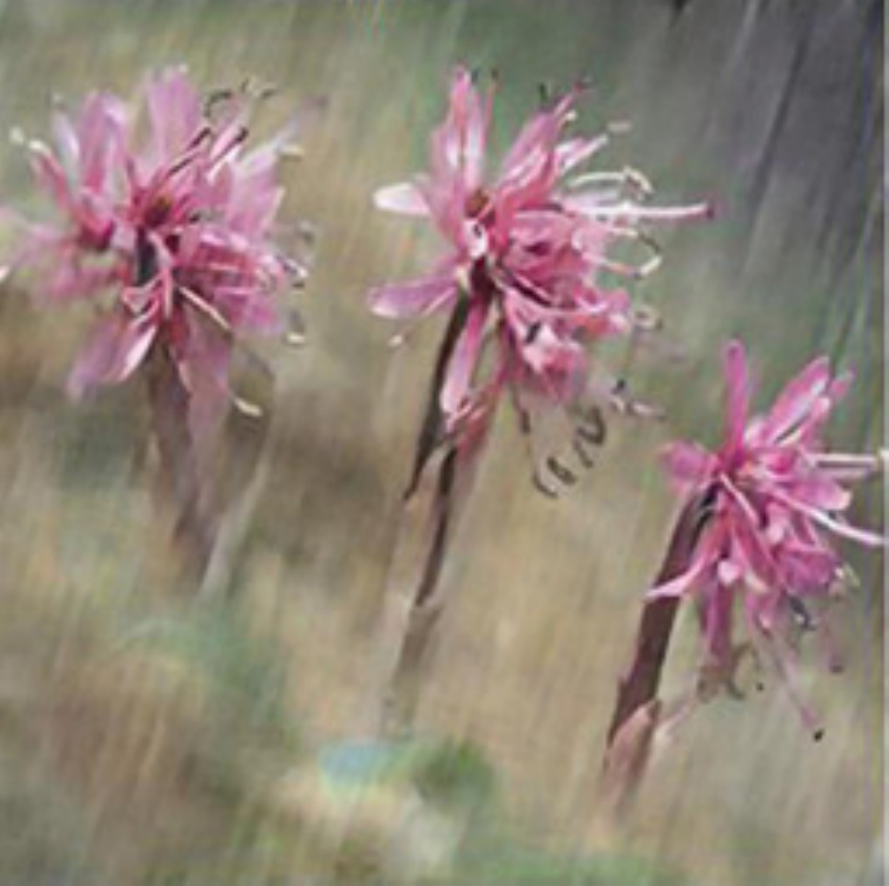}}
\centerline{(c)}
\end{minipage}
\hfill
\begin{minipage}{0.24\linewidth}
\centering{\includegraphics[width=1\linewidth]{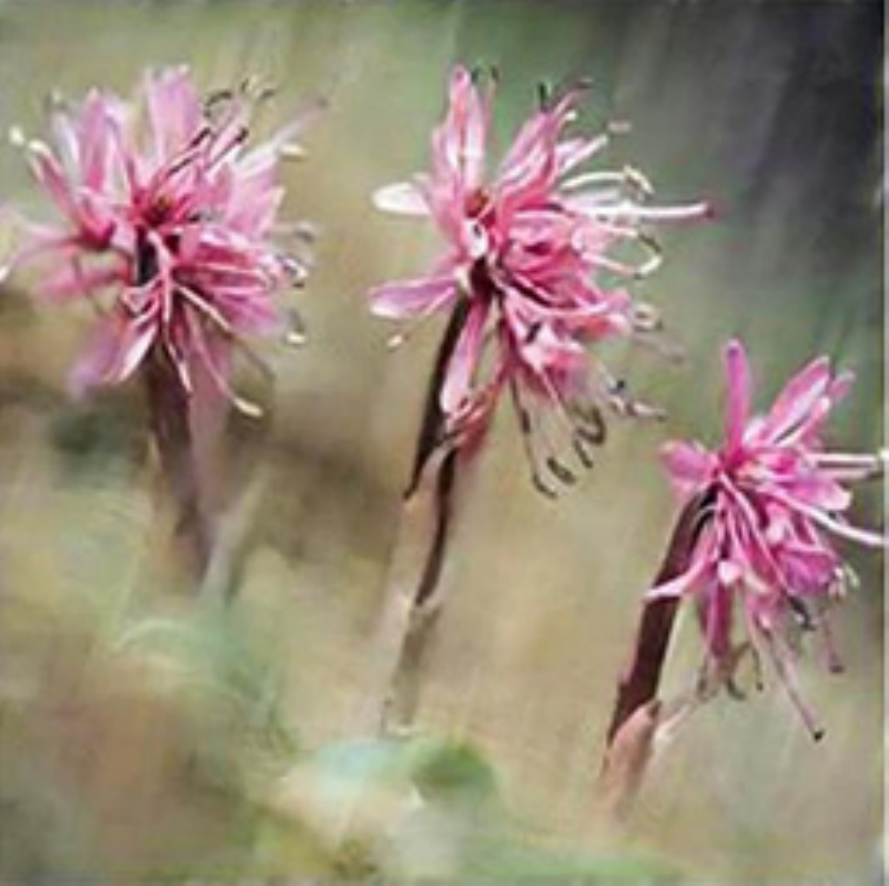}}
\centerline{(d)}
\end{minipage}
\vfill
\begin{minipage}{0.24\linewidth}
\centering{\includegraphics[width=1\linewidth]{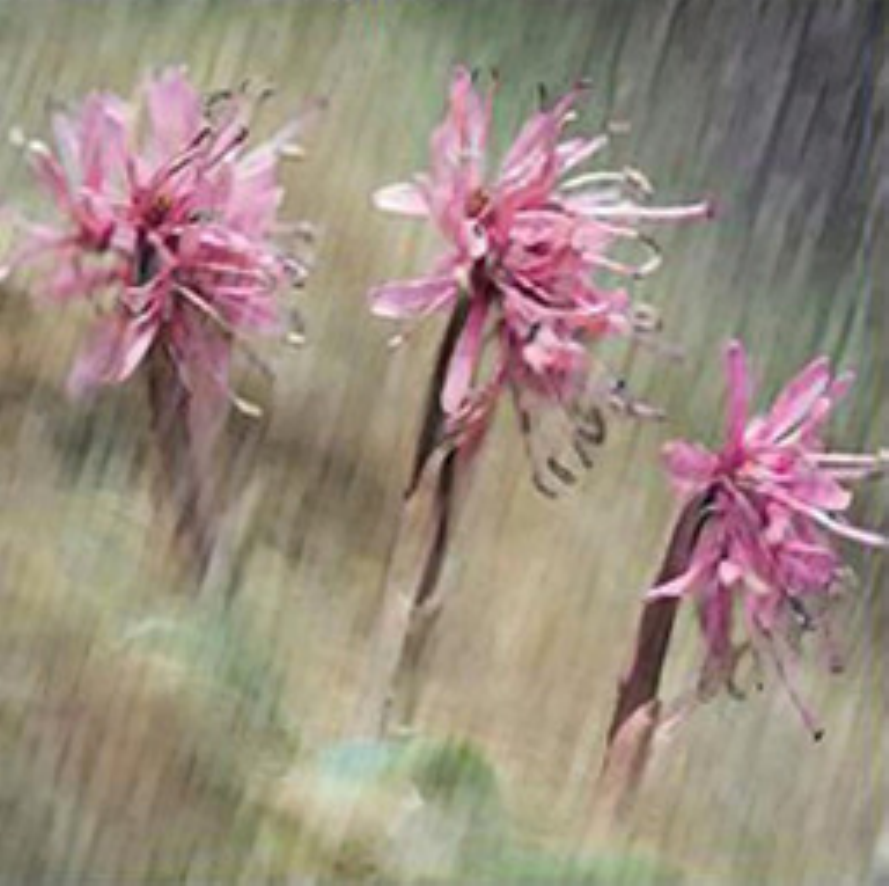}}
\centerline{(e)}
\end{minipage}
\hfill
\begin{minipage}{0.24\linewidth}
\centering{\includegraphics[width=1\linewidth]{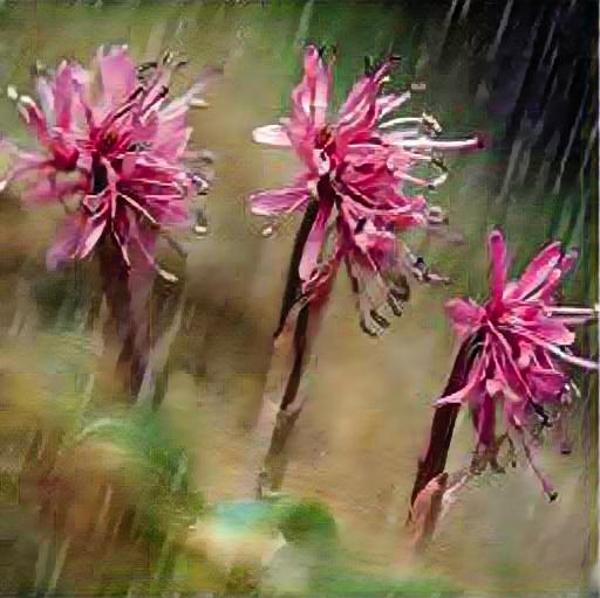}}
\centerline{(f)}
\end{minipage}
\hfill
\begin{minipage}{0.24\linewidth}
\centering{\includegraphics[width=1\linewidth]{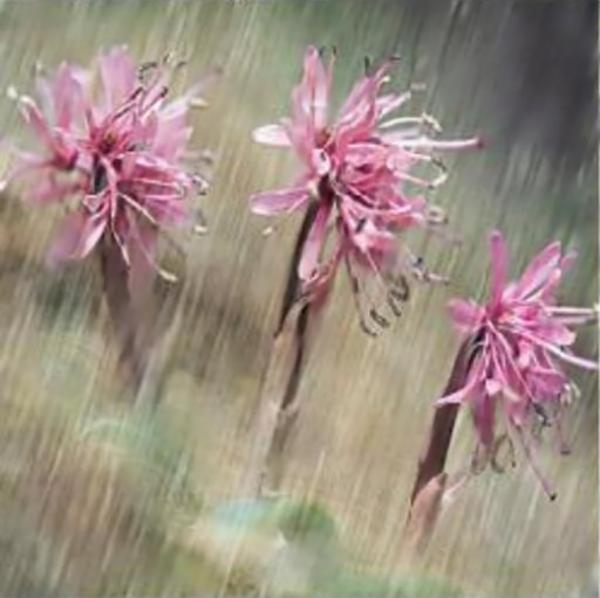}}
\centerline{(g)}
\end{minipage}
\hfill
\begin{minipage}{0.24\linewidth}
\centering{\includegraphics[width=1\linewidth]{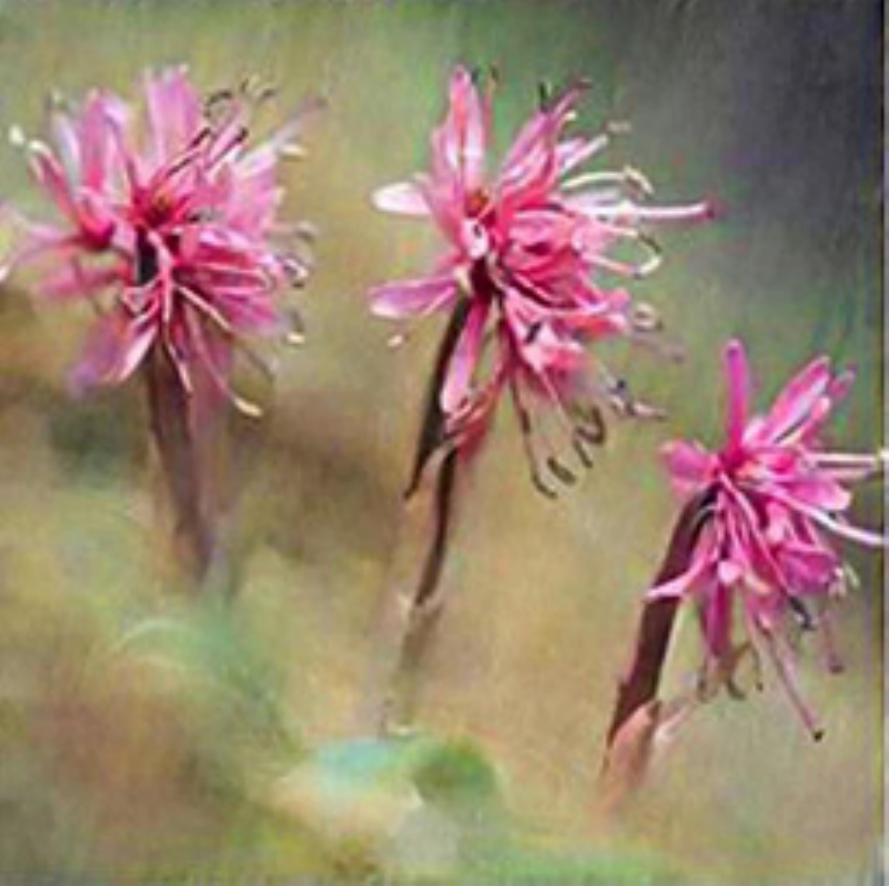}}
\centerline{(h)}
\end{minipage}
\end{center}
\caption{{(a) Input rainy image. (b)-(f) Deraining results of \cite{Fu_2017_CVPR,Yang_2017_CVPR,Zhang_2018_CVPR,Li_2018_ECCV,Li_rt_2019_CVPR,Wang_ty_2019_CVPR} and our method. Our method removes the obvious heavy rain streaks and recovers the colors of the scene by removing the haze-like effect.}}
\label{fig:example1}
\end{figure}

The prevalence of rain, particularly in some locations, not only severely reduces the images quality captured by cameras, but more importantly impacts negatively upon the robustness of devices and/or algorithms that must operate continuously irrespective of the weather. For example, the inability of driverless cars to operate in the rain has become a notorious issue\footnote{See the Bloomberg Businessweek article `Self-Driving Cars Can Handle Neither Rain nor Sleet nor Snow' on 17 Sept. 2018}.
Most of the early attempts use videos (e.g.  \cite{Garg_2004_CVPR}) as utilizing temporal correlation helps to improve the results. Dictionary learning \cite{Mairal_2010_JMLR} contributes a lot to single image deraining (e.g. \cite{Wang_2017_TIP,Wang_2016_ICIP}). Recently, deep learning has also been applied to rain removal and achieved remarkable results \cite{Fu_2017_CVPR,Yang_2017_CVPR,Zhang_2018_CVPR,Fu_2017_TIP,Li_2018_MM}. Given an observed rainy image $\bI$, the deep learning-based methods either train a network to estimate a clear (rain-free) image $\mathbf{B}$ from $\mathbf{I}$ directly, e.g., \cite{Pan_2018_arxiv}, or estimate a residual (rain layer) $\mathbf{R}$ and obtain $\mathbf{B}$ via $\mathbf{B}=\mathbf{I}-\mathbf{R}$, e.g., \cite{Zhang_2018_CVPR}.
Learning to estimate the clear image directly usually ignores the imaging process of the rainy images. The methods seek to learn an end-to-end mapping function from the appearance of the observations and the latent clear images from limited synthetic image pairs, which leads to limited robustness and generalization \cite{Pan_2018_arxiv,gong2018learning}.
The residual estimation based methods (e.g., \cite{Fu_2017_CVPR}) treat the rainy images as a simple summation of the clear background and a rain layer. Since estimating rain layer is usually easier than predicting the diverse background, these models work more reliably.

\begin{figure*}[t]
\begin{center}
\begin{minipage}{1\linewidth}
\centering{\includegraphics[width=1\linewidth]{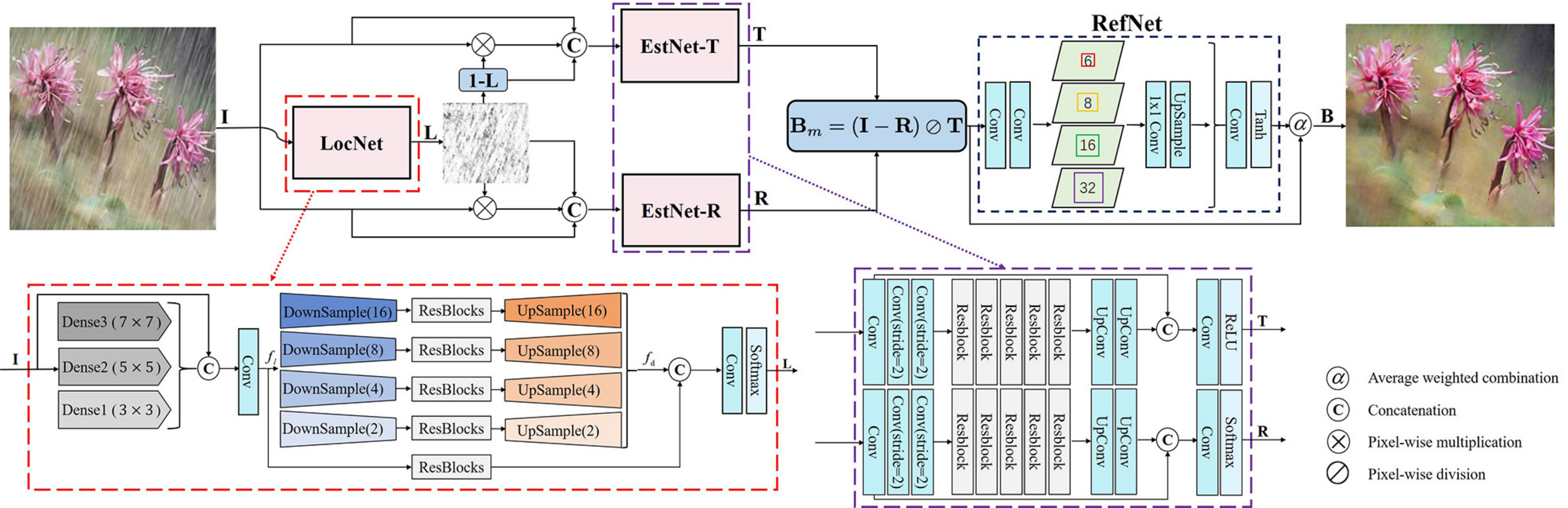}}
\end{minipage}
\end{center}
\caption{The architecture of AMPE-Net. LocNet captures the location of the strong rain streaks.
EstNet-T and EstNet-R estimate the parameters corresponding to $\mathbf{T}$ and $\mathbf{R}$ in the rain model. In RefNet, we use SPP modules with the factors $6$, $8$, $16$ and $32$, respectively. The component $\alpha$ represents a weighted combination operation with coefficient $\alpha$.}
\label{fig:whole_net}
\end{figure*}

However, existing deep learning methods often fail to obtain satisfactory results for many realistic cases, as shown in Figure ~\ref{fig:example1}. Under many realistic rainy scenes, not only the rain streaks may affect the visual quality of the images, but also the rainy images usually suffer from a layer of haze-like effect or mist \cite{Li_sy_2019_CVPR}, especially in heavy rain conditions. This is mainly caused by the scattering of accumulated tiny raindrops, which leads to the color degeneration \cite{Narasimhan_2002_IJCV}. Moreover, the boundaries of the rain streaks tend to be blurry under such situation, which renders more difficulties in the deraining task.

To tackle the above issues and remove the rain effect more completely, we take the mist effect into consideration while modeling the rainy image $\mathbf{I}$.
By letting $\bB$ denote the rain-effect-free clear image, we model the rainy image $\mathbf{I}$ as $\mathbf{I}=\mathbf{T} \circ \mathbf{B}+\mathbf{R}$, where $\mathbf{T}$ denotes the transmission of the mist effect, $\bR$ models the apparent rain streaks and $\circ$ denotes pixel-wise multiplication.
Unlike the commonly used residual model $\mathbf{I}=\mathbf{B}+\mathbf{R}$, we use an additional coefficient $\mathbf{T}$ to model the transmission of the haze-like mist effect explicitly.
Instead of merely training an end-to-end network for recovering $\bB$ from $\bI$, considering that the rain model itself contains a more specific structure of the deraining task, we integrate the rain model into the design of the deep neural network.
Specifically, we propose a two-branch model-based deep network for single image deraining via rAin Model Parameter Estimation, referred to as \emph{AMPE-Net}. The network framework is shown in Figure \ref{fig:whole_net}.

In the proposed model, two processing branches (i.e., \emph{EstNet-T} and \emph{EstNet-R}) are jointly trained to predict the coefficients corresponding to $\mathbf{T}$ and $\bR$, and the network estimates the results $\bB$ according to the rain model. We train a subnetwork (\emph{LocNet}) to help to localize the rain streaks, which provides a guide for the following EstNet-T and EstNet-R as shown in Figure \ref{fig:whole_net}.
In some realistic cases, it is observed that the haze-like mist effect may be ``over-removed'', which may result in over bright colors in the results. Although it is a subjective assessment due to the lack of the ground truth for the real image, we add an additional component (\emph{RefNet}) to give the model flexibility to control the strength for removing the mist effects.

We summarize the main contributions as the following.
\begin{itemize}
\item We model the rainy image via a new rain model that explicitly formulates the haze-like mist effect (caused by the scattering of the accumulated tiny raindrops). As a result, not only the rain streaks, but also the haze-like mist can be taken into consideration for deraining.
\item According to our rain model, we propose a two-branch network, i.e. AMPE-Net, for deraining, which incorporates the rain imaging model and the deep learning based methods. In the proposed AMPE-Net, two branches are jointly trained to estimate the rain model parameters, which are explicitly applied for recovering the clear image. Moreover, a refinement subnetwork is designed to give flexibility to the model to control the strength of mist effect removal.
\item
Unlike previous methods mainly focusing on removing the rain streaks, our model can remove various effects caused by rain, from large raindrops to haze-like effect.
It can also flexibly control the haze-like effect to produce different visual effects.
We conduct extensive experiments on different datasets to show the effectiveness of the proposed techniques.
We also simply show the potentials of our method on dehazing task.
\end{itemize}

\section{Related Work}

Single-image rain removal has gained much success and popularity recently. The main attempt is using dictionary learning \cite{Mairal_2010_JMLR} to decompose rainy images \cite{Kang_2012_TIP,Wang_2017_TIP}.
Very recently, deep learning has been used in many image restoration tasks \cite{ledig2017photo,gong2017motion,yang2018seeing} including rain removal.
A deep detail network was proposed to reduce the mapping range from input to output \cite{Fu_2017_CVPR}, to make the learning process easier. Moreover, they extended the work by decomposing the rain image into low and high-frequency components and extract image details from the high-frequency component \cite{Fu_2017_TIP}. These two methods are particularly good for removing light rain but have issues of removing bright or blurry rain streaks.
Yang et al. add a binary map to locate the rain streak. They create a new model to represent rain streak accumulation, and various shapes and directions of overlapping rain streaks \cite{Yang_2017_CVPR}. Their method is very good for removing bright rain streaks but often fails for removing blurry rain streaks. Zhang et al. \cite{Zhang_2018_CVPR} propose a multi-stream dense network that can automatically determine the rain-density information and thus can efficiently remove the corresponding rain-streaks according to the estimated rain-density label. This method can handle a diverse range of rainy images, but sometimes causes blur in image details. To model and remove rain streaks of various size and the veiling effect, a multi-stage network consisting of several parallel sub-networks was designed, each of which models a different scale of rain streaks \cite{Li_2017_arxiv}.
Li et al. \cite{Li_2018_ECCV} remove the rain streaks via multiple stages and use a recurrent neural network to exchange information across stages.
A non-locally enhanced encoder-decoder network framework is proposed, which captures long-range spatial dependencies via skip-connections and learns increasingly abstract feature representation while preserving the image detail by pooling indices guided decoding \cite{Li_2018_MM}.

\section{The Proposed Method}
Given an observed rainy image $\mathbf{I}$, the goal of deraining is to recover a clean image $\mathbf{B}$. Our target is to train a neural network (i.e. AMPE-Net) to estimate $\mathbf{B}$ from $\mathbf{I}$ by:
\begin{equation}
\mathbf{B} = \mathcal{U} (\mathbf{I}),
\end{equation}
where $\mathcal{U}(\cdot)$ denotes the proposed AMPE-Net. Unlike the commonly-used rain model $\mathbf{I}=\mathbf{B}+\mathbf{R}$, we propose to integrate our rain model into the neural network to guide the estimation of parameters. Before introducing the network $\mathcal{U}(\cdot)$, we will first
remodel a rainy image $\mathbf{I}$ to express the relationship between $\mathbf{I}$
and $\mathbf{B}$ more completely.

\subsection{Rainy Image Modeling}
\label{sec:linear_rain_model}

Majority of existing deraining methods \cite{Zhang_2018_CVPR,Yang_2017_CVPR} model a rainy image $\mathbf{I}$ as a summation of the background $\mathbf{B}$ and the rain layer $\mathbf{R}$:
\begin{equation}\label{eq:original_model}
\mathbf{I} = \mathbf{B} + \mathbf{R}.
\end{equation}
It has been proven that the residual estimation models based on Eq. \eqref{eq:original_model} can work well for handling the rain streaks \cite{Zhang_2018_CVPR,Yang_2017_CVPR,Li_2018_ECCV}. The apparent rain streaks and drops are usually significant in the rainy images and are assumed to have similar falling directions and shapes \cite{Li_2018_ECCV}. However, in many realistic scenarios, not only large rain streaks but also the accumulated tiny raindrops may influence the imaging quality \cite{Yang_2017_CVPR}. The tiny raindrops in the air impair the images by accumulating together and impeding the
propagation of light via scattering. In the rainy images, they appear as a layer of haze-like mist effect, which desaturates the colors of the background and leads to low contrast.
In this case, the edges of rain streaks also become blurry and merge into the mist,
rendering more difficulties for deraining (e.g. Figure \ref{fig:example1}).

Considering that the haze-like mist generally exists in realistic rainy images, model \eqref{eq:original_model} based methods, e.g., \cite{Yang_2017_CVPR}, often produce unsatisfactory results suffering from haze-like effects and remaining rain streaks, as shown in Figure \ref{fig:example1}.
To handle above problems, we take the scattering of accumulated tiny raindrops into consideration and add a variable $\mathbf{T}$ to formulate the influence of tiny raindrops:
\begin{equation}\label{eq:linear_model}
\mathbf{I} = \mathbf{T} \circ \mathbf{B} + \mathbf{R},
\end{equation}
where $\mathbf{T}$ is to model haze-like effect and $\mathbf{R}$ models the apparent rain streaks.
In the following, we design the deraining network based on our improved model in \eqref{eq:linear_model}.

\subsection{The Proposed AMPE-Net for Deraining}
According to the rain model in Eq. \eqref{eq:linear_model}, given a rainy image $\mathbf{I}$, if we can obtain the corresponding parameters $\mathbf{T}$ and $\mathbf{R}$, the clean image $\mathbf{B}$ can be predicted through:
\begin{equation}
\widehat{\mathbf{B}}_{m} = (\mathbf{I}-{\mathbf{R}}) \oslash {\mathbf{T}},
\label{eq:linear_model_inverse}
\end{equation}
where $\widehat{\mathbf{B}}_{m}$ denotes the estimation of $\mathbf{B}$ by our rain model, and $\oslash$ is the point-wise division.
Eq. \eqref{eq:linear_model_inverse} formulates the internal structure of the deraining task.
However, estimating $\mathbf{T}$ and $\mathbf{R}$ from $\mathbf{I}$ is non-trivial.
We thus incorporate the model in Eq. \eqref{eq:linear_model_inverse} and the deep learning methods.
Since high-quality estimation of the clear image $\mathbf{B}$ is the final objective, instead of imposing supervision on $\mathbf{T}$ and $\mathbf{R}$ to train the network, we integrate the model \eqref{eq:linear_model_inverse} into the neural network and optimize the quality of the output $\mathbf{B}$. Specifically, we estimate $\mathbf{T}$ and $\mathbf{R}$ via two subnetworks in the whole model which is supervised on $\mathbf{B}$. Note that no explicit supervisions on $\mathbf{T}$ and $\mathbf{R}$ are applied. Figure \ref{fig:whole_net} shows the architecture of our model, i.e., AMPE-Net.

Our proposed AMPE-Net is mainly a two-branch model, which consists of two parallel subnetworks EstNet-T and EstNet-R to estimate $\mathbf{T}$ and $\mathbf{R}$, respectively, for the estimation model in Eq. \eqref{eq:linear_model_inverse}. Considering that $\mathbf{R}$ is mainly used to model the significant rain streaks, we apply a rainy streak localization subnetwork LocNet, which predicts a map of the significant rain streaks as the guidance of the following EstNet-T and EstNet-R, as shown in Figure \ref{fig:whole_net}.

After obtaining EstNet-T and EstNet-R, the model predicts the clear image via Eq. \eqref{eq:linear_model_inverse}. Eq. \eqref{eq:linear_model_inverse} is differentiable, the two-branch unit (with EstNet-T and EstNet-R) is trained via the supervision signal on the final estimation $\mathbf{B}$. Considering that previous residual estimation based methods, e.g., \cite{Li_2018_ECCV} obtain the results via $\mathbf{B}=\mathbf{I}-\mathbf{R}$ (where $\mathbf{R}$ is estimated by a network), the proposed model can be seen as an advanced version of the model-based estimation. The proposed model is based on a more realistic model and renders higher flexibility and strength for universally removing rain effects.

\par
Based on the model \eqref{eq:linear_model_inverse}, we can estimate a clear image in which the rain streaks and mist effect are well removed.
Although the model-based estimation can obtain high-quality results, the model trained on synthetic data may produce visually over bright results on some realistic images, due to that the haze-like mist may be ``over-removed''. Considering that the visual assessment is subjective, we add an additional refining module RefNet to refine the estimation. It is jointly trained with the two-branch unit and gives the model flexibility to control the strength for removing the mist.

\par
In details, we formulate each subcomponent below and introduce the implementation details in the following.

\noindent \textbf{LocNet} LocNet takes the rainy image $\mathbf{I}$ as input and predicts a location map $\mathbf{L}$ of the rain pixels in $\mathbf{I}$ by
\begin{equation}\label{eq:L}
\widehat{\mathbf{L}}=\mathcal{H}(\mathbf{I}),
\end{equation}
where $\mathcal{H}(\cdot)$ denotes LocNet, and $\widehat{\mathbf{L}}$ is a continuous (non-binary) estimation distributed in $[0,1]$. The high values in $\widehat{\mathbf{L}}$ indicate the pixels suffering from significant rain streaks with high probabilities. LocNet $\mathcal{H}(\cdot)$ is trained under the supervision of the binary version of $\mathbf{L}$.

\noindent \textbf{EstNet-R} EstNet-R is used to estimate the coefficient corresponding to $\mathbf{R}$. $\mathbf{R}$ is expected to mainly represent the degeneration caused by the significant rain streaks. EstNet-R thus takes $\widehat{\mathbf{L}}$ and $\mathbf{I}\circ \widehat{\mathbf{L}}$ as input, apart from the observed image $\mathbf{I}$. It is the way that $\mathbf{L}$ guides the model-based estimation. By letting $\mathcal{G}(\cdot)$ represent EstNet-R, $\mathbf{R}$ can be estimated by
\begin{equation}\label{eq:B1}
\widehat{\mathbf{R}} = \mathcal{G}(\mathbf{I}, ~\widehat{\mathbf{L}}, ~\mathbf{I}\circ \widehat{\mathbf{L}} ).
\end{equation}

\noindent \textbf{EstNet-T} Similar to EstNet-R, EstNet-T is the coefficient corresponding to $\mathbf{T}$.
Apart from $\mathbf{I}$, it also takes $\mathbf{1}-\widehat{\mathbf{L}}$ and $\mathbf{I} \circ (\mathbf{1}-\widehat{\mathbf{L}})$ as input. Let $\mathcal{F}(\cdot)$ denote EstNet-T. $\mathbf{T}$ can be estimated by
\begin{equation}\label{eq:A1}
\widehat{\mathbf{T}} = \mathcal{F}(\mathbf{I}, \mathbf{1}-\widehat{\mathbf{L}}, \mathbf{I}\circ(\mathbf{1}-\widehat{\mathbf{L}})).
\end{equation}

\noindent \textbf{Rain-free image estimation}
After obtaining $\widehat{\mathbf{T}}$ and $\widehat{\mathbf{R}}$, we then obtain the train-effect-free clear image based on the model in Eq. \eqref{eq:linear_model_inverse}:
\begin{equation}\label{eq:model_results}
\widehat{\mathbf{B}}_{m} = (\mathbf{I}-\widehat{\mathbf{R}}) \oslash \widehat{\mathbf{T}}.
\end{equation}

\noindent \textbf{AMPE-Net} As discussed above, we define a RefNet $\mathcal{R}(\cdot)$ to further manipulate the results by $\mathcal{R}( \widehat{\mathbf{B}}_{m} )$. To control the strength of the refinement, we introduce a coefficient $\alpha$ to obtain the final estimation via a linear combination of $\mathcal{R}(\widehat{\mathbf{B}}_m )$ and  $\widehat{\mathbf{B}}_m$. We arrive the final estimation of the proposed AMPE-Net:
\begin{equation}\label{eq:S1}
\widehat{\mathbf{B}} = \mathcal{U}(I) = \alpha \widehat{\mathbf{B}}_{m}
                            + (1-\alpha) \mathcal{R}(\widehat{\mathbf{B}}_{m}).
\end{equation}
where $\alpha\in[0,1]$.
During training, we set $\alpha=0.9$ to fit the samples. In testing, the strength of ``removing haze-like effect'' can be controlled by tuning $\alpha$. More analysis and experiments are left in the following sections.

\subsection{Network Structure of LocNet $\mathcal{H}(\cdot)$}

Since rain streaks and raindrops usually have different sizes and scales, only using convolutional kernels with single size cannot always extract useful features.
Inspired by \cite{Zhang_2018_CVPR,yan2019attention}, three densely connected convolutional modules \cite{Huang_2017_CVPR} are utilized in LocNet (Figure \ref{fig:whole_net}) to extract multi-scale shallow features of $\mathbf{I}$.
The kernel sizes of the three densely-connected blocks are $7 \times 7$, $5 \times 5$ and $3 \times 3$, respectively.
We concatenate the obtained features with $\mathbf{I}$ to form the shallow-layer feature $f_{l}$ after a Conv layer.

\begin{figure}[t!]
\begin{center}
\begin{minipage}{0.24\linewidth}
\centering{\includegraphics[width=1\linewidth]{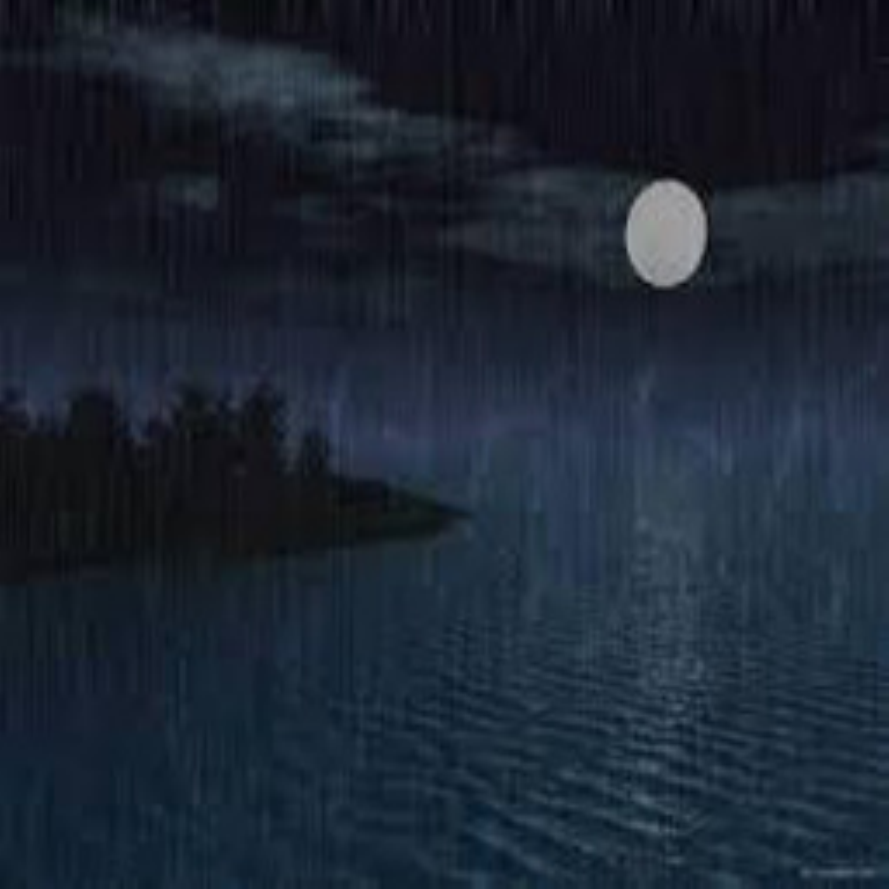}}
\end{minipage}
\hfill
\begin{minipage}{0.24\linewidth}
\centering{\includegraphics[width=1\linewidth]{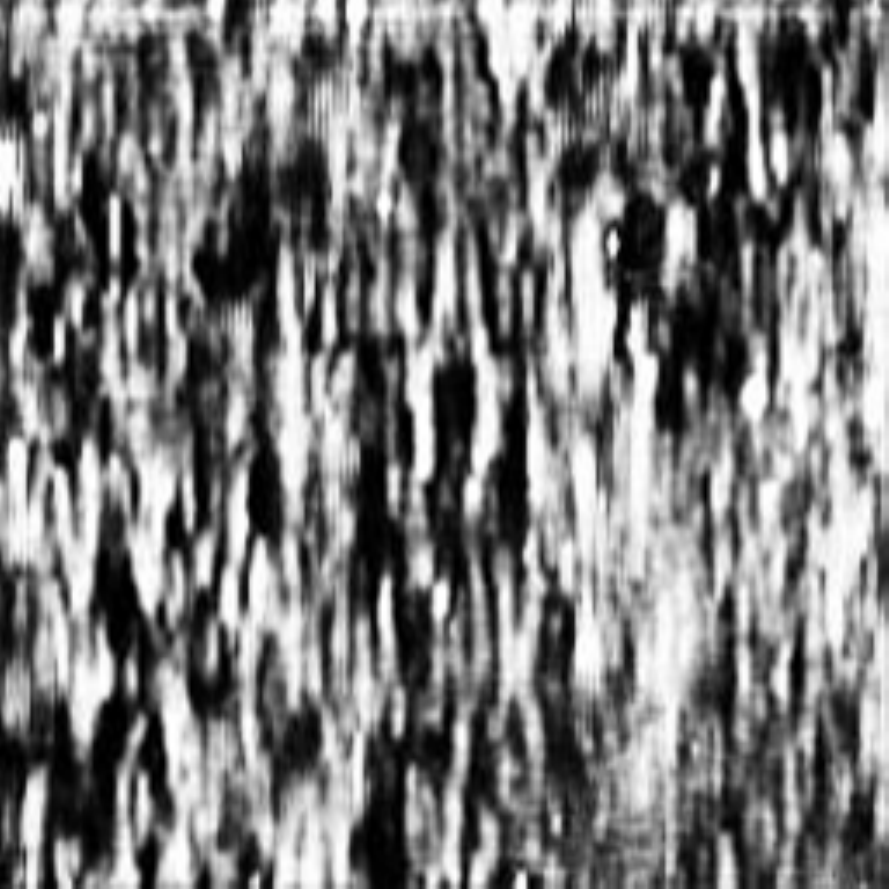}}
\end{minipage}
\hfill
\begin{minipage}{0.24\linewidth}
\centering{\includegraphics[width=1\linewidth]{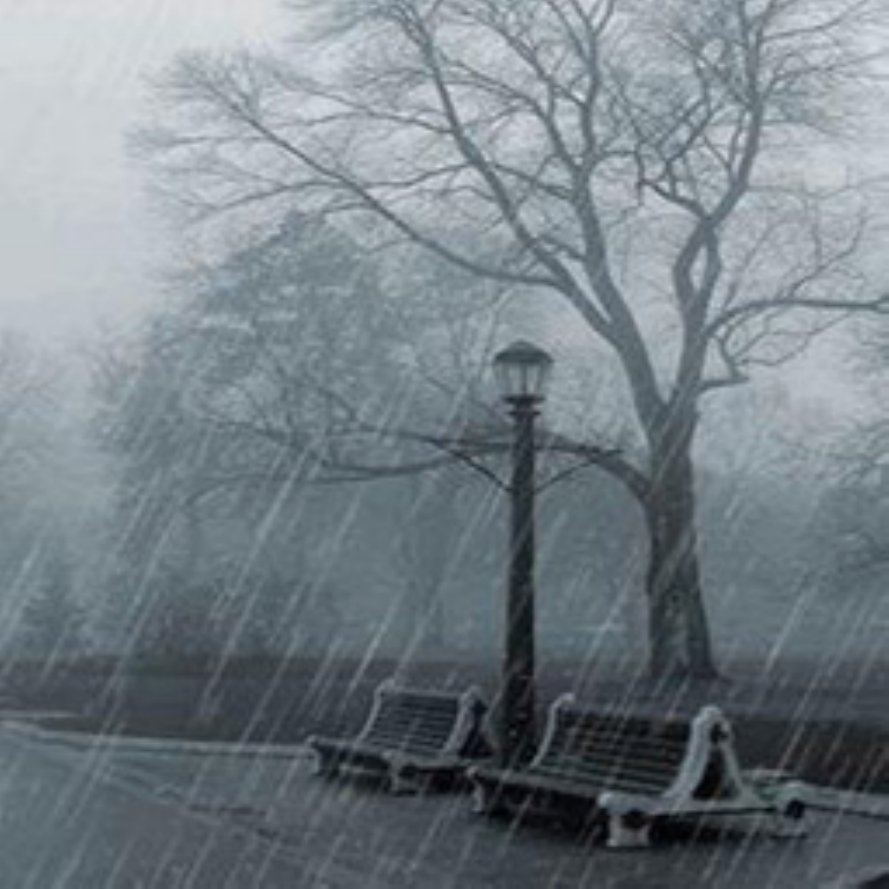}}
\end{minipage}
\hfill
\begin{minipage}{0.24\linewidth}
\centering{\includegraphics[width=1\linewidth]{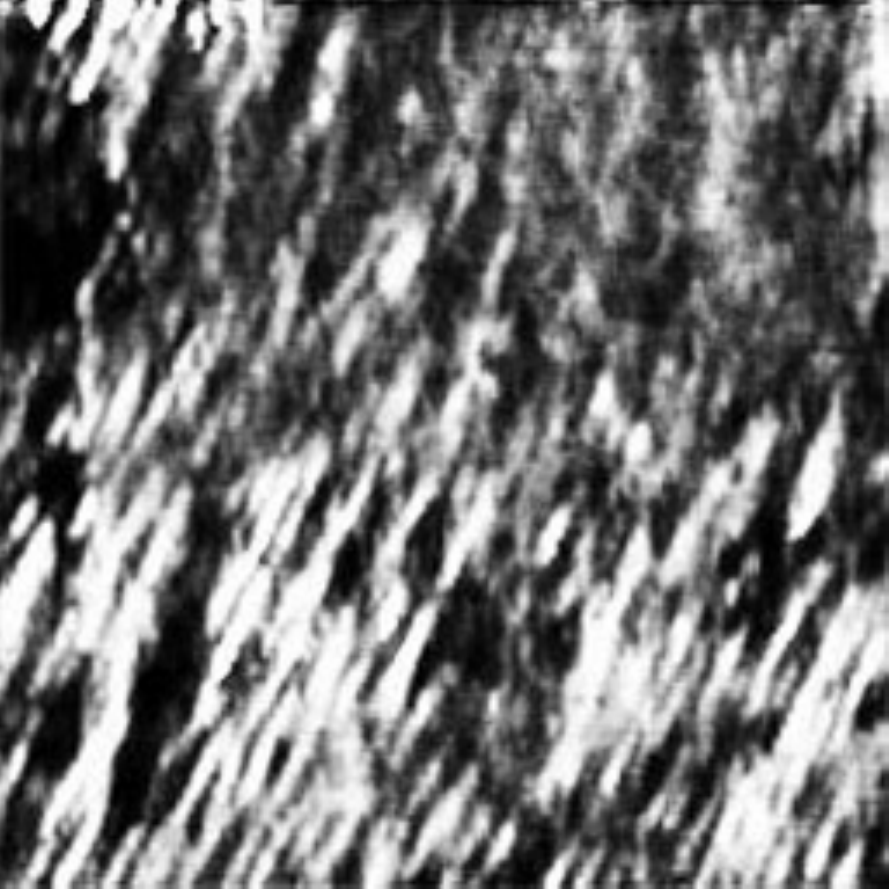}}
\end{minipage}
\end{center}
\caption{From left to right are two rainy images and their corresponding location maps.}
\label{fig:loca}
\end{figure}

The core part of our LocNet composes of down-sampling, details extraction, and up-sampling operations with four different scale factors (i.e., $16$, $8$, $4$ and $2$).
We use $5$ ResBlocks \cite{He_2015_CVPR} to extract deep features, and then up-sample the features to original size to form the deep-layer features $f_{d}$.
After concatenating the shallow and deep features, a convolutional layer is used to fuse the combined features.
At last, we utilize a softmax function to estimate the location map $\widehat{\mathbf{L}}$.
Two examples of the estimated location maps are shown in Figure \ref{fig:loca}.

\subsection{Network Structure of EstNet-R $\mathcal{G}(\cdot)$ and EstNet-T $\mathcal{F}(\cdot)$}

In the two-branch estimation unit, a $9 \times 9$ convolution layer is firstly utilized to extract the features of the guided input.
Two down-sampling layers are then applied for extracting features in larger areas.
Furthermore, we use five ResBlocks and two times up-samplings.
To avoid checkerboard artifacts, we up-sample directly with a following convolution layer to replace to deconvolve the feature maps.
After up-sampling the feature map, we concatenate it with the output of the first convolution layer.
At last, $\mathbf{T}$ is estimated by a composition of convolution and ReLU operation.
As shown in Figure \ref{fig:whole_net}, apart from the activation function at the output layer, the network structures of EstNet-R is the same as EstNet-T.

\subsection{Network Structure of RefNet $\mathcal{R}(\cdot)$}

The main goal of RefNet $\mathcal{R}(\cdot)$ is to adjust the color of the results.
In $\mathcal{R}(\cdot)$, two convolution layers are first used.
Then an SPP module \cite{He_2015_arxiv} is utilized to obtain multi-scale features.
The scale factors are $4$, $8$, $16$ and $32$, respectively.
For the feature maps with different size, we adopt pointwise convolution \cite{Sifre_2014_Phd_thesis} to reduce their channels
and up-sample them by the nearest interpolation method to original size.
The refined result is obtained via convolution and Tanh activation function
on the concatenated multi-scale features successively.
At last, the weighted average combination is utilized to integrate refined result and the result before being refined to obtain our final
rain removed results.

\subsection{Training Loss}
During training, given a dataset $\{(\mathbf{I}_{t}, \mathbf{L}_{t}) \}^{N}_{t=1}$ with ground truth rain streak map $\mathbf{L}$, we first train the LocNet $\mathcal{H}(\cdot)$ by predicting $\mathbf{L}$. We then train the rest part of the AMPE-Net (with subnetworks $\mathcal{F}(\cdot)$, $\mathcal{G}(\cdot)$ and  $\mathcal{R}(\cdot)$) jointly on the training set $\{(\mathbf{I}_{t}, \mathbf{B}_{t}) \}^{M}_{t=1}$.

\noindent \textbf{Training loss for LocNet}~
In LocNet, we use a softmax layer to obtain the estimation of the rain streak map. By letting the output $\mathcal{H}(\mathbf{I}_{t})$ indicate the location of the rain streaks, we apply an MSE loss function to fit $\mathbf{L}_{t}$:
\begin{equation}\label{eq:loss_function_loca}
\mathcal{L}_{\mathbf{L}} = \sum^{N}_{t=1} \| \mathcal{H}(\mathbf{I}_{t})-\mathbf{L}_{t} \|^{2}_{F}.
\end{equation}
In practice, we find that training with MSE loss can obtain satisfactory guidance for the subsequent networks more stably and quickly.

\noindent \textbf{Training loss for rain-removal}~
To fully use the constraints of our rain model to optimize the parameters of network,
we minimize the following two MSE loss functions:
\begin{equation}\label{eq:loss_function2}
\mathcal{L}_{1} = \sum^{M}_{t=1} \| \mathbf{B}_{t} - (\mathbf{I}_{t} - \widehat{\mathbf{R}}) \oslash \widehat{\mathbf{T}} \|^{2}_{F},
\end{equation}
\begin{equation}\label{eq:loss_function1}
\mathcal{L}_{2} = \sum^{M}_{t=1} \| \mathbf{I}_{t} - \widehat{\mathbf{T}} \circ \mathbf{B}_{t} - \widehat{\mathbf{R}} \|^{2}_{F}.
\end{equation}
By using the loss function $\mathcal{L}_2$ in Eq. \eqref{eq:loss_function1}, the training is further restricted and achieves more robust results. To avoid involving an additional hyper-parameter for balancing $\mathcal{L}_1$ and $\mathcal{L}_2$, in training, the two loss functions are applied alternatively with different batches of training samples.
Hence, we arrive the loss function $\mathcal{L}_{m}$ for optimizing EstNet-R and EstNet-T and obtaining the rain-free result $\widehat{\mathbf{B}}_{m}$:
\begin{equation}\label{eq:loss_function3}
\mathcal{L}_{m} = \left \{
                                \begin{array}{lr}
                                \mathcal{L}_{1}, & i =1, 3, 5, ...  \\
                                \mathcal{L}_{2}, & i =2, 4, 6, ...
                                \end{array},
                           \right.
\end{equation}
where $i$ denote the iteration (or data mini batch) index during the training.

\par
In the proposed model, the estimated $\widehat{\mathbf{B}}_{m}$ is refined via the RefNet $\mathcal{R}(\cdot)$ and the weighted average combination defined in Eq. \eqref{eq:S1}. The RefNet is jointly trained using the following loss function
\begin{equation}\label{eq:loss_function4}
\mathcal{L}_{r} = \sum^{M}_{t=1} \| \alpha \widehat{\mathbf{B}}_{m} + (1-\alpha)\mathcal{R}(\widehat{\mathbf{B}}_{m})-\mathbf{B}_{t} \|^{2}_{F}.
\end{equation}
Note that $\alpha$ is fixed as $0.9$ in training and can be turned during testing.

\par
Accordingly, we arrive the final loss function:
\begin{equation}\label{eq:loss_function5}
\mathcal{L} = \mathcal{L}_{m} + \mathcal{L}_{r}.
\end{equation}

\begin{figure}[!t]
\begin{center}
\begin{minipage}{0.19\linewidth}
\centering{\includegraphics[width=1\linewidth]{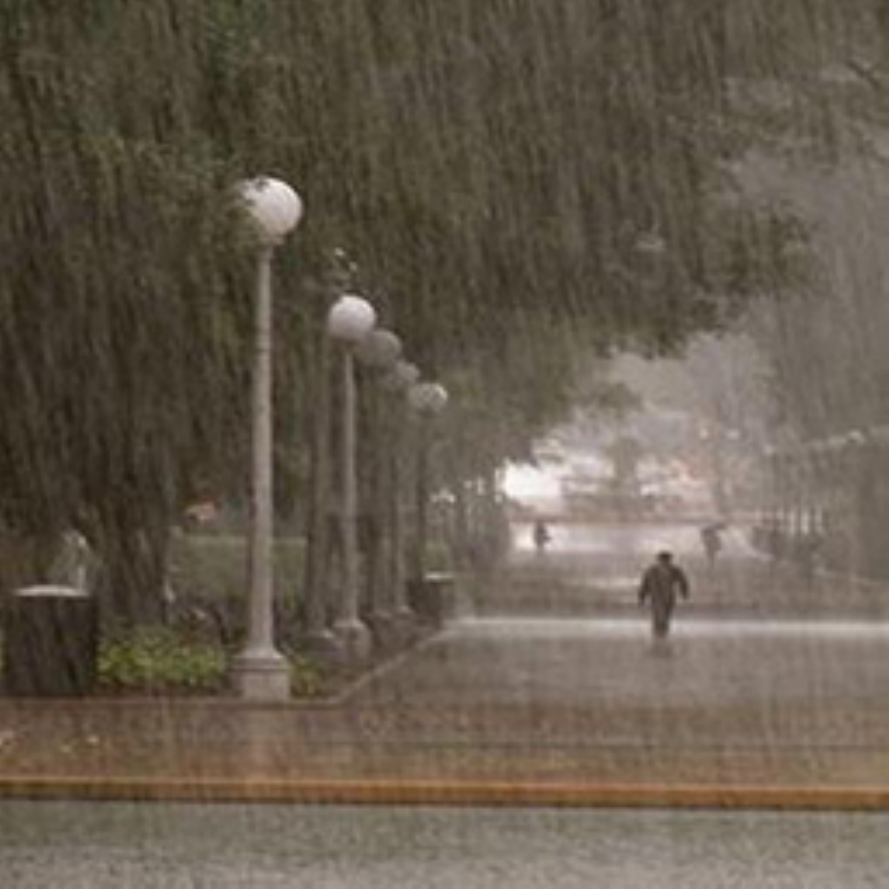}}
\end{minipage}
\hfill
\begin{minipage}{0.19\linewidth}
\centering{\includegraphics[width=1\linewidth]{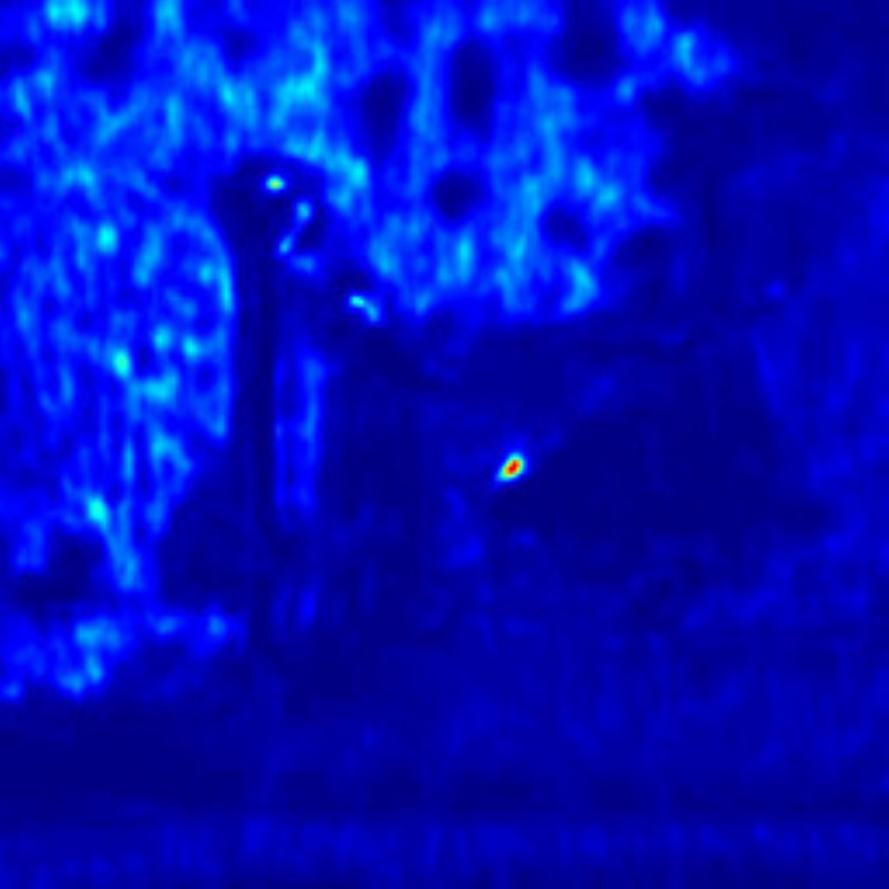}}
\end{minipage}
\hfill
\begin{minipage}{0.19\linewidth}
\centering{\includegraphics[width=1\linewidth]{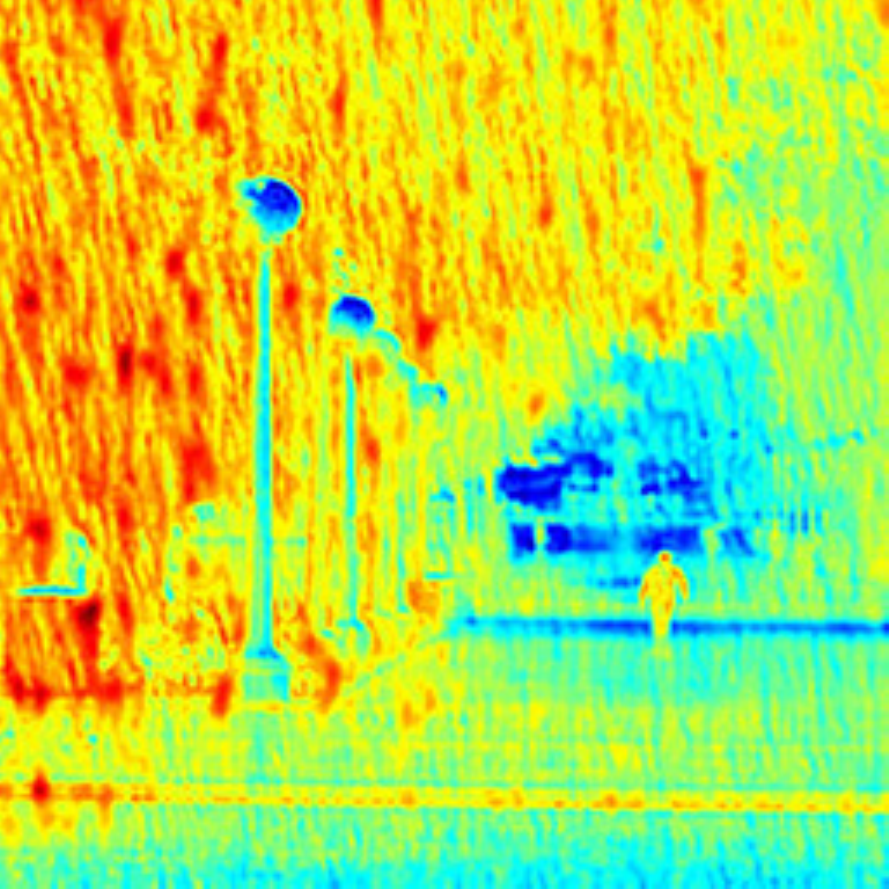}}
\end{minipage}
\hfill
\begin{minipage}{0.19\linewidth}
\centering{\includegraphics[width=1\linewidth]{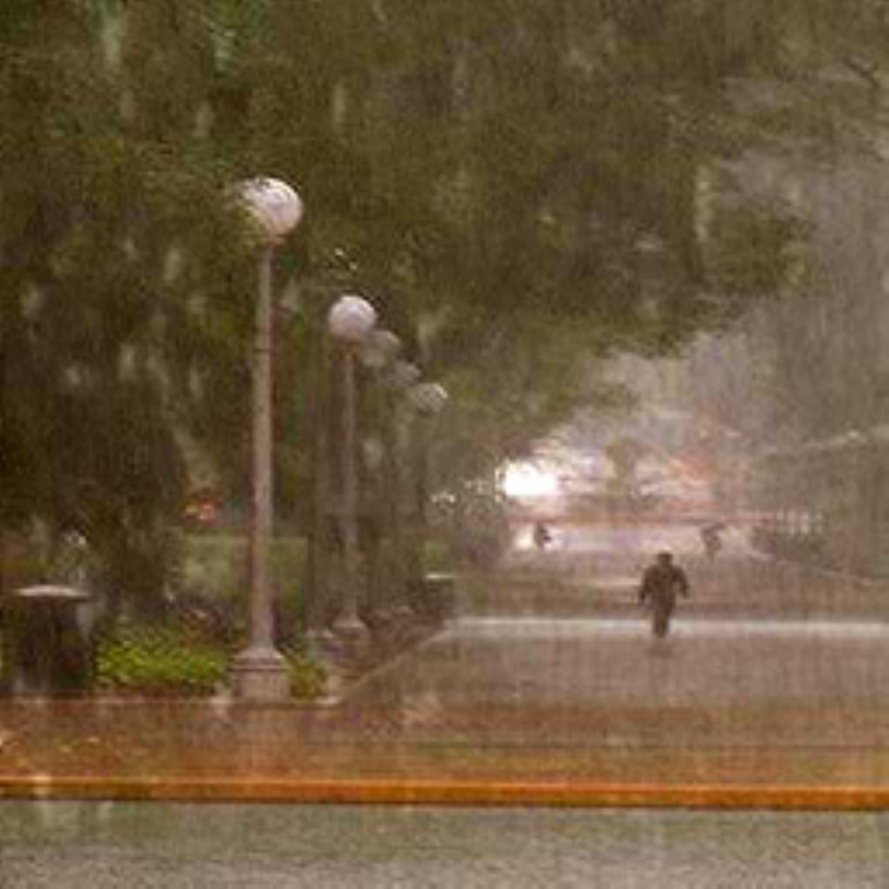}}
\end{minipage}
\hfill
\begin{minipage}{0.19\linewidth}
\centering{\includegraphics[width=1\linewidth]{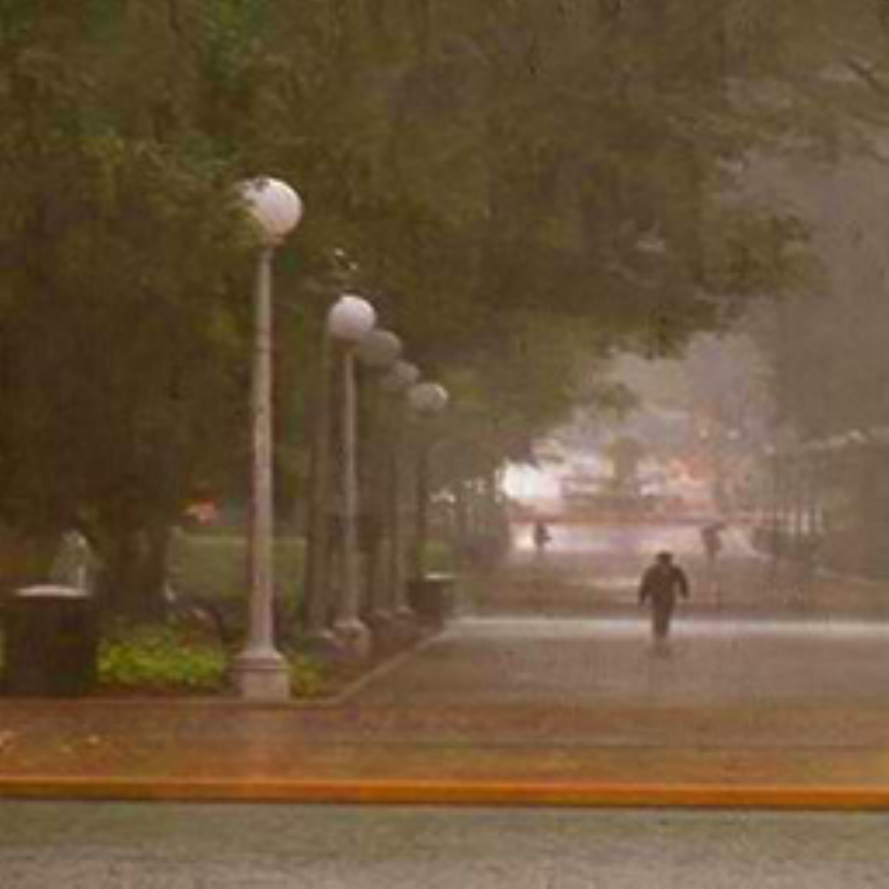}}
\end{minipage}
\vfill
\begin{minipage}{0.19\linewidth}
\centering{\includegraphics[width=1\linewidth]{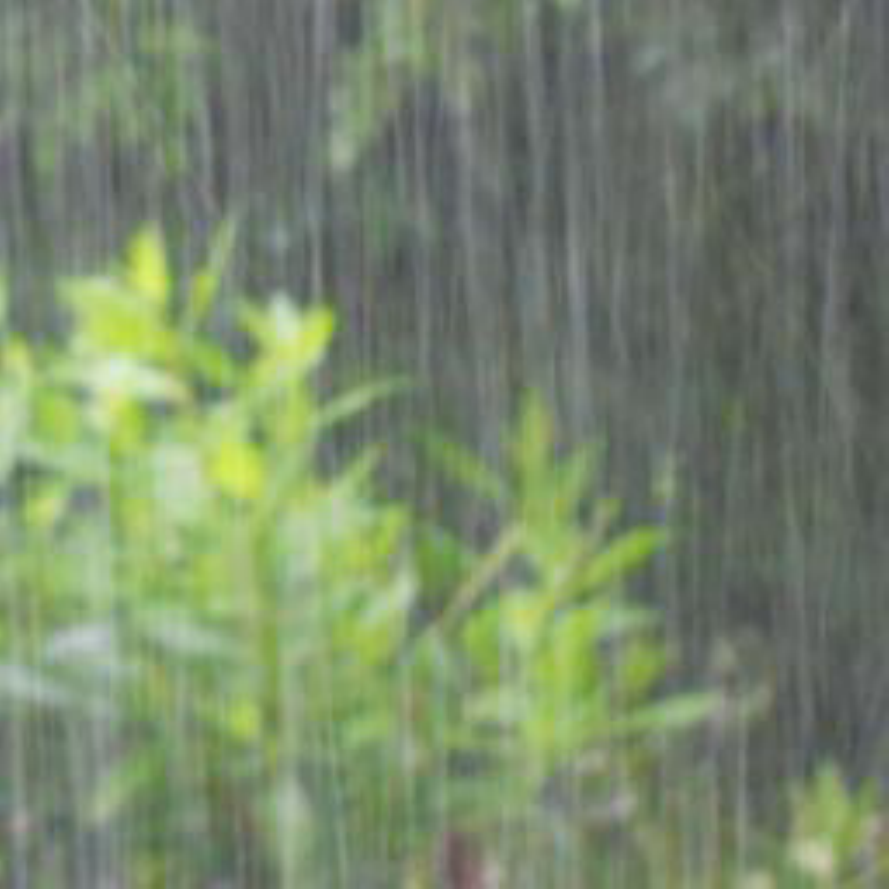}}
\centerline{(a)}
\end{minipage}
\hfill
\begin{minipage}{0.19\linewidth}
\centering{\includegraphics[width=1\linewidth]{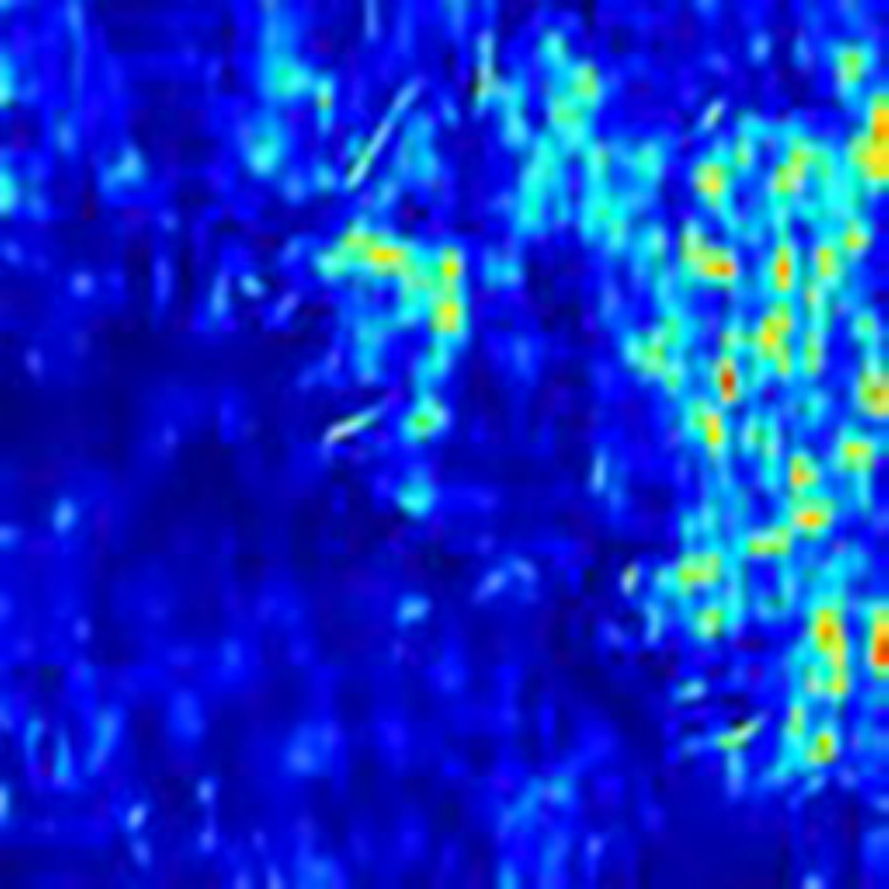}}
\centerline{(b)}
\end{minipage}
\hfill
\begin{minipage}{0.19\linewidth}
\centering{\includegraphics[width=1\linewidth]{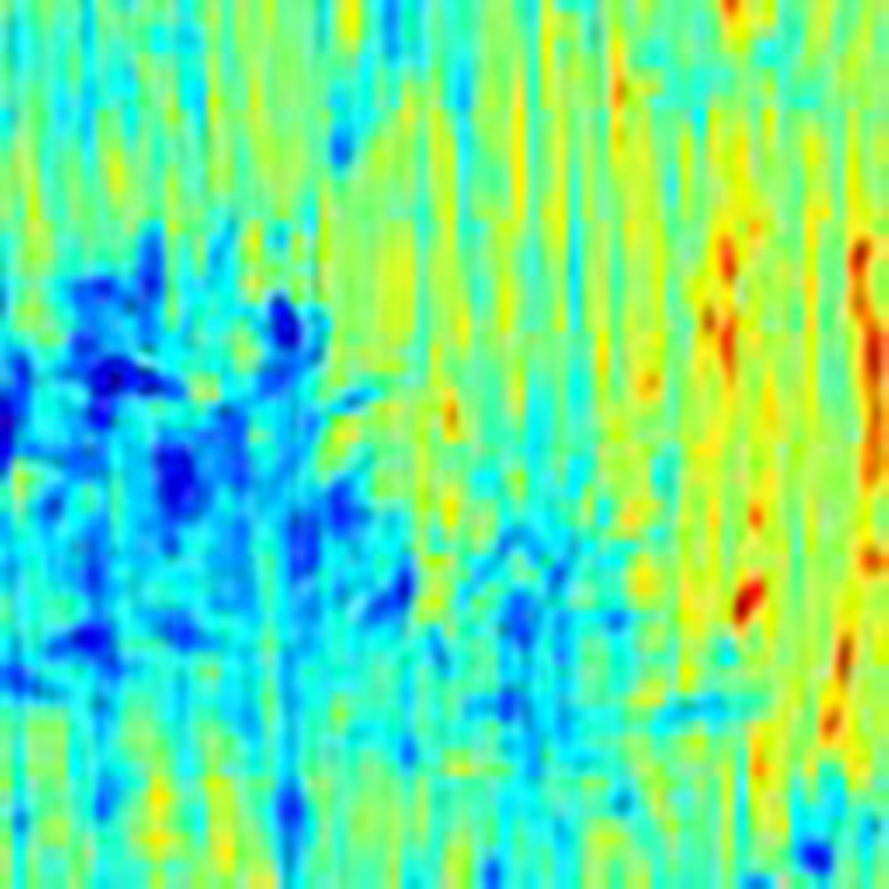}}
\centerline{(c)}
\end{minipage}
\hfill
\begin{minipage}{0.19\linewidth}
\centering{\includegraphics[width=1\linewidth]{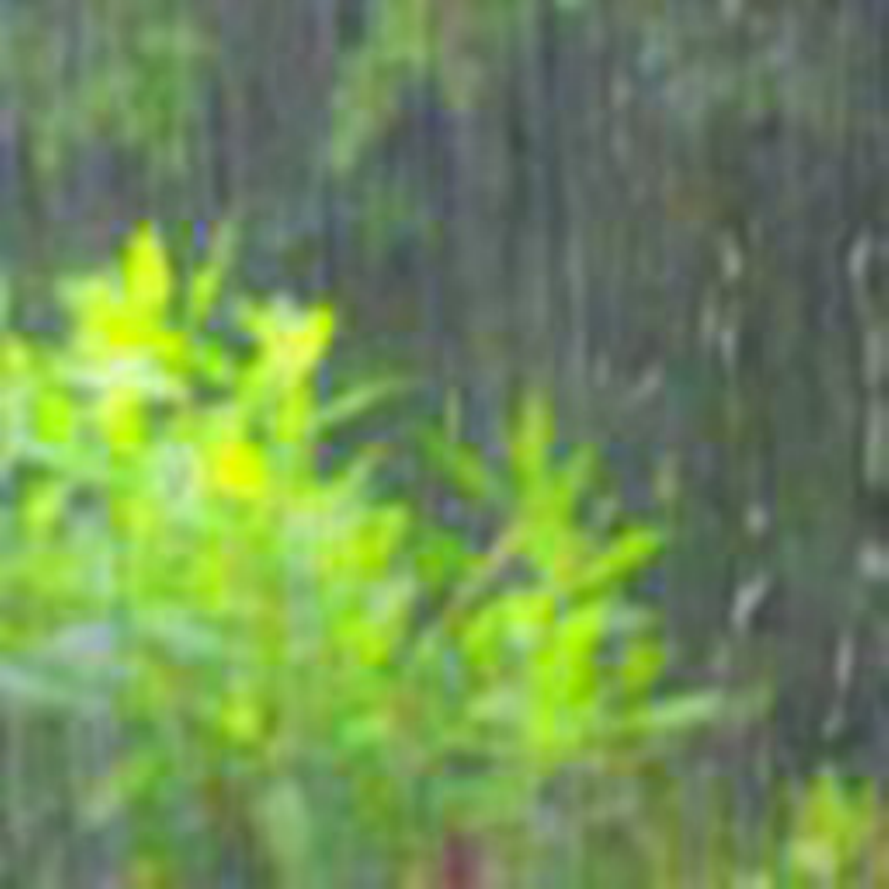}}
\centerline{(d)}
\end{minipage}
\hfill
\begin{minipage}{0.19\linewidth}
\centering{\includegraphics[width=1\linewidth]{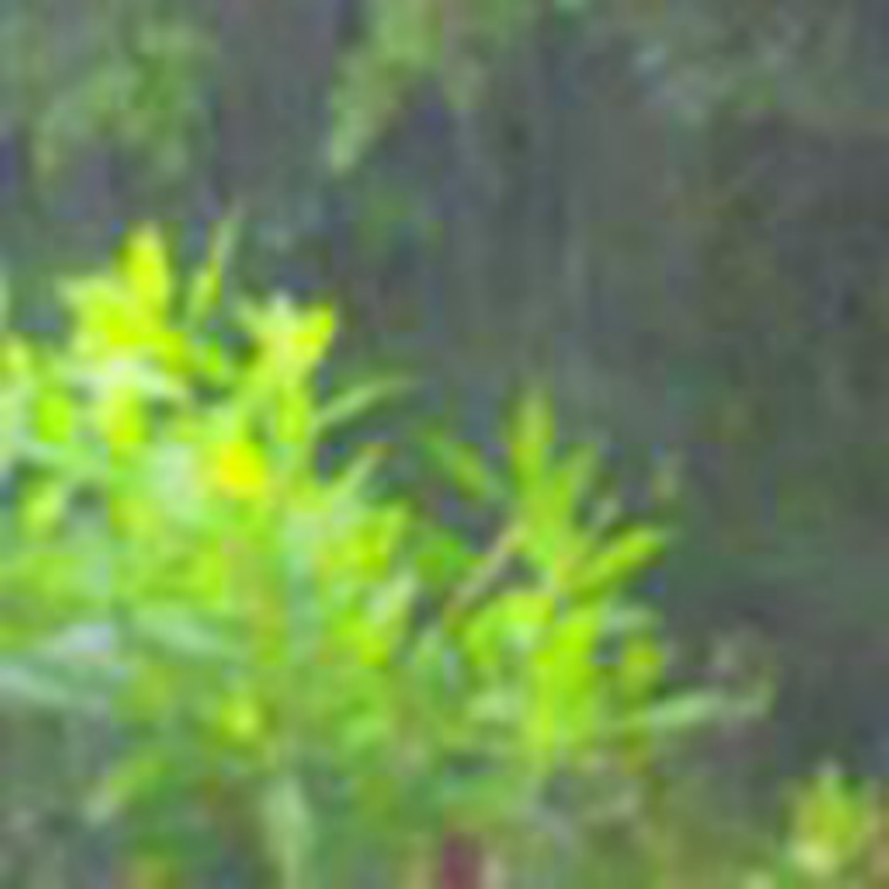}}
\centerline{(e)}
\end{minipage}
\end{center}
\caption{Visualization of the intermediate estimation. (a) Rainy images. (b) $\mathbf{1}-\mathbf{T}$. (c) $\mathbf{R}$. (d) $\mathbf{I}-\mathbf{R}$. (e) ($(\mathbf{I}-\mathbf{R}) \oslash \mathbf{T}$). $\mathbf{T}$ and $\mathbf{R}$ are normalized to $[0, 1]$ for visualization. The estimated $\mathbf{T}$ and $\bR$ can reflect the image areas degenerated by the haze-like mist and rain streaks. In (d), after removing the rain streaks reflected by $\mathbf{R}$, the image still suffers from the mist. We also observe that the operation with $\mathbf{T}$ can compensate the underestimate of $\mathbf{R}$. Please zoom in for better visualization.}
\label{fig:betas}
\end{figure}

\begin{figure*}[t]
\begin{center}
\begin{minipage}{0.078\linewidth}
\centering{\includegraphics[width=1\linewidth]{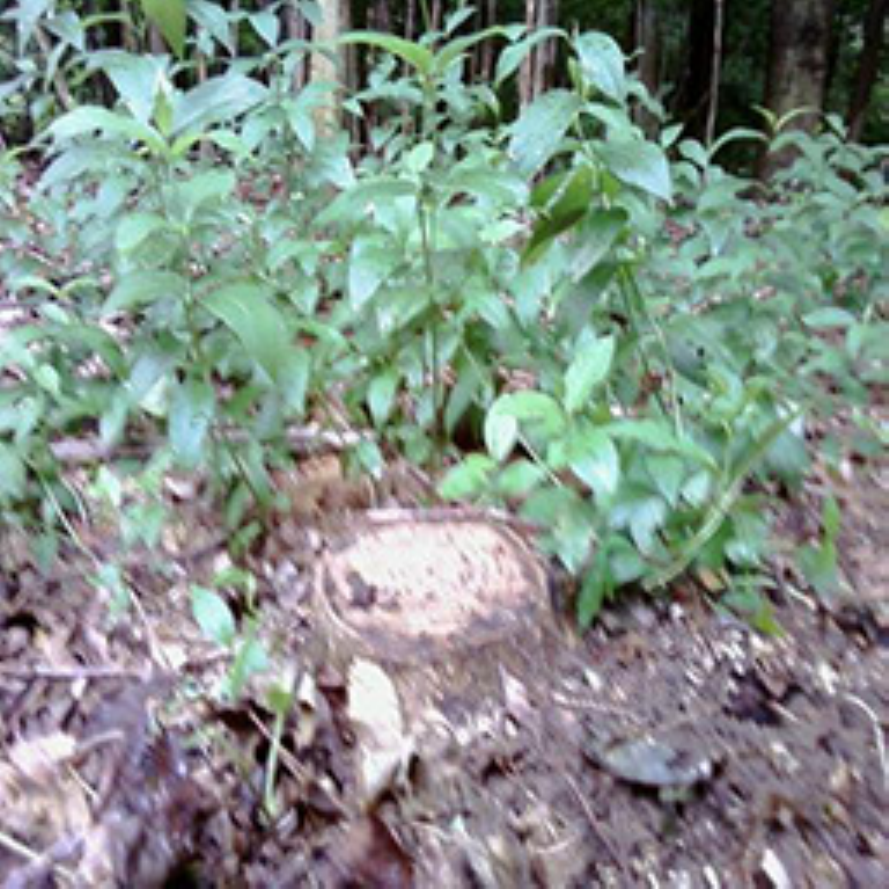}}
\end{minipage}
\hfill
\begin{minipage}{0.078\linewidth}
\centering{\includegraphics[width=1\linewidth]{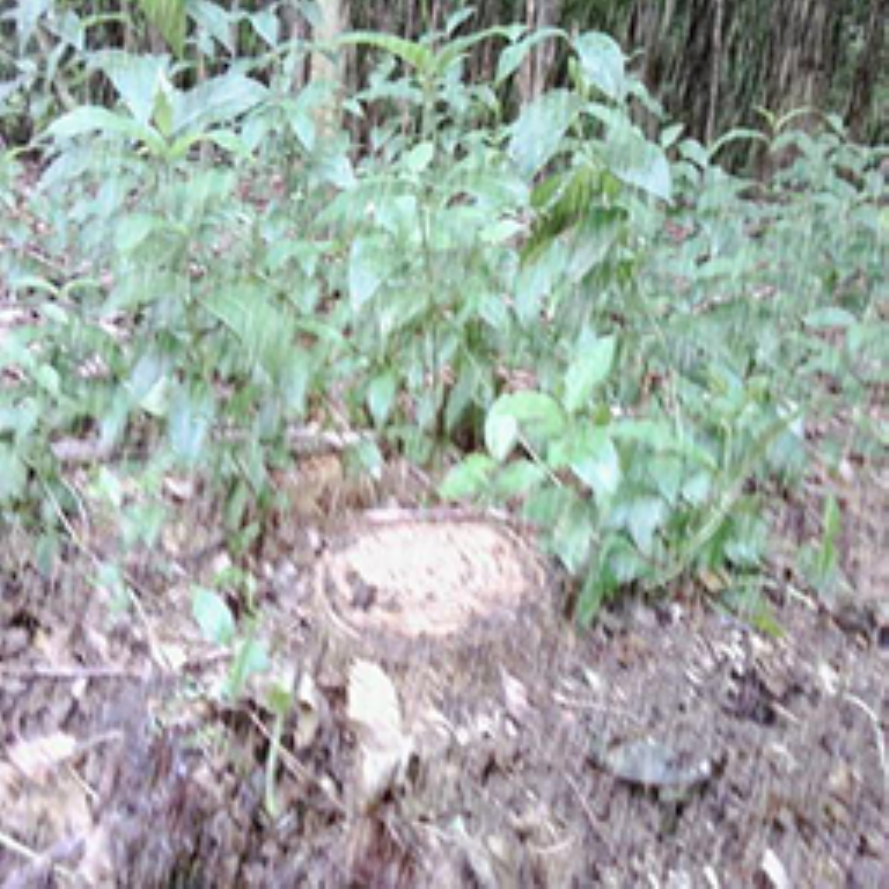}}
\end{minipage}
\hfill
\begin{minipage}{0.078\linewidth}
\centering{\includegraphics[width=1\linewidth]{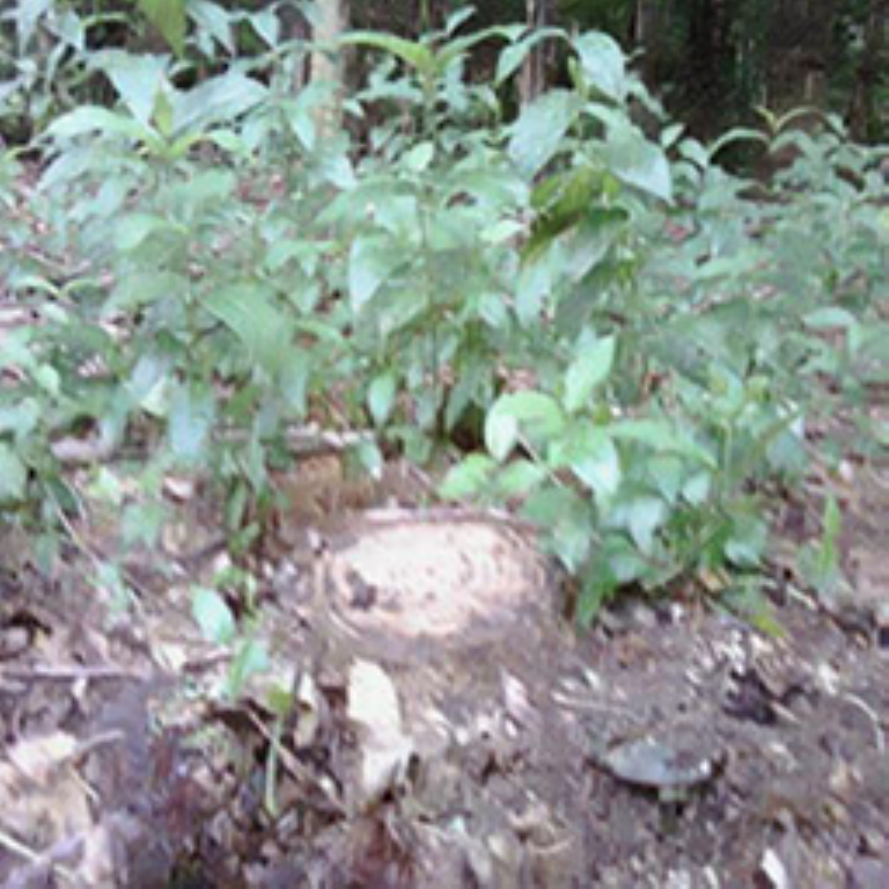}}
\end{minipage}
\hfill
\begin{minipage}{0.078\linewidth}
\centering{\includegraphics[width=1\linewidth]{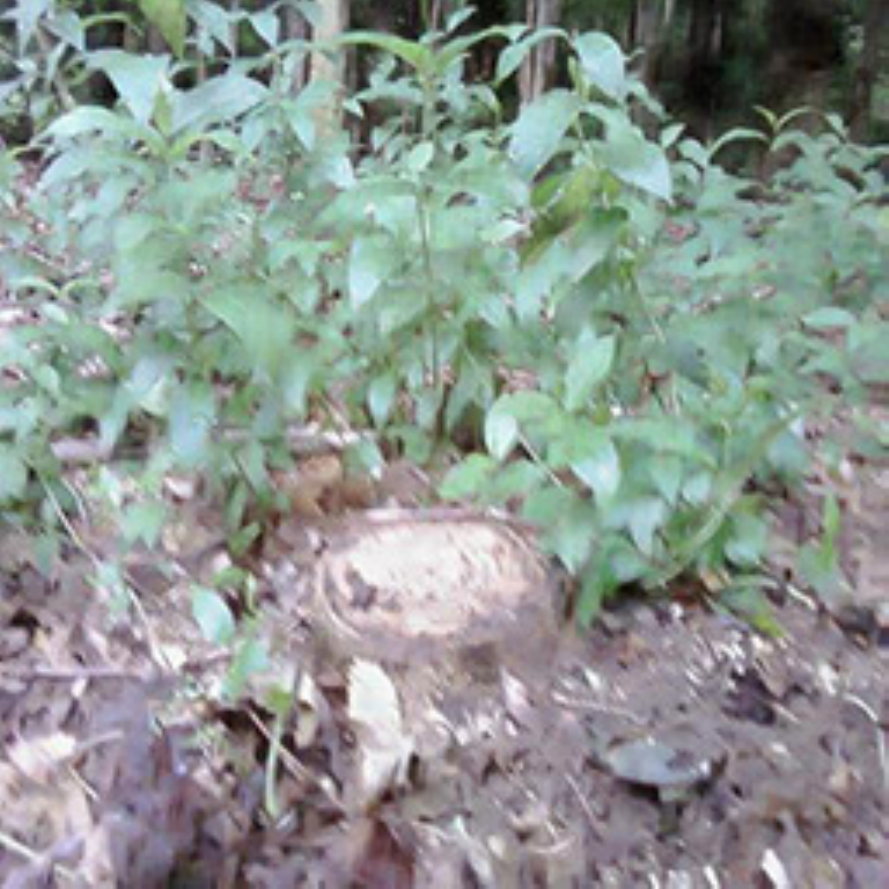}}
\end{minipage}
\hfill
\begin{minipage}{0.078\linewidth}
\centering{\includegraphics[width=1\linewidth]{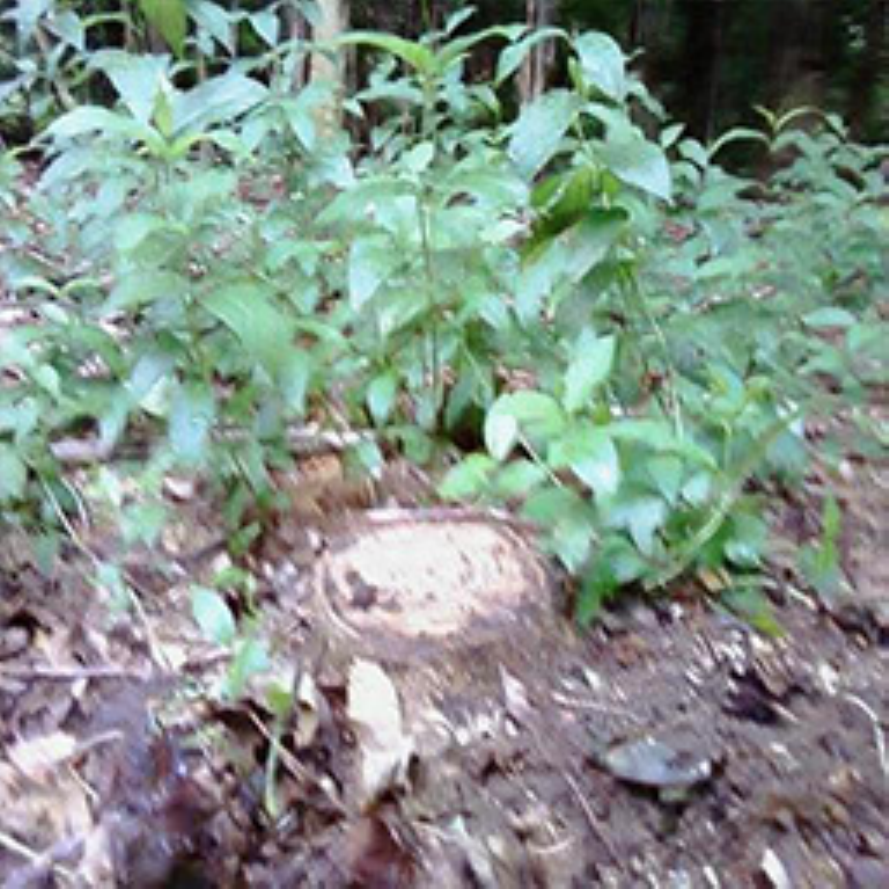}}
\end{minipage}
\hfill
\begin{minipage}{0.078\linewidth}
\centering{\includegraphics[width=1\linewidth]{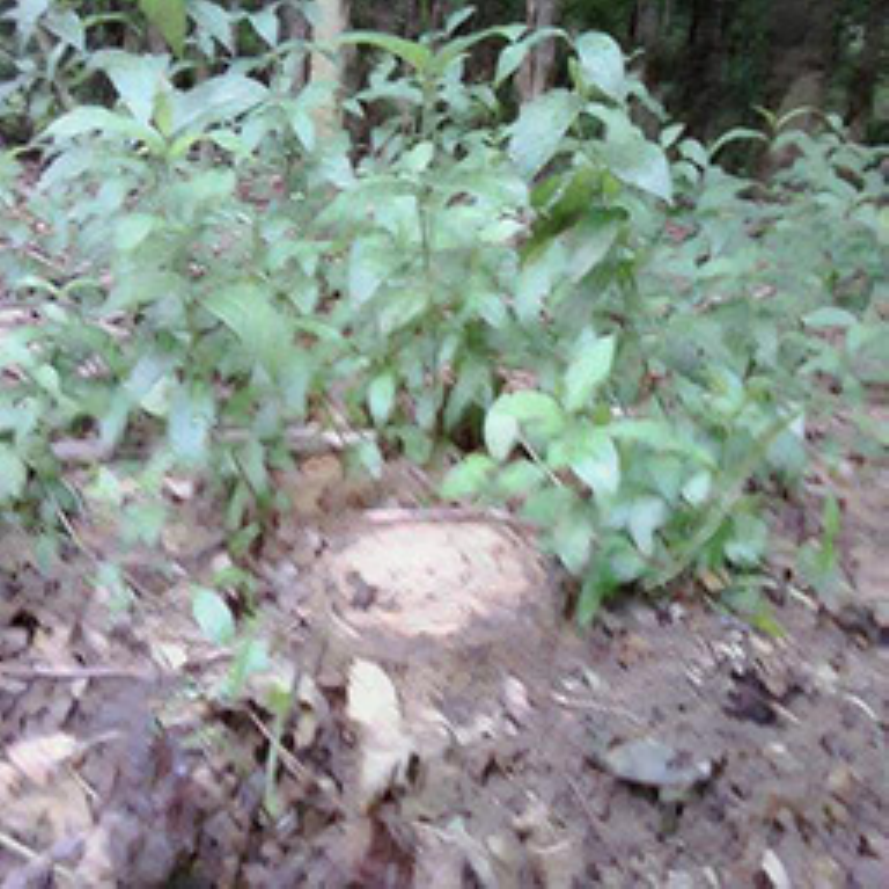}}
\end{minipage}
\hfill
\begin{minipage}{0.078\linewidth}
\centering{\includegraphics[width=1\linewidth]{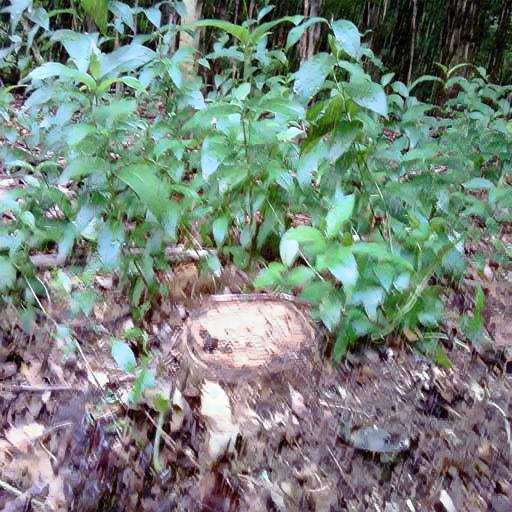}}
\end{minipage}
\hfill
\begin{minipage}{0.078\linewidth}
\centering{\includegraphics[width=1\linewidth]{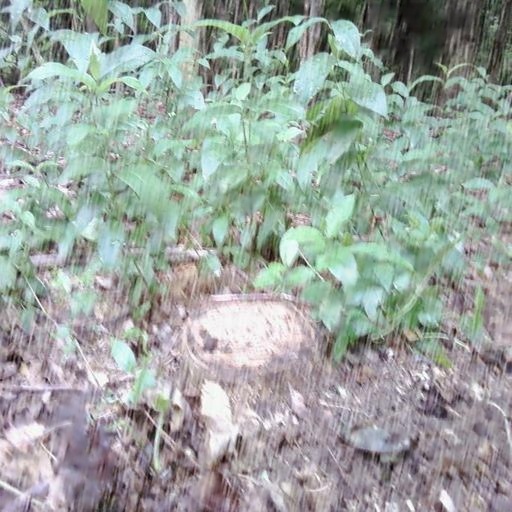}}
\end{minipage}
\hfill
\begin{minipage}{0.078\linewidth}
\centering{\includegraphics[width=1\linewidth]{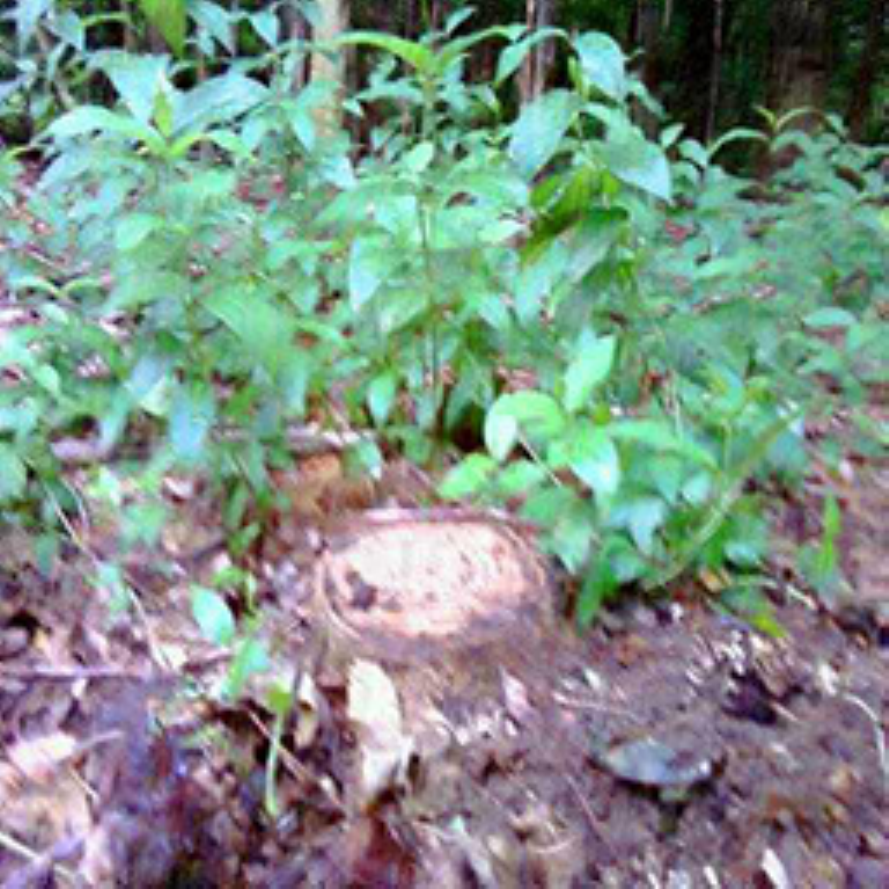}}
\end{minipage}
\hfill
\begin{minipage}{0.078\linewidth}
\centering{\includegraphics[width=1\linewidth]{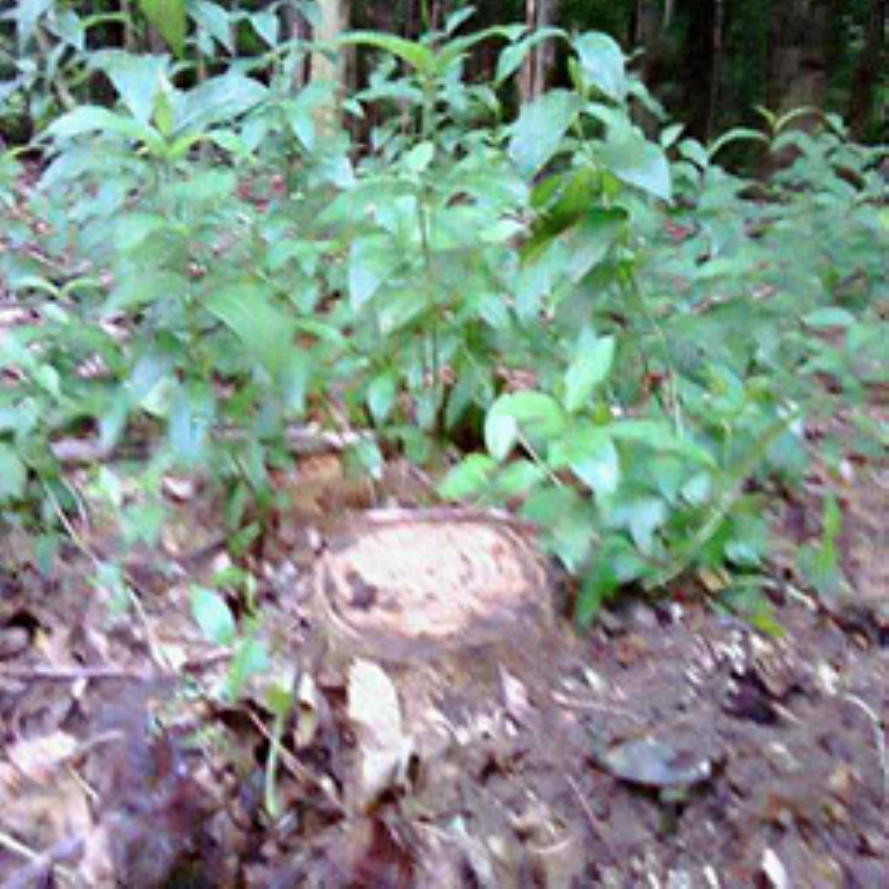}}
\end{minipage}
\hfill
\begin{minipage}{0.078\linewidth}
\centering{\includegraphics[width=1\linewidth]{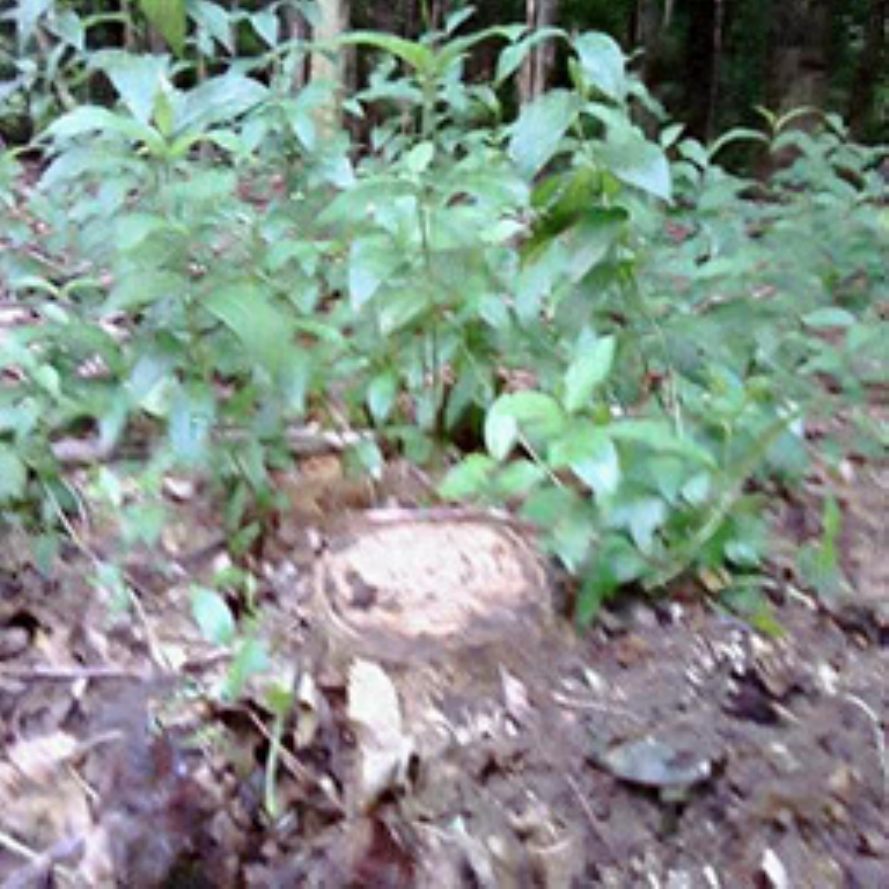}}
\end{minipage}
\hfill
\begin{minipage}{0.078\linewidth}
\centering{\includegraphics[width=1\linewidth]{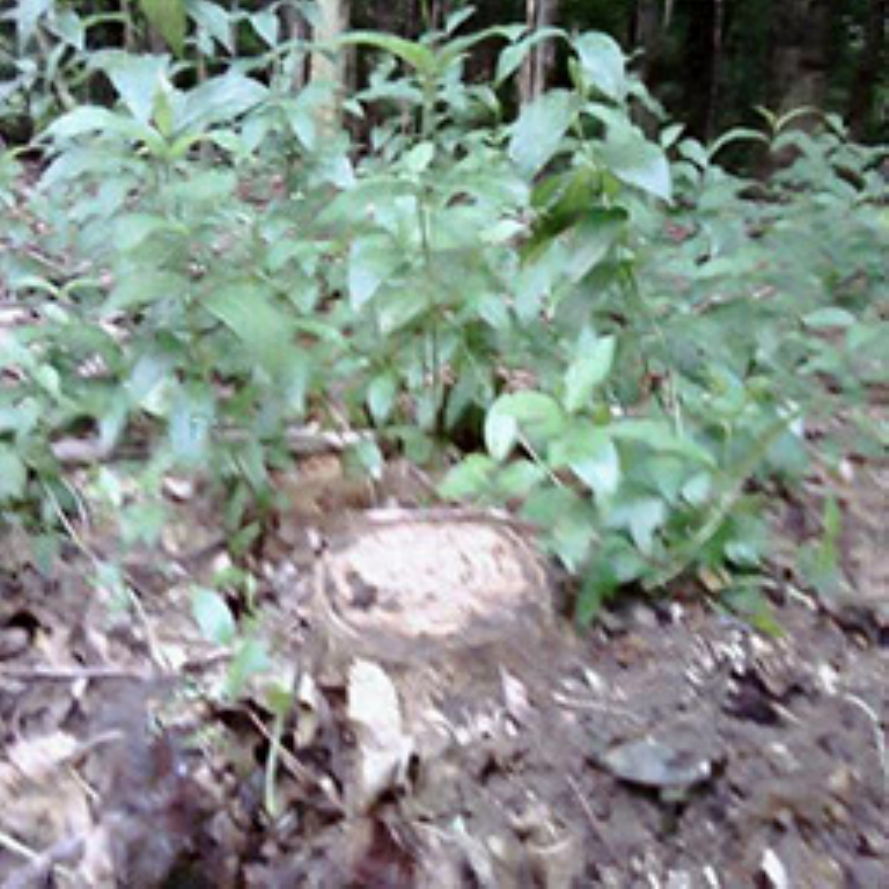}}
\end{minipage}
\vfill
\begin{minipage}{0.078\linewidth}
\centering{\includegraphics[width=1\linewidth]{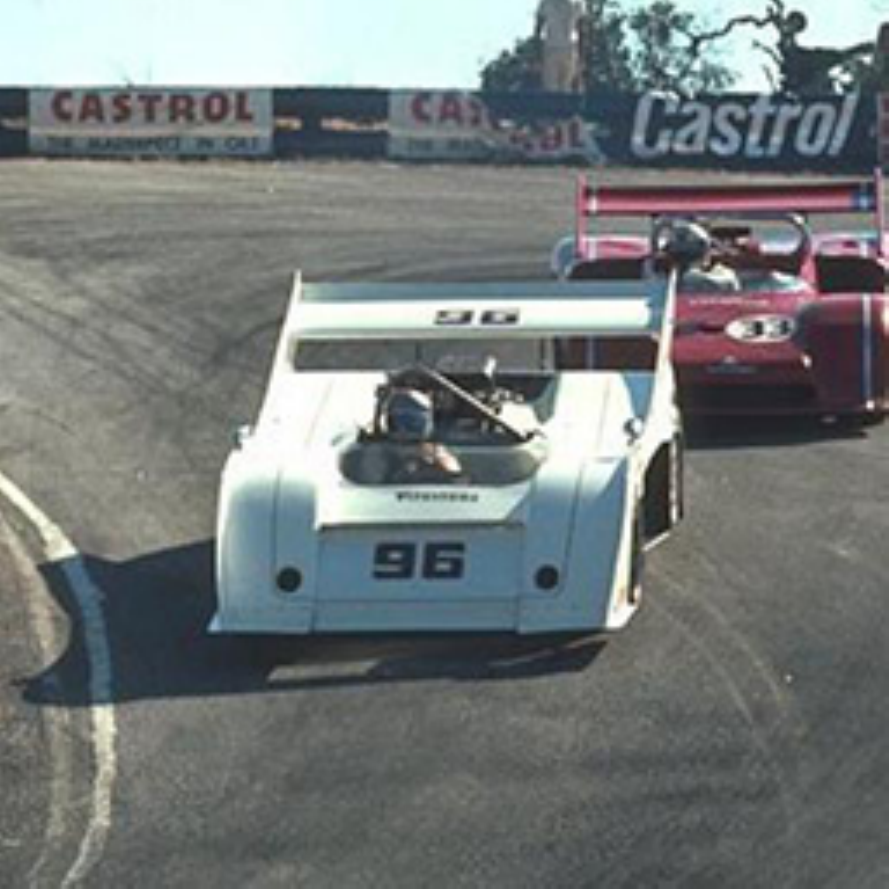}}
\centerline{(a)}
\end{minipage}
\hfill
\begin{minipage}{0.078\linewidth}
\centering{\includegraphics[width=1\linewidth]{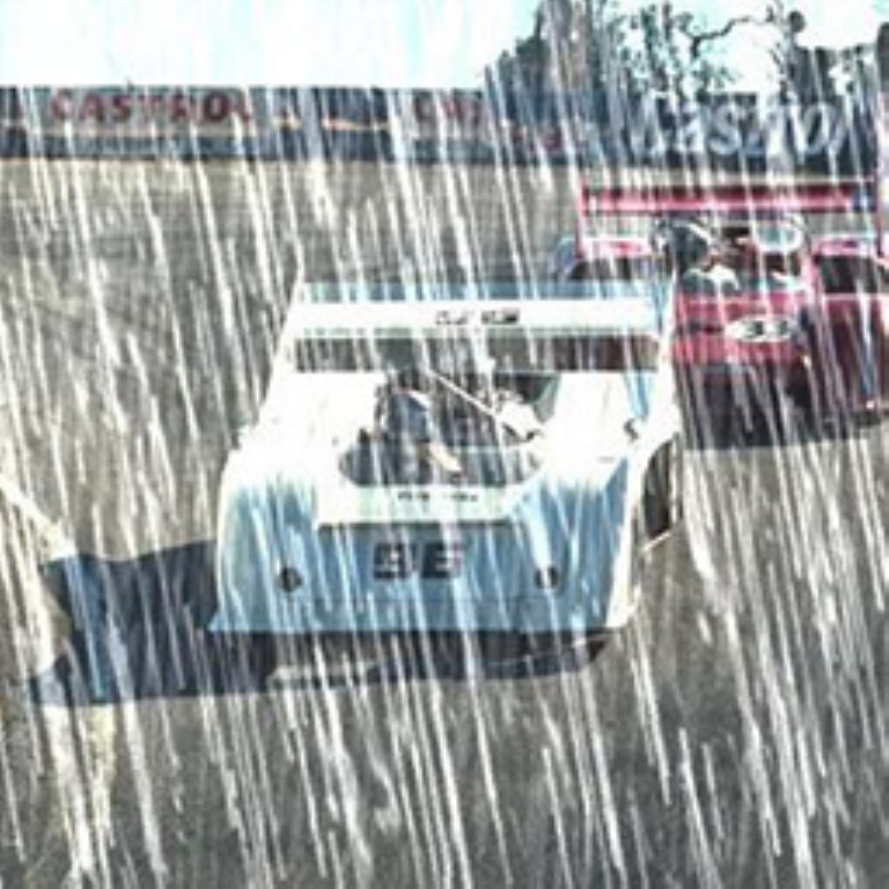}}
\centerline{(b)}
\end{minipage}
\hfill
\begin{minipage}{0.078\linewidth}
\centering{\includegraphics[width=1\linewidth]{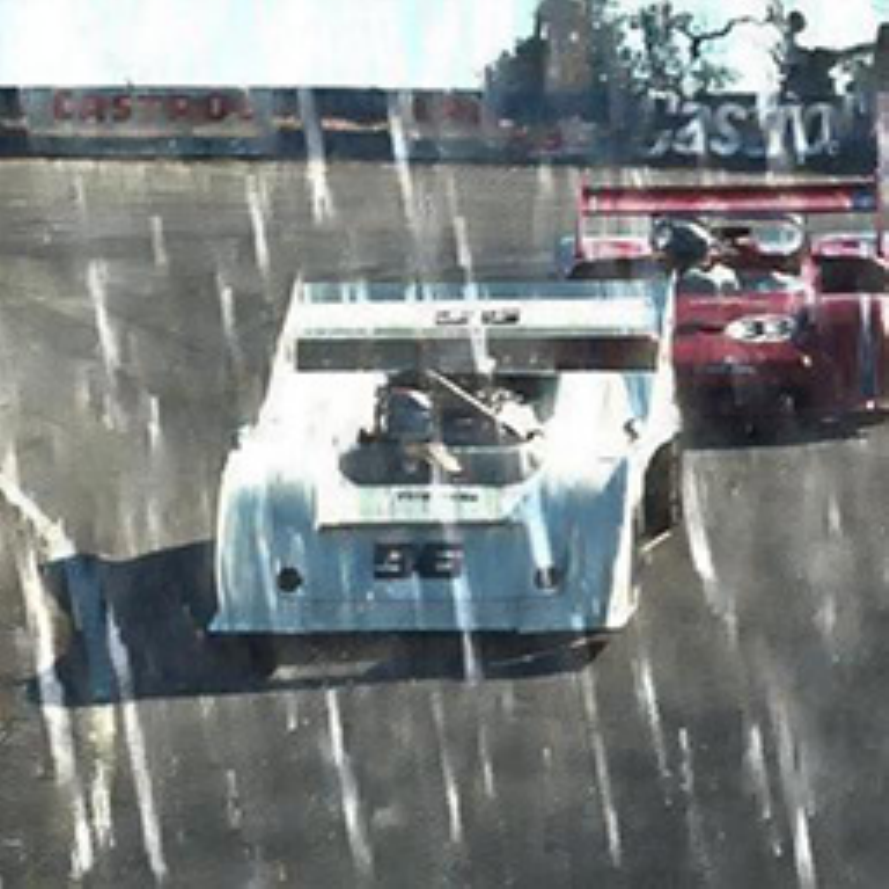}}
\centerline{(c)}
\end{minipage}
\hfill
\begin{minipage}{0.078\linewidth}
\centering{\includegraphics[width=1\linewidth]{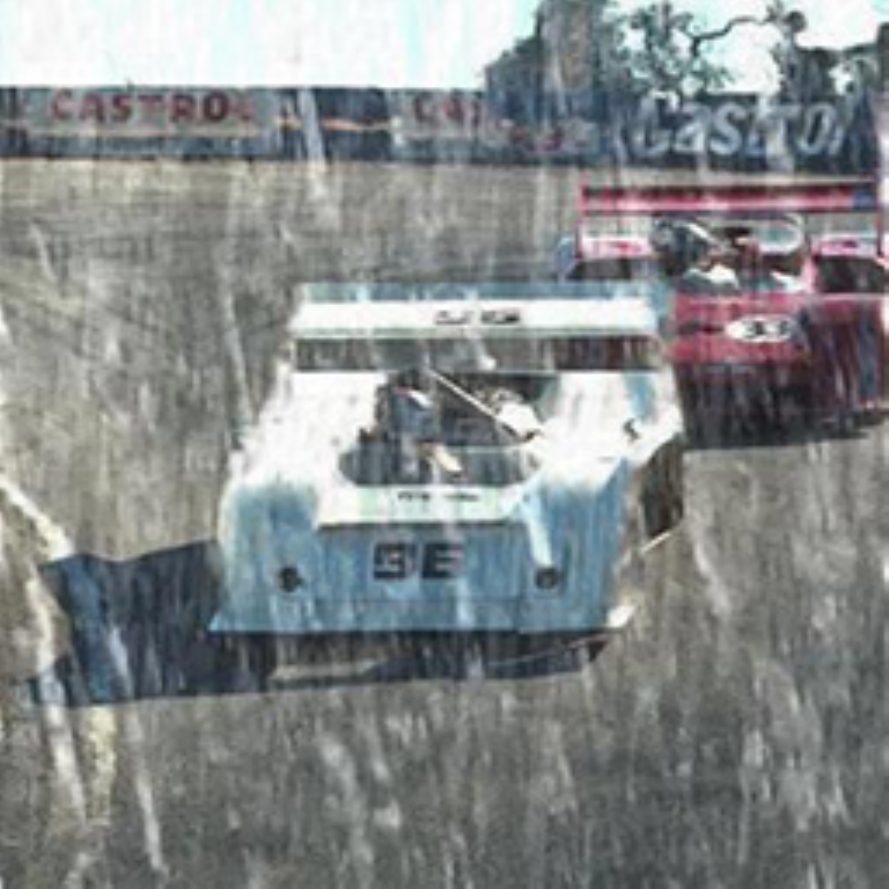}}
\centerline{(d)}
\end{minipage}
\hfill
\begin{minipage}{0.078\linewidth}
\centering{\includegraphics[width=1\linewidth]{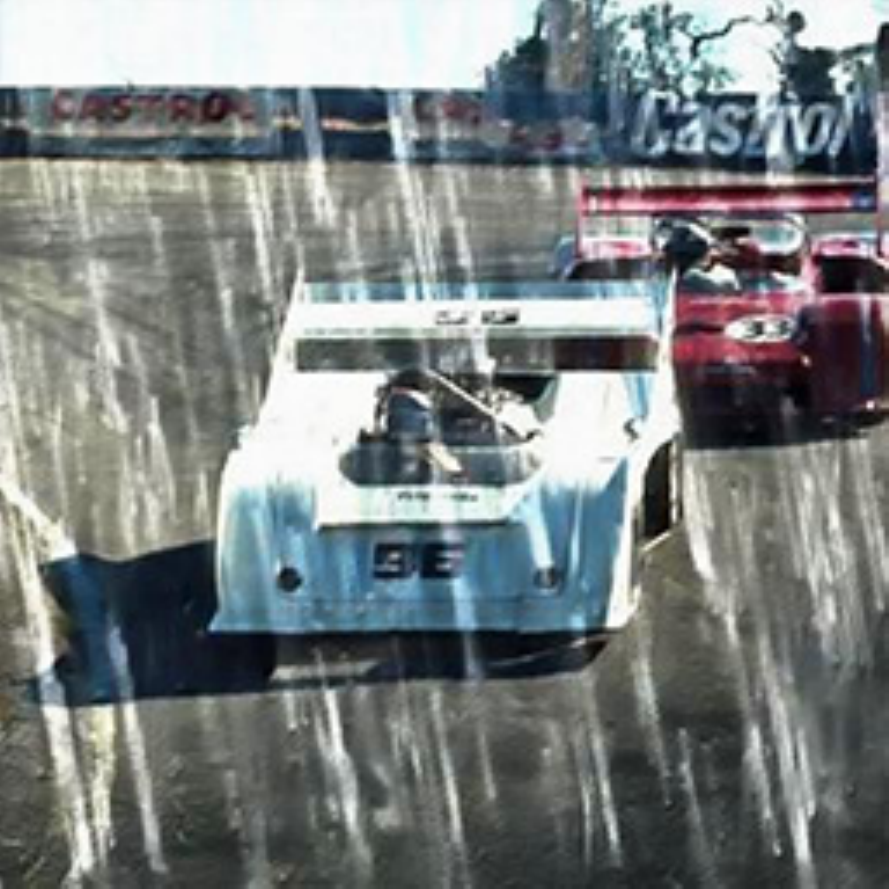}}
\centerline{(e)}
\end{minipage}
\hfill
\begin{minipage}{0.078\linewidth}
\centering{\includegraphics[width=1\linewidth]{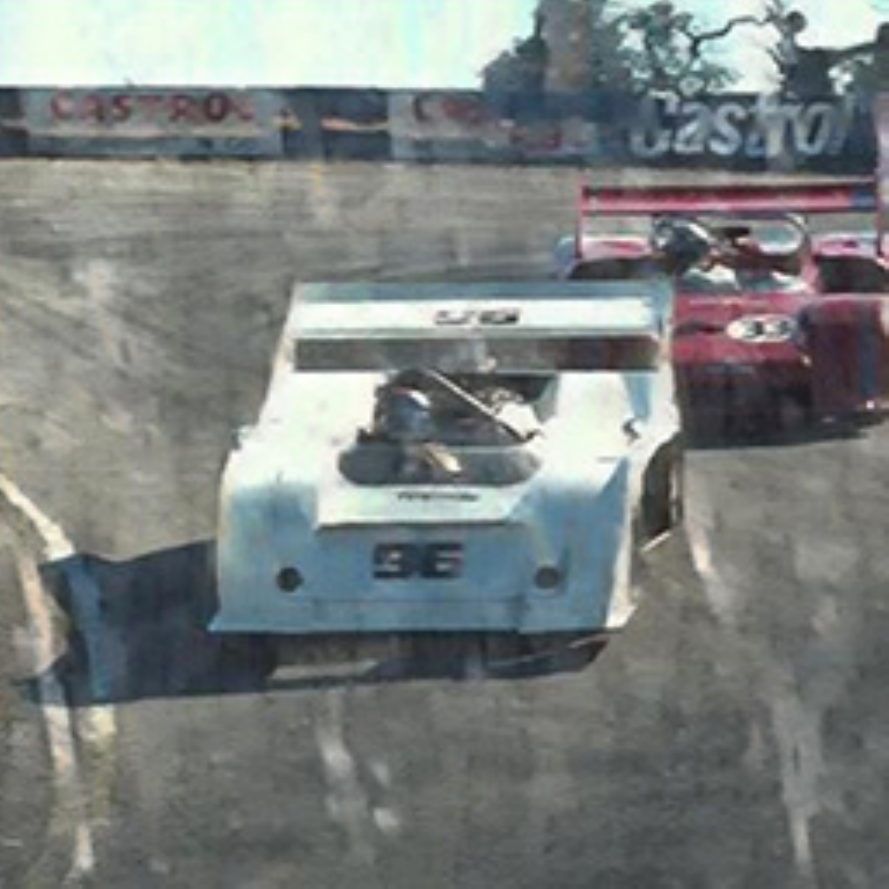}}
\centerline{(f)}
\end{minipage}
\hfill
\begin{minipage}{0.078\linewidth}
\centering{\includegraphics[width=1\linewidth]{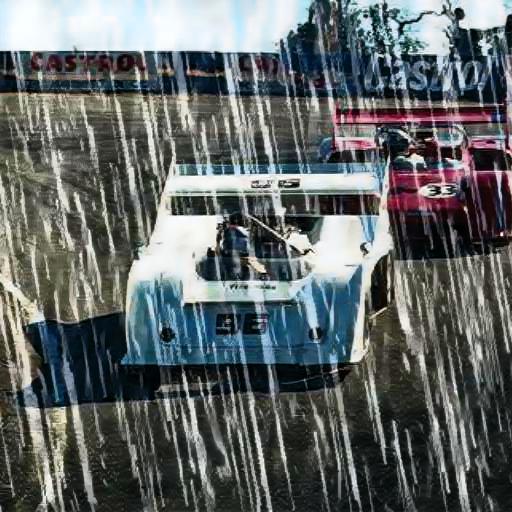}}
\centerline{(g)}
\end{minipage}
\hfill
\begin{minipage}{0.078\linewidth}
\centering{\includegraphics[width=1\linewidth]{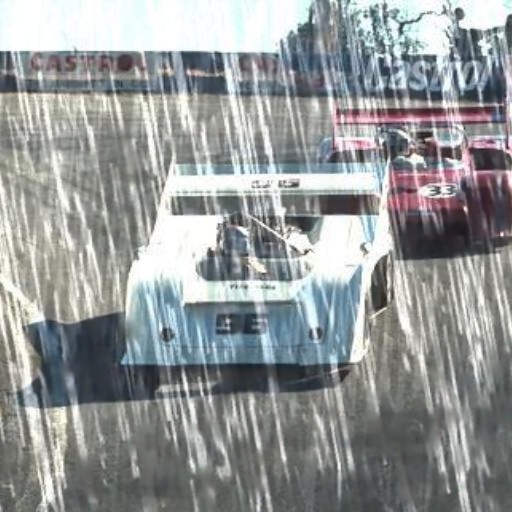}}
\centerline{(h)}
\end{minipage}
\hfill
\begin{minipage}{0.078\linewidth}
\centering{\includegraphics[width=1\linewidth]{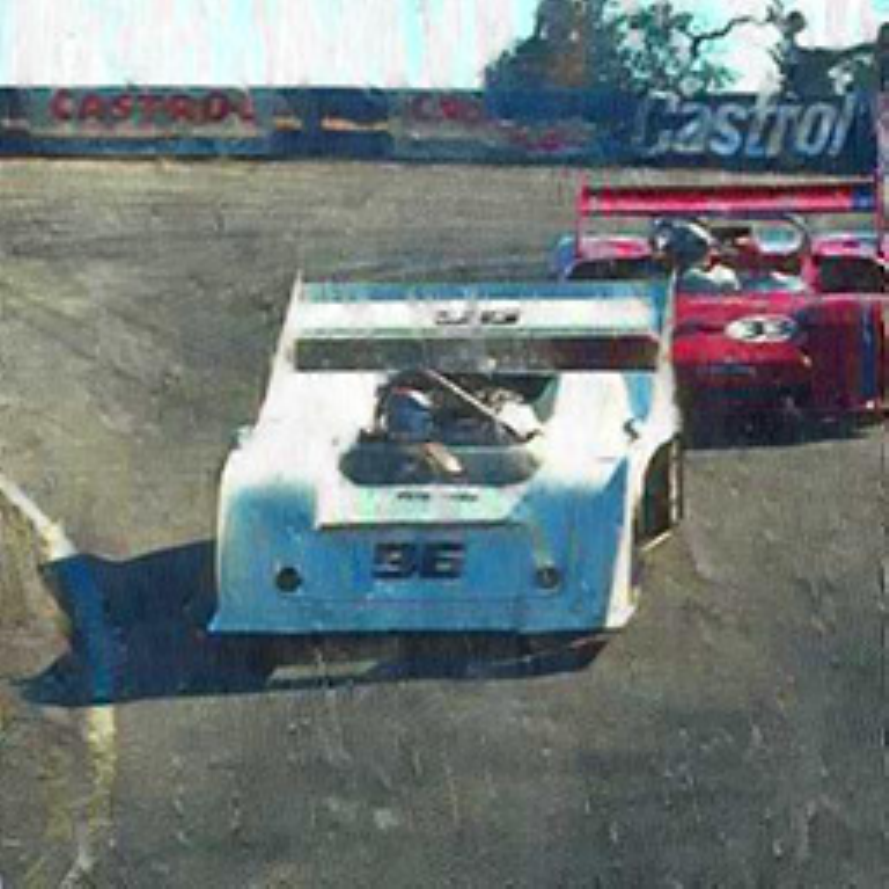}}
\centerline{(i)}
\end{minipage}
\hfill
\begin{minipage}{0.078\linewidth}
\centering{\includegraphics[width=1\linewidth]{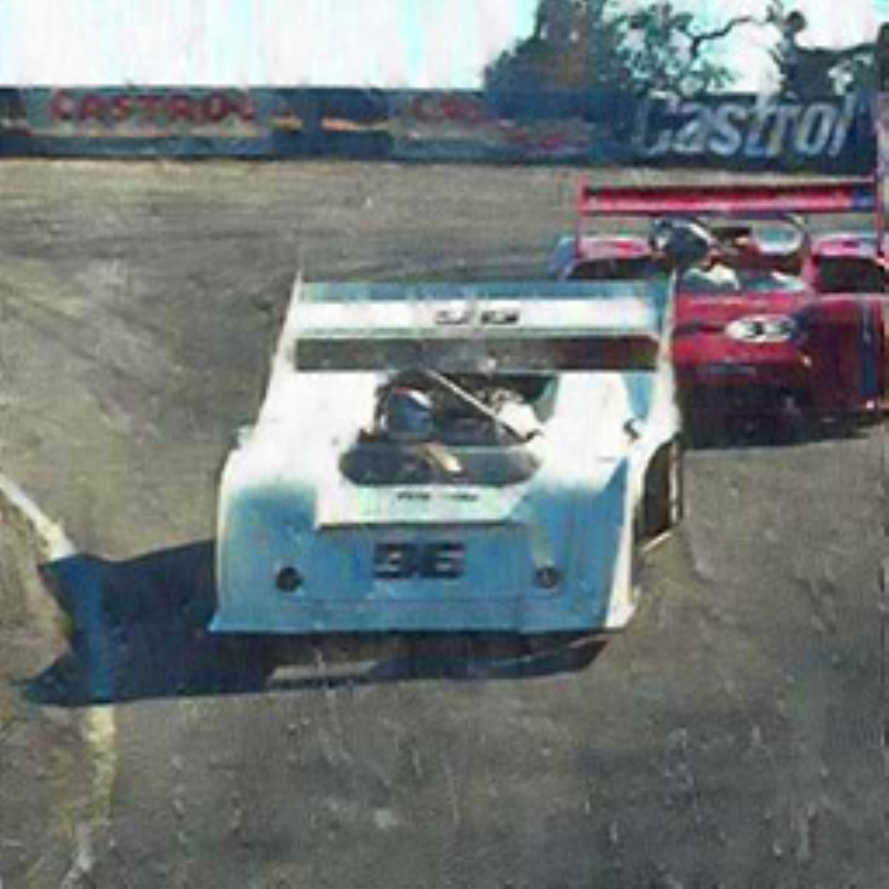}}
\centerline{(j)}
\end{minipage}
\hfill
\begin{minipage}{0.078\linewidth}
\centering{\includegraphics[width=1\linewidth]{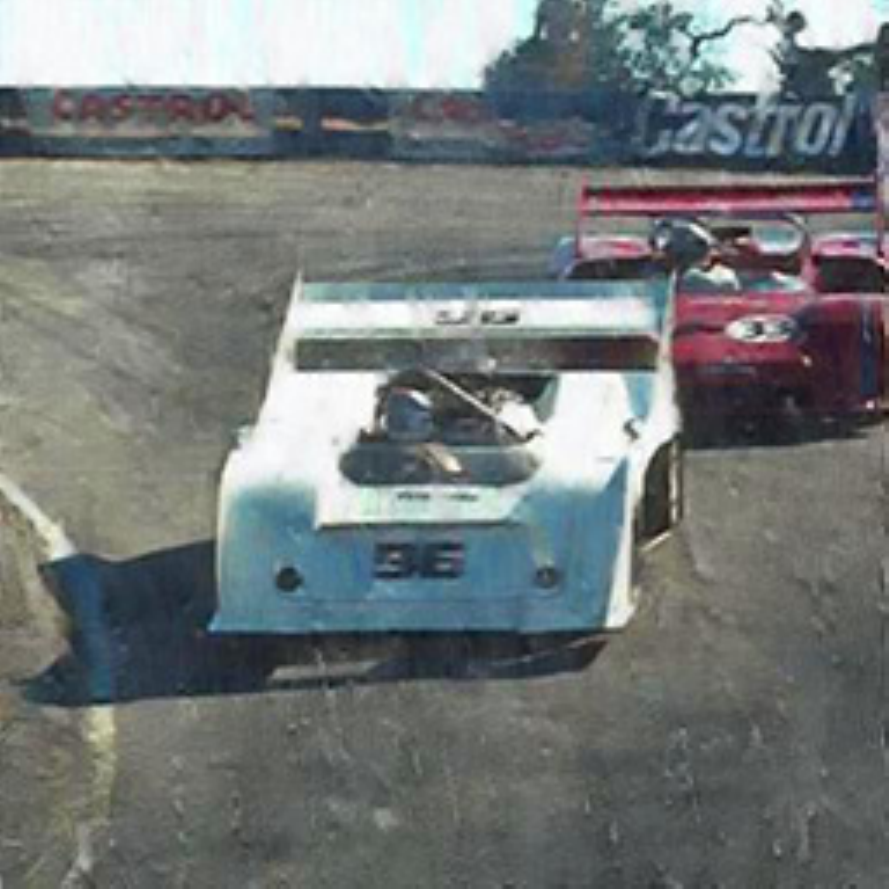}}
\centerline{(k)}
\end{minipage}
\hfill
\begin{minipage}{0.078\linewidth}
\centering{\includegraphics[width=1\linewidth]{images/ysrain-33_lambda_0_3_Derain.pdf}}
\centerline{(l)}
\end{minipage}
\end{center}
\caption{Rain removal results for synthetic rainy images. (a) Ground truth. (b) Input rainy image. (c)-(h) The results of \cite{Fu_2017_CVPR,Yang_2017_CVPR,Zhang_2018_CVPR,Li_2018_ECCV,Li_rt_2019_CVPR,Wang_ty_2019_CVPR}. (i)-(l) Our results with $\alpha$ equaling to $1.0$, $0.6$, $0.3$, $0.0$. Please zoom in to see clearly.}
\label{fig:synthetic_1}
\end{figure*}

\begin{figure*}[t!]
\begin{center}
\begin{minipage}{0.085\linewidth}
\centering{\includegraphics[width=1\linewidth]{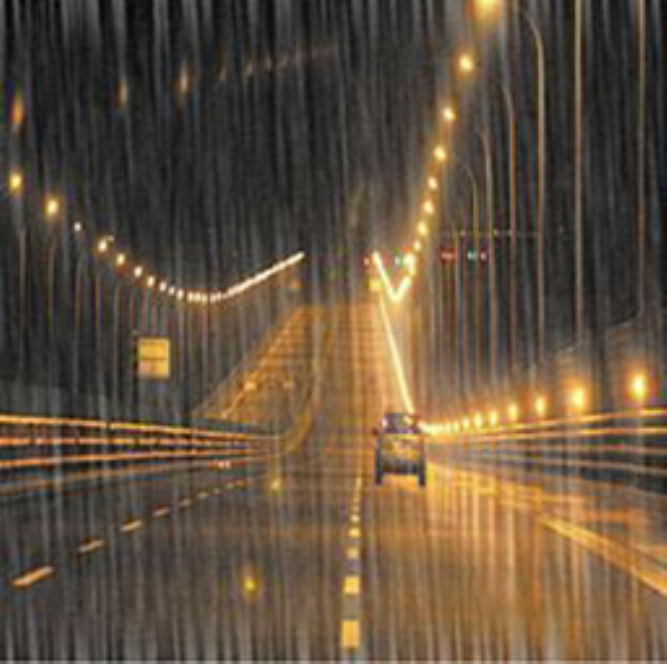}}
\end{minipage}
\hfill
\begin{minipage}{0.085\linewidth}
\centering{\includegraphics[width=1\linewidth]{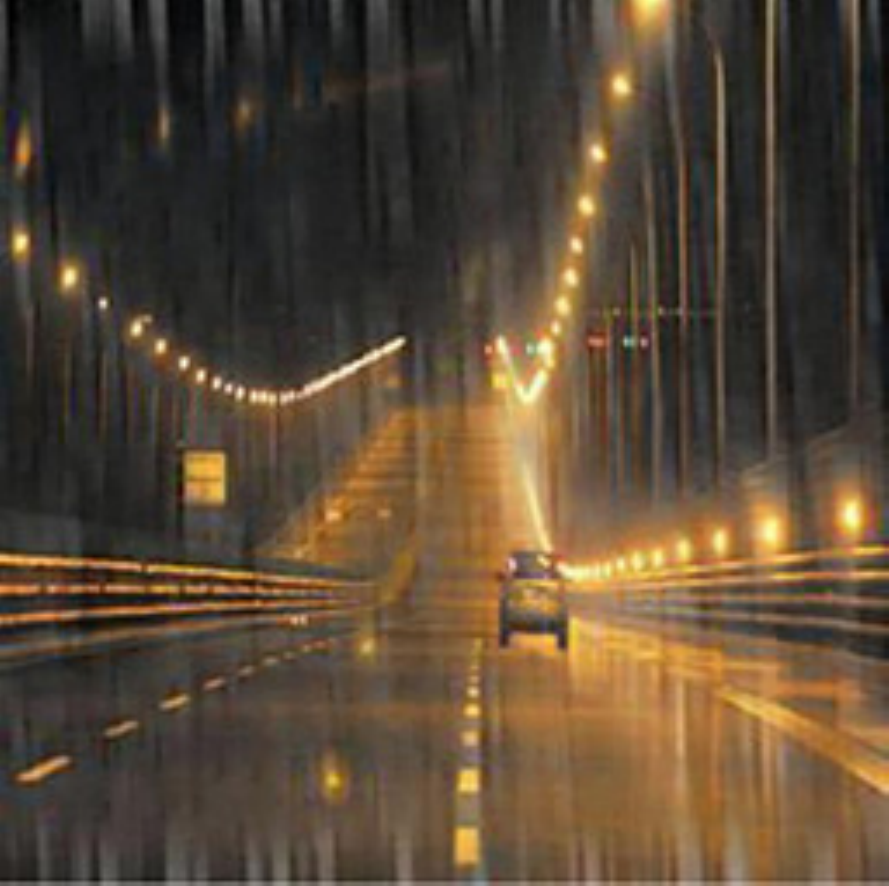}}
\end{minipage}
\hfill
\begin{minipage}{0.085\linewidth}
\centering{\includegraphics[width=1\linewidth]{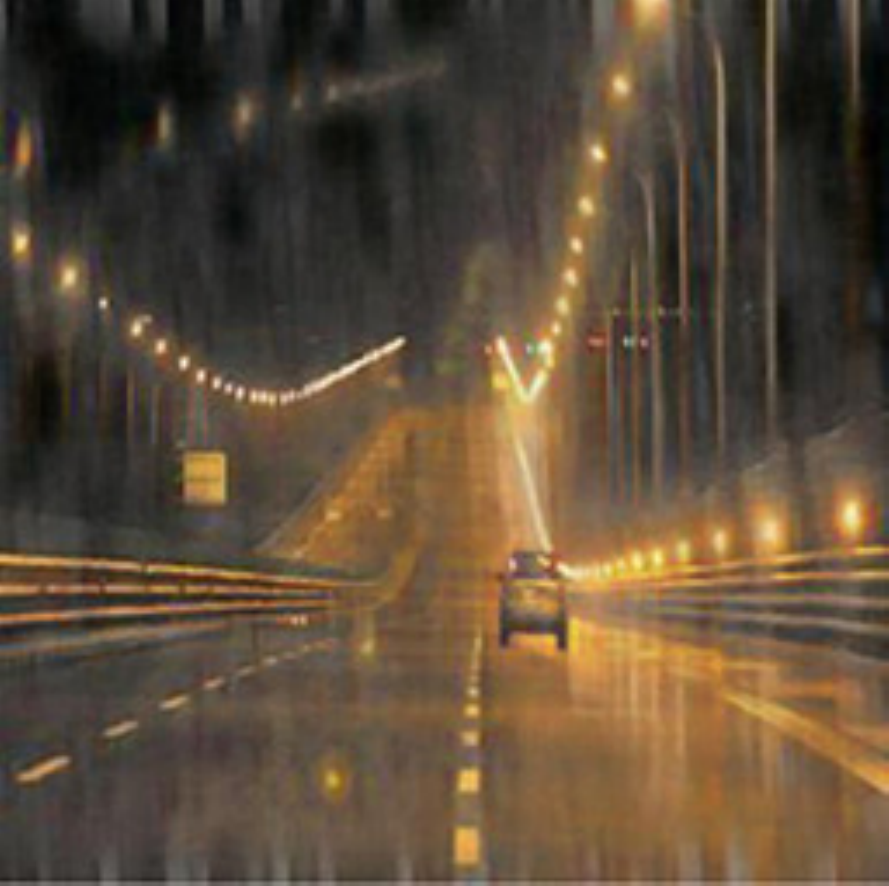}}
\end{minipage}
\hfill
\begin{minipage}{0.085\linewidth}
\centering{\includegraphics[width=1\linewidth]{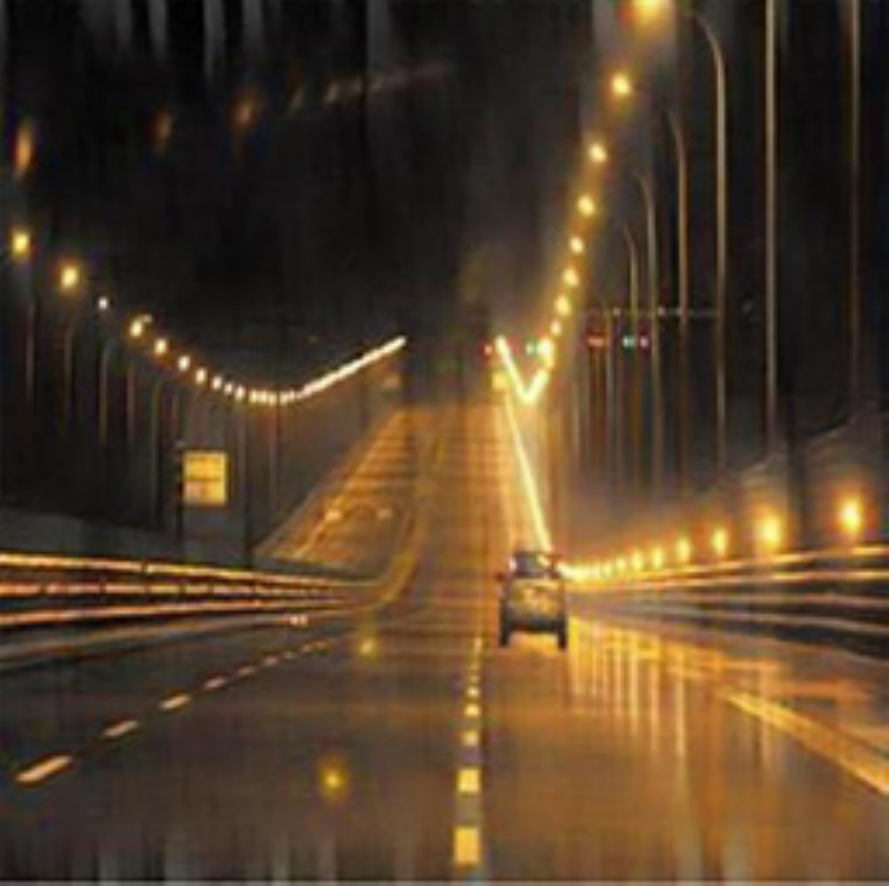}}
\end{minipage}
\hfill
\begin{minipage}{0.085\linewidth}
\centering{\includegraphics[width=1\linewidth]{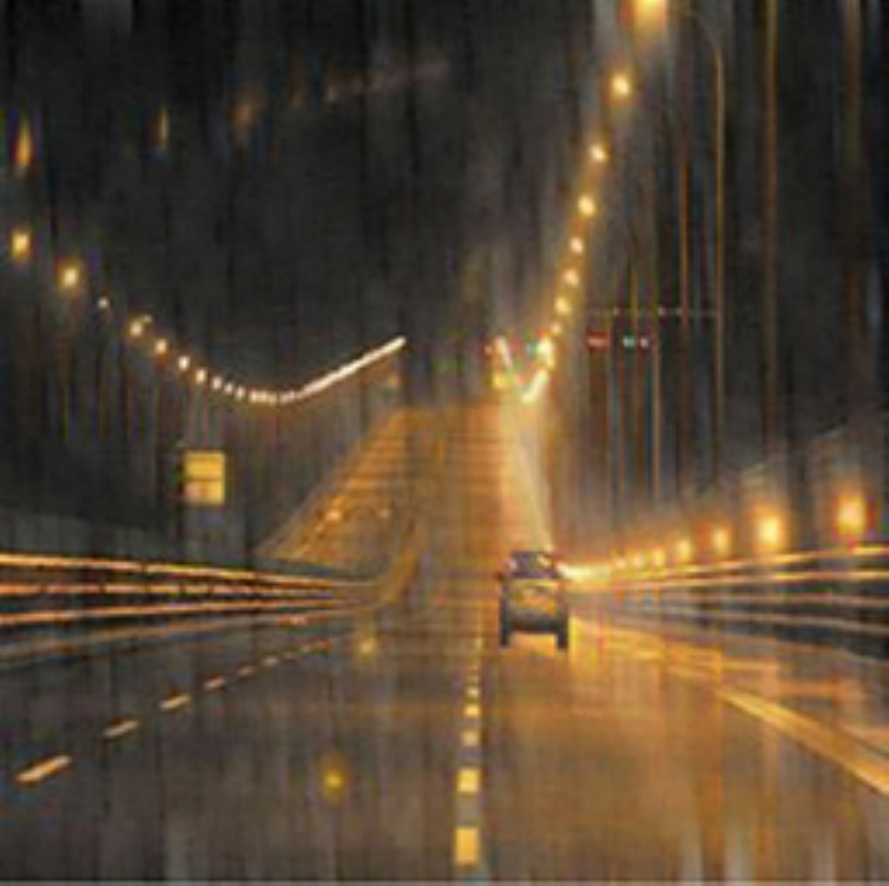}}
\end{minipage}
\hfill
\begin{minipage}{0.085\linewidth}
\centering{\includegraphics[width=1\linewidth]{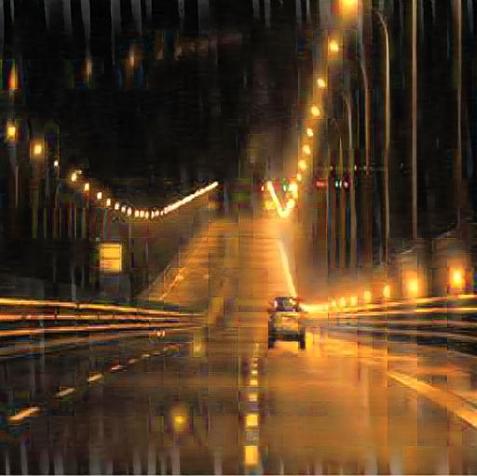}}
\end{minipage}
\hfill
\begin{minipage}{0.085\linewidth}
\centering{\includegraphics[width=1\linewidth]{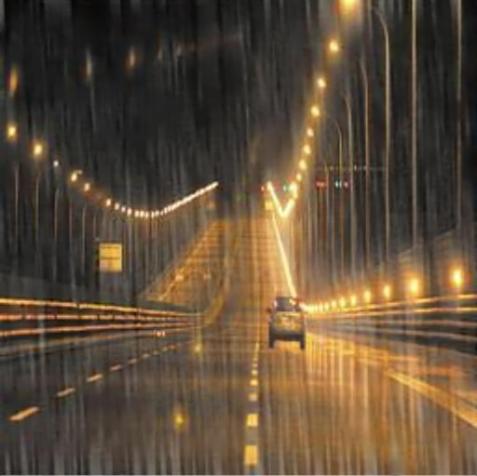}}
\end{minipage}
\hfill
\begin{minipage}{0.085\linewidth}
\centering{\includegraphics[width=1\linewidth]{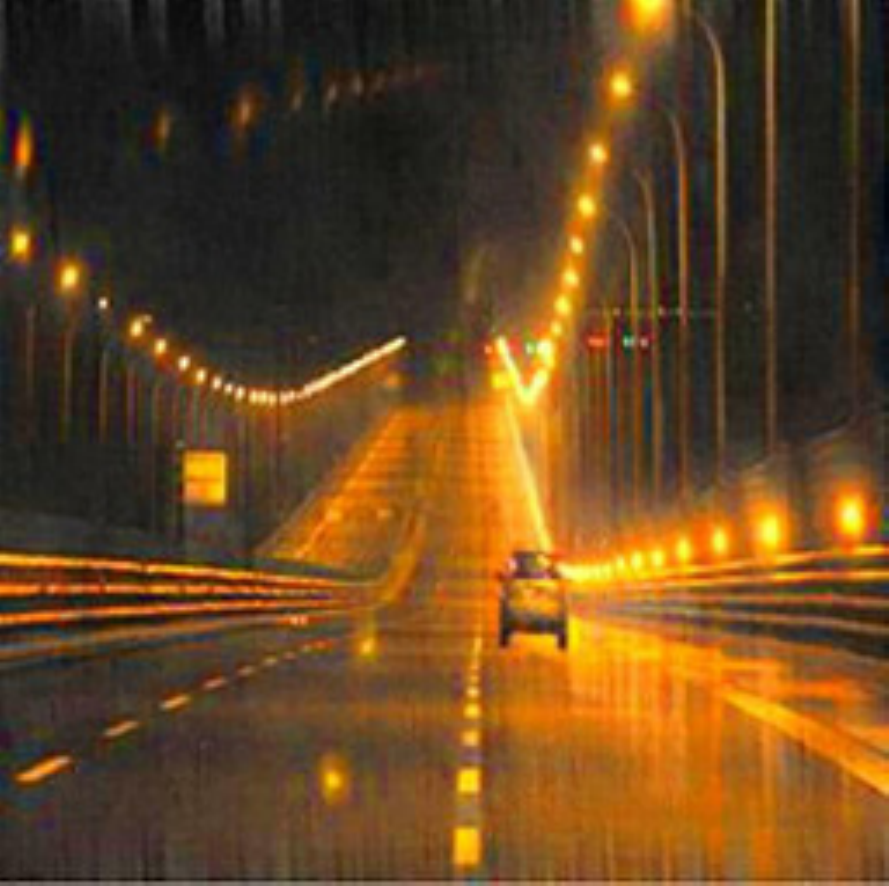}}
\end{minipage}
\hfill
\begin{minipage}{0.085\linewidth}
\centering{\includegraphics[width=1\linewidth]{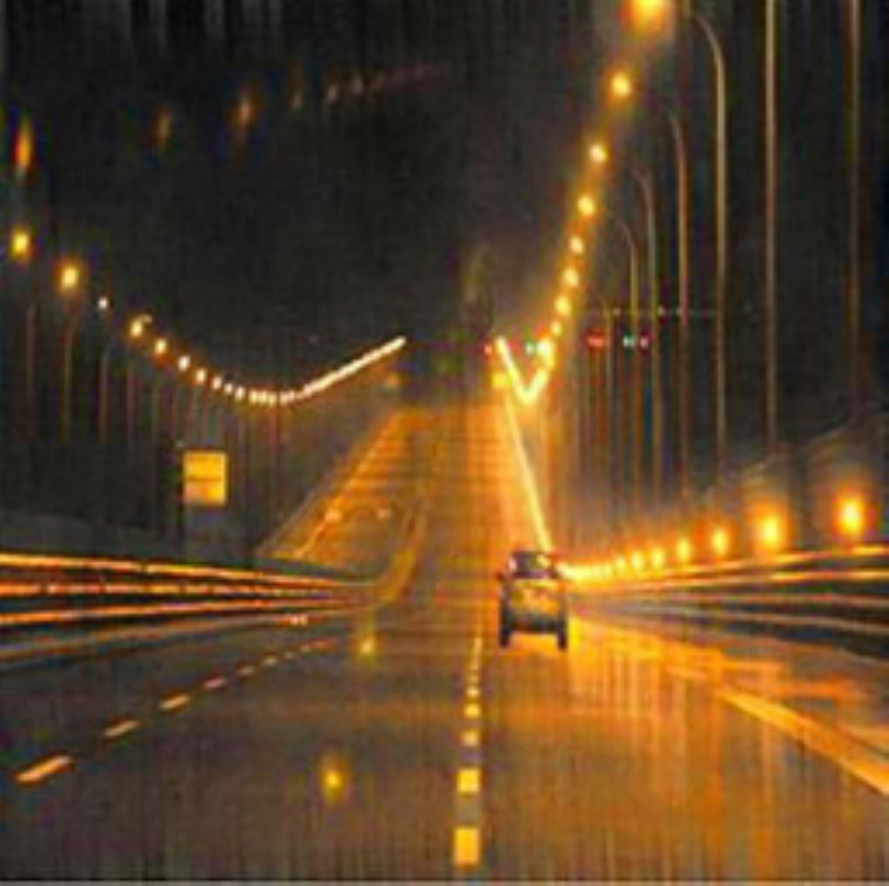}}
\end{minipage}
\hfill
\begin{minipage}{0.085\linewidth}
\centering{\includegraphics[width=1\linewidth]{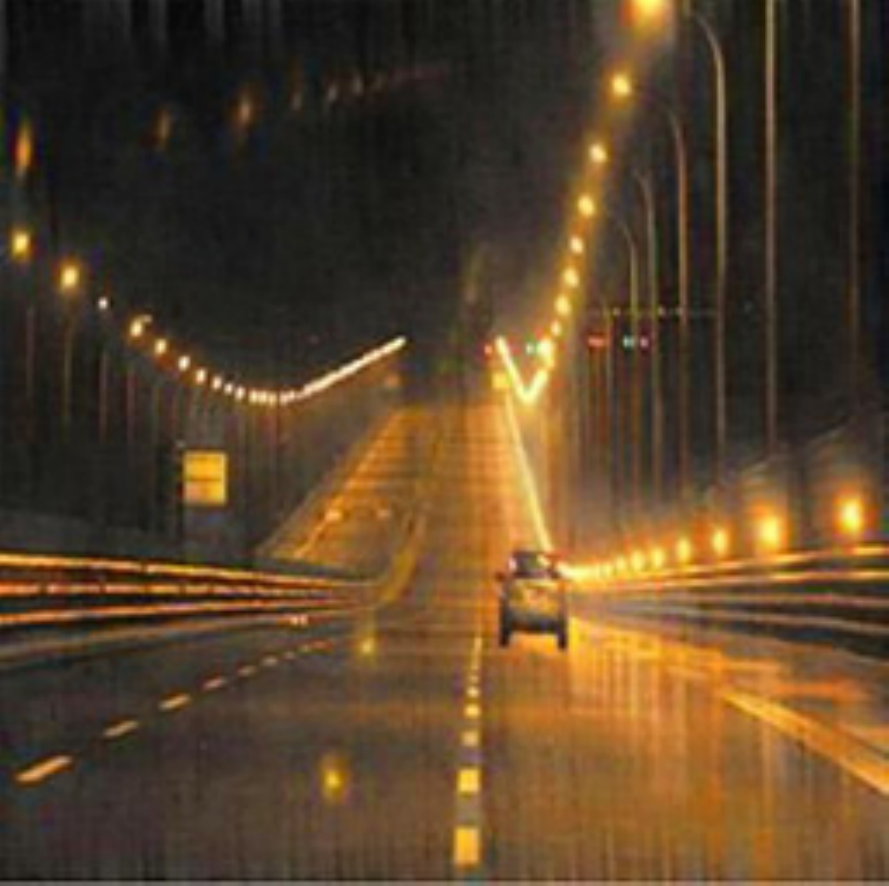}}
\end{minipage}
\hfill
\begin{minipage}{0.085\linewidth}
\centering{\includegraphics[width=1\linewidth]{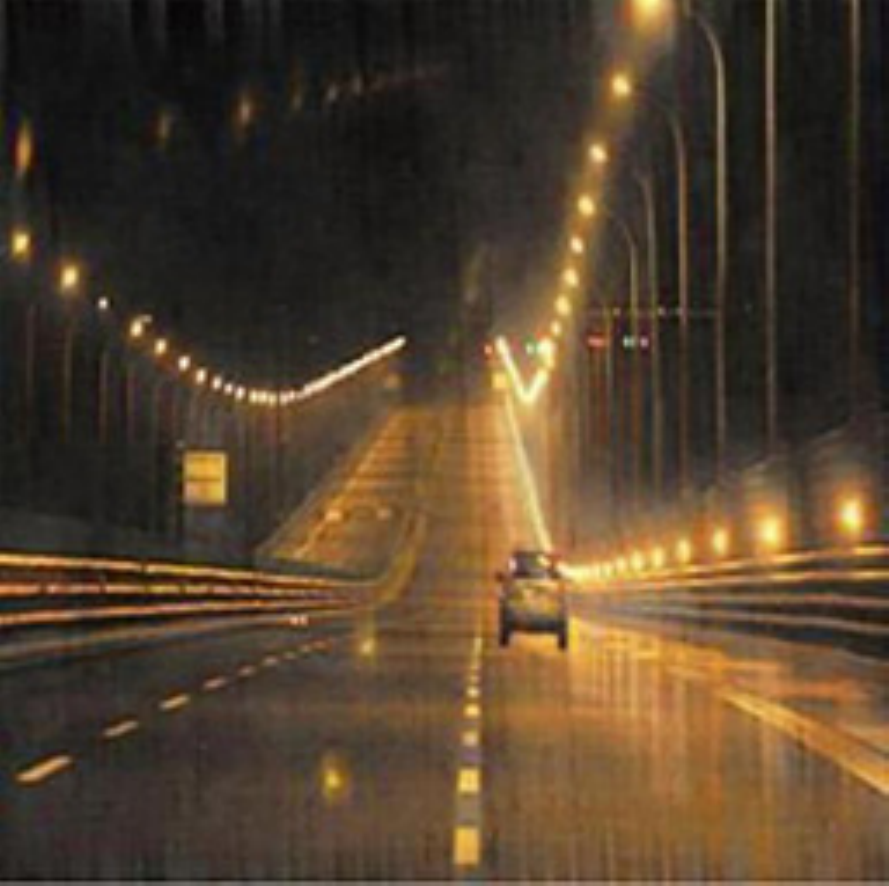}}
\end{minipage}
\vfill
\begin{minipage}{0.085\linewidth}
\centering{\includegraphics[width=1\linewidth]{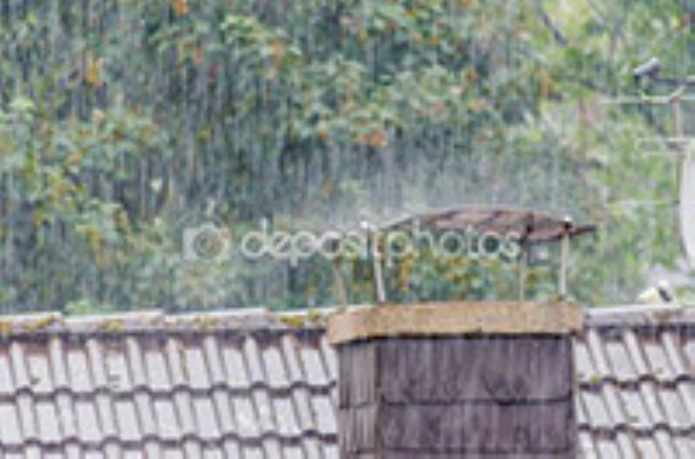}}
\end{minipage}
\hfill
\begin{minipage}{0.085\linewidth}
\centering{\includegraphics[width=1\linewidth]{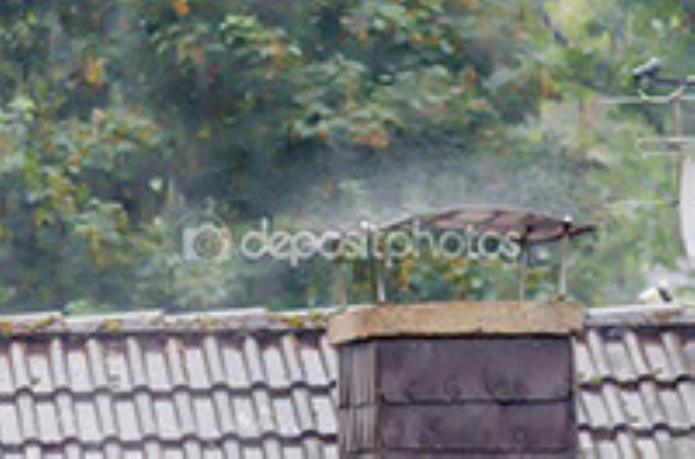}}
\end{minipage}
\hfill
\begin{minipage}{0.085\linewidth}
\centering{\includegraphics[width=1\linewidth]{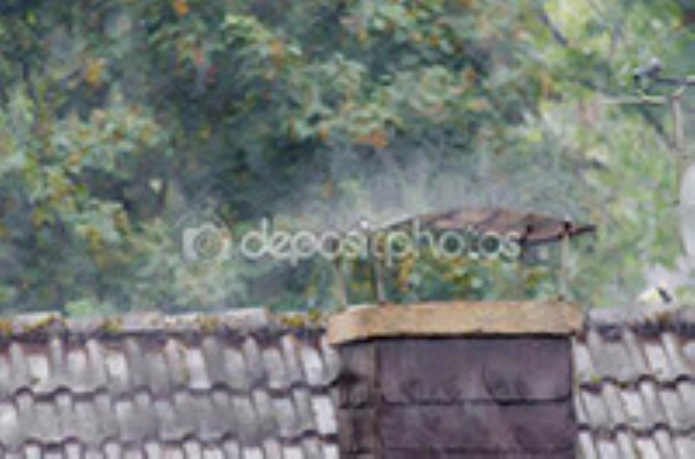}}
\end{minipage}
\hfill
\begin{minipage}{0.085\linewidth}
\centering{\includegraphics[width=1\linewidth]{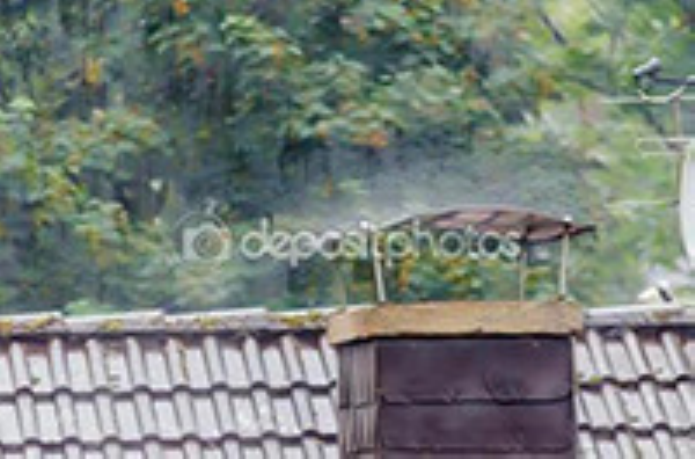}}
\end{minipage}
\hfill
\begin{minipage}{0.085\linewidth}
\centering{\includegraphics[width=1\linewidth]{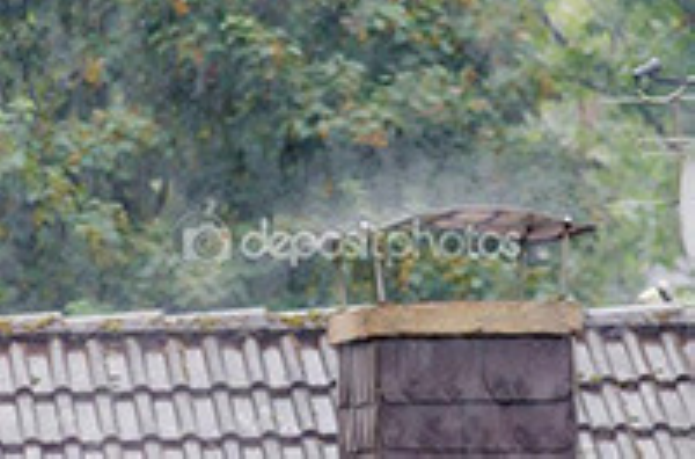}}
\end{minipage}
\hfill
\begin{minipage}{0.085\linewidth}
\centering{\includegraphics[width=1\linewidth]{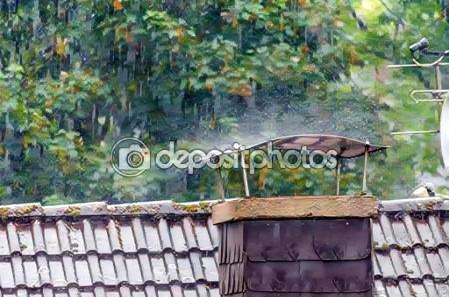}}
\end{minipage}
\hfill
\begin{minipage}{0.085\linewidth}
\centering{\includegraphics[width=1\linewidth]{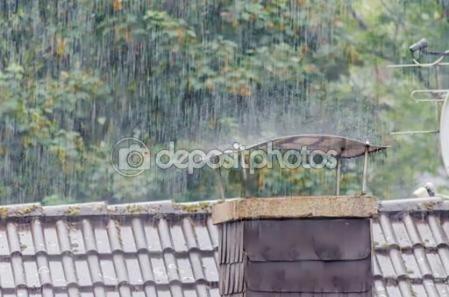}}
\end{minipage}
\hfill
\begin{minipage}{0.085\linewidth}
\centering{\includegraphics[width=1\linewidth]{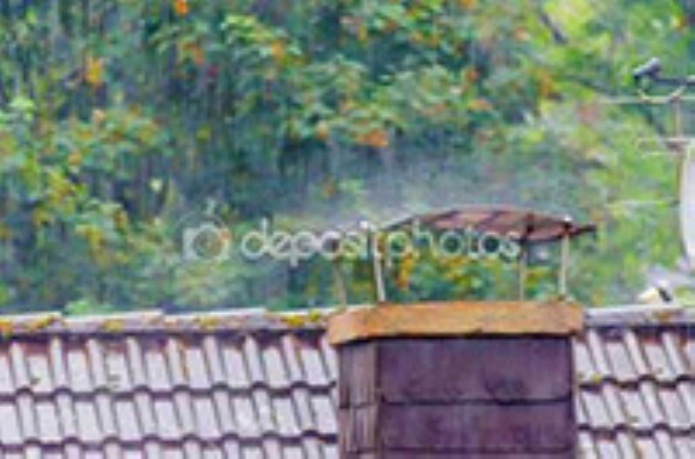}}
\end{minipage}
\hfill
\begin{minipage}{0.085\linewidth}
\centering{\includegraphics[width=1\linewidth]{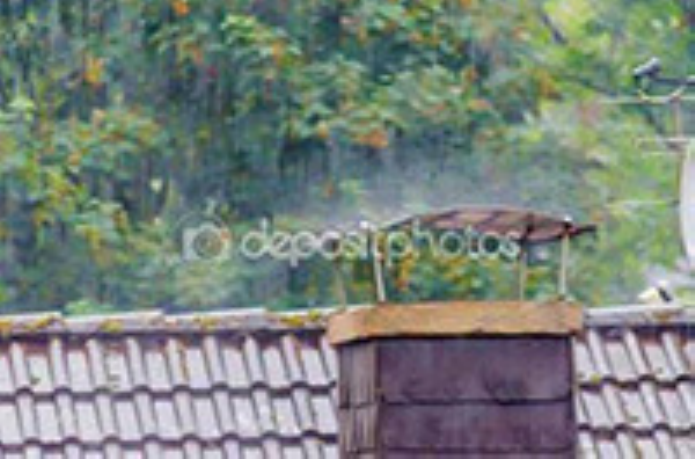}}
\end{minipage}
\hfill
\begin{minipage}{0.085\linewidth}
\centering{\includegraphics[width=1\linewidth]{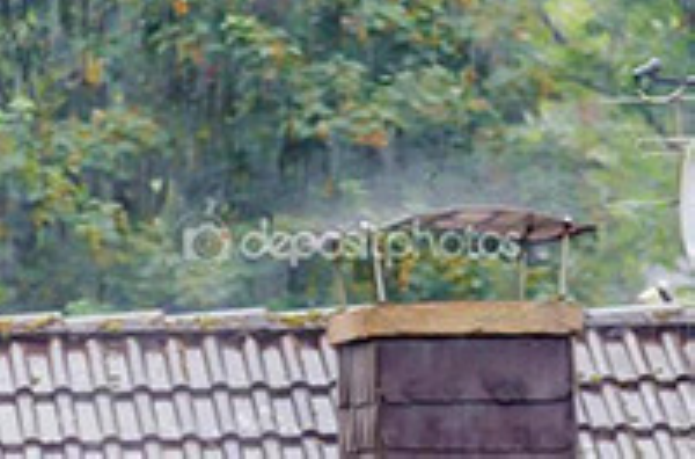}}
\end{minipage}
\hfill
\begin{minipage}{0.085\linewidth}
\centering{\includegraphics[width=1\linewidth]{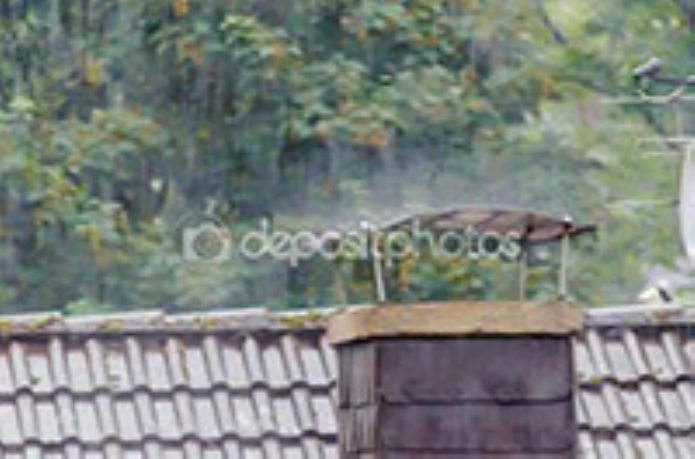}}
\end{minipage}
\vfill
\begin{minipage}{0.085\linewidth}
\centering{\includegraphics[width=1\linewidth]{images/rain-24.pdf}}
\centerline{(a)}
\end{minipage}
\hfill
\begin{minipage}{0.085\linewidth}
\centering{\includegraphics[width=1\linewidth]{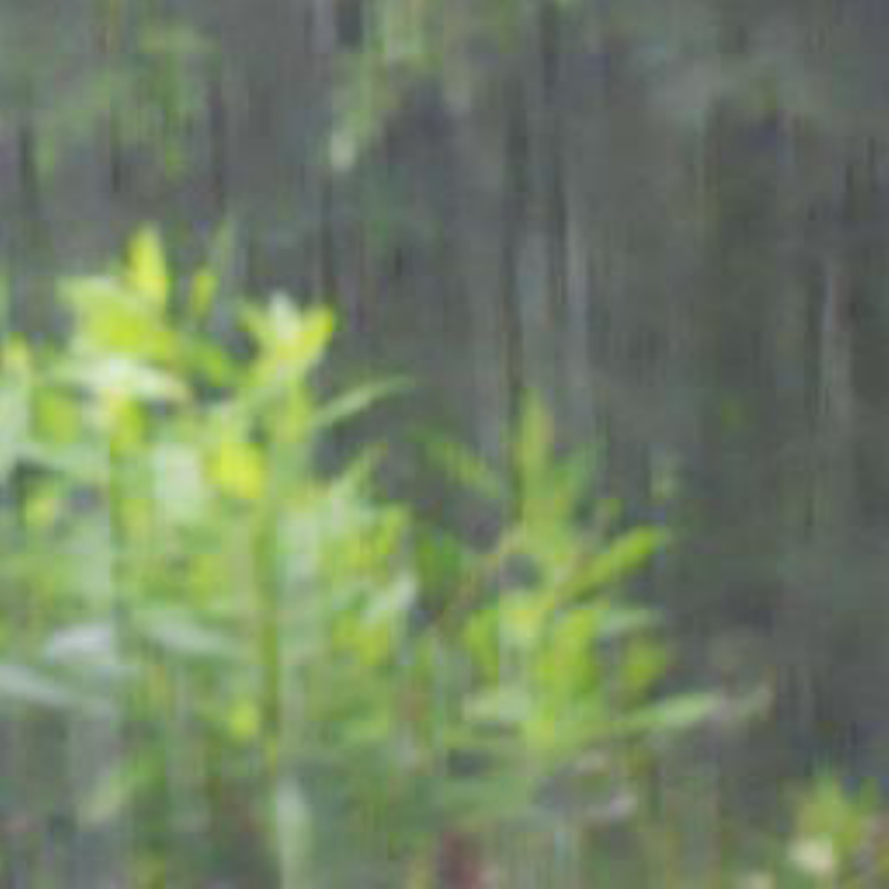}}
\centerline{(b)}
\end{minipage}
\hfill
\begin{minipage}{0.085\linewidth}
\centering{\includegraphics[width=1\linewidth]{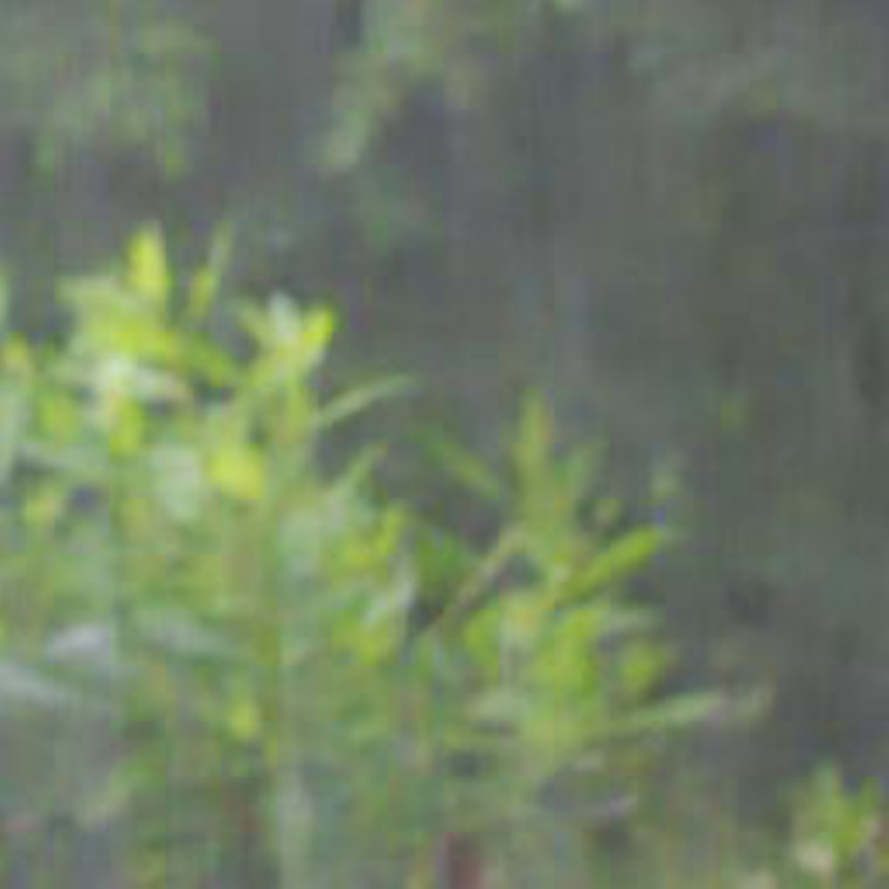}}
\centerline{(c)}
\end{minipage}
\hfill
\begin{minipage}{0.085\linewidth}
\centering{\includegraphics[width=1\linewidth]{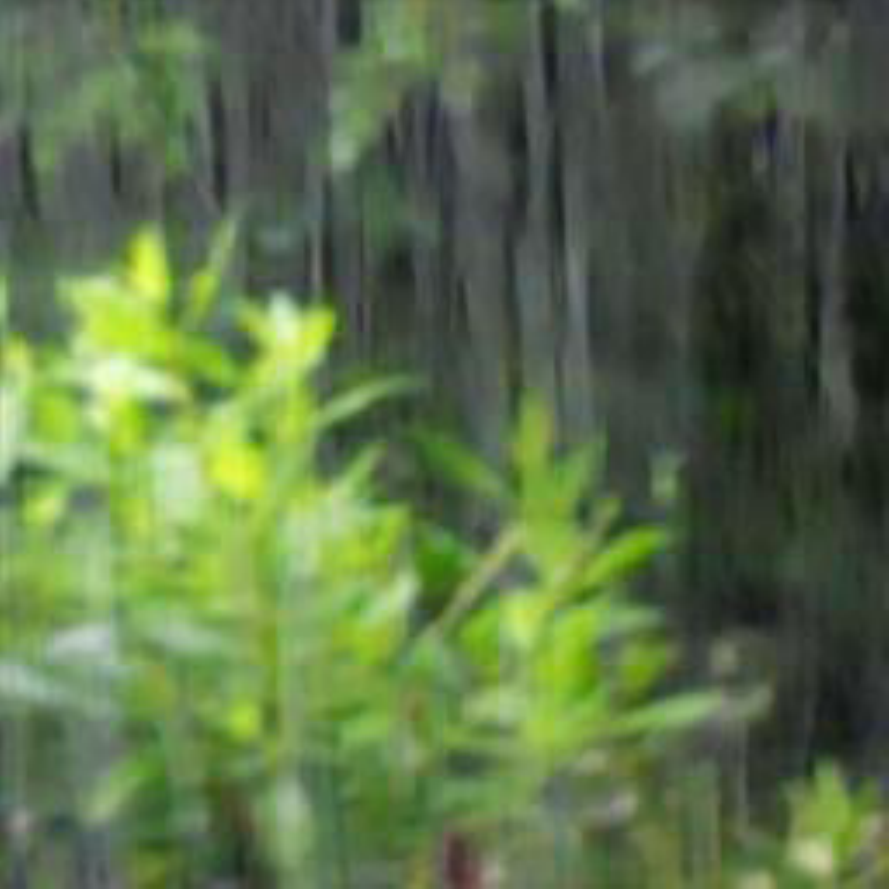}}
\centerline{(d)}
\end{minipage}
\hfill
\begin{minipage}{0.085\linewidth}
\centering{\includegraphics[width=1\linewidth]{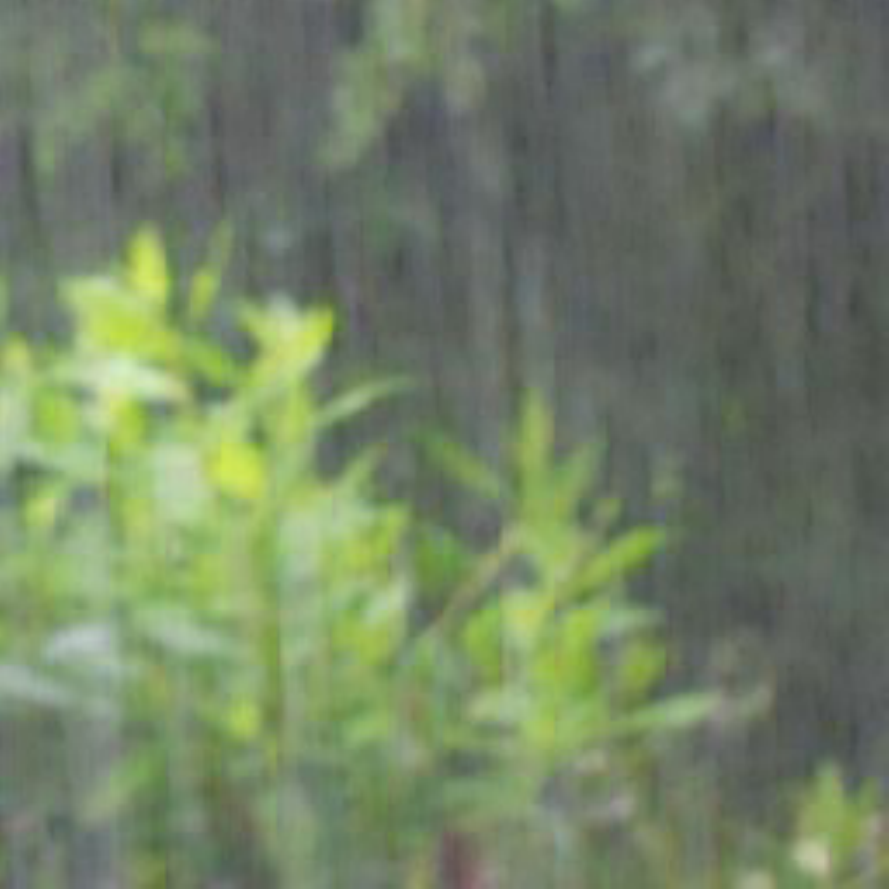}}
\centerline{(e)}
\end{minipage}
\hfill
\begin{minipage}{0.085\linewidth}
\centering{\includegraphics[width=1\linewidth]{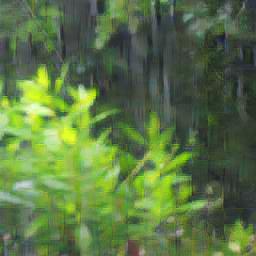}}
\centerline{(f)}
\end{minipage}
\hfill
\begin{minipage}{0.085\linewidth}
\centering{\includegraphics[width=1\linewidth]{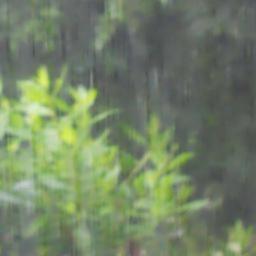}}
\centerline{(g)}
\end{minipage}
\hfill
\begin{minipage}{0.085\linewidth}
\centering{\includegraphics[width=1\linewidth]{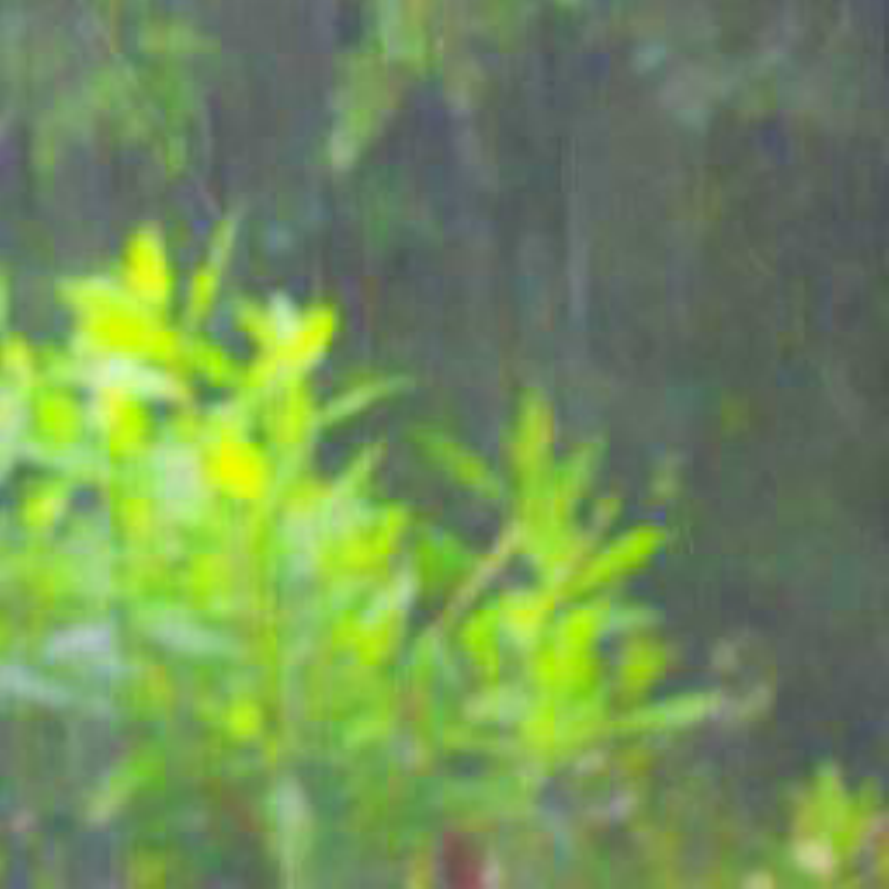}}
\centerline{(h)}
\end{minipage}
\hfill
\begin{minipage}{0.085\linewidth}
\centering{\includegraphics[width=1\linewidth]{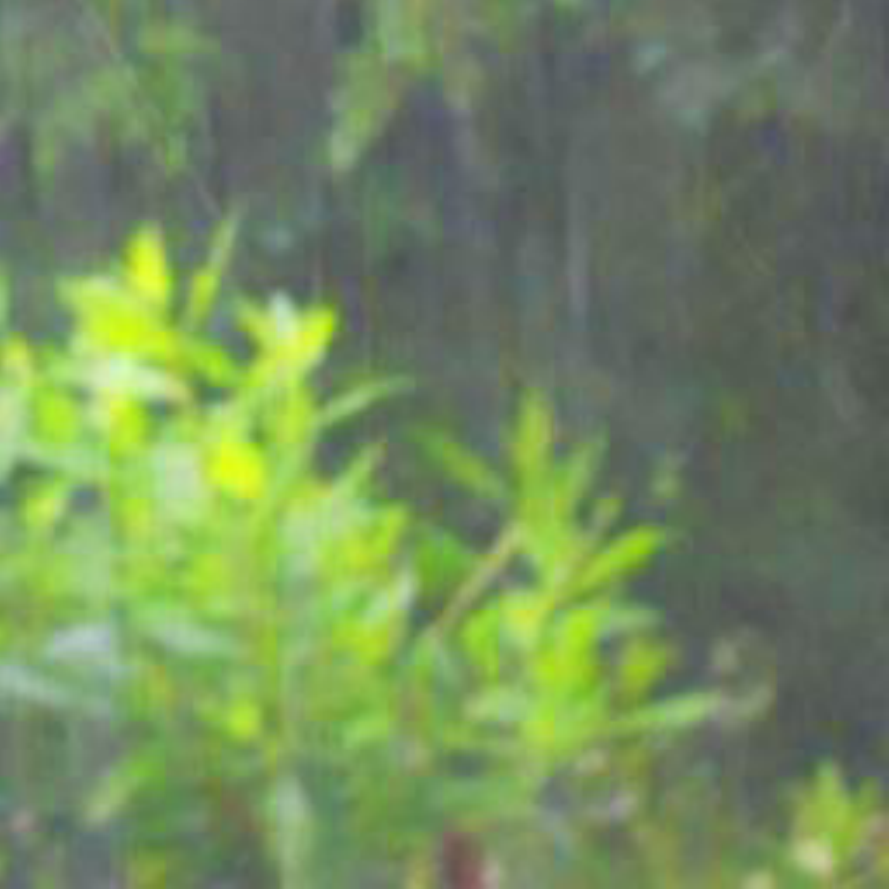}}
\centerline{(i)}
\end{minipage}
\hfill
\begin{minipage}{0.085\linewidth}
\centering{\includegraphics[width=1\linewidth]{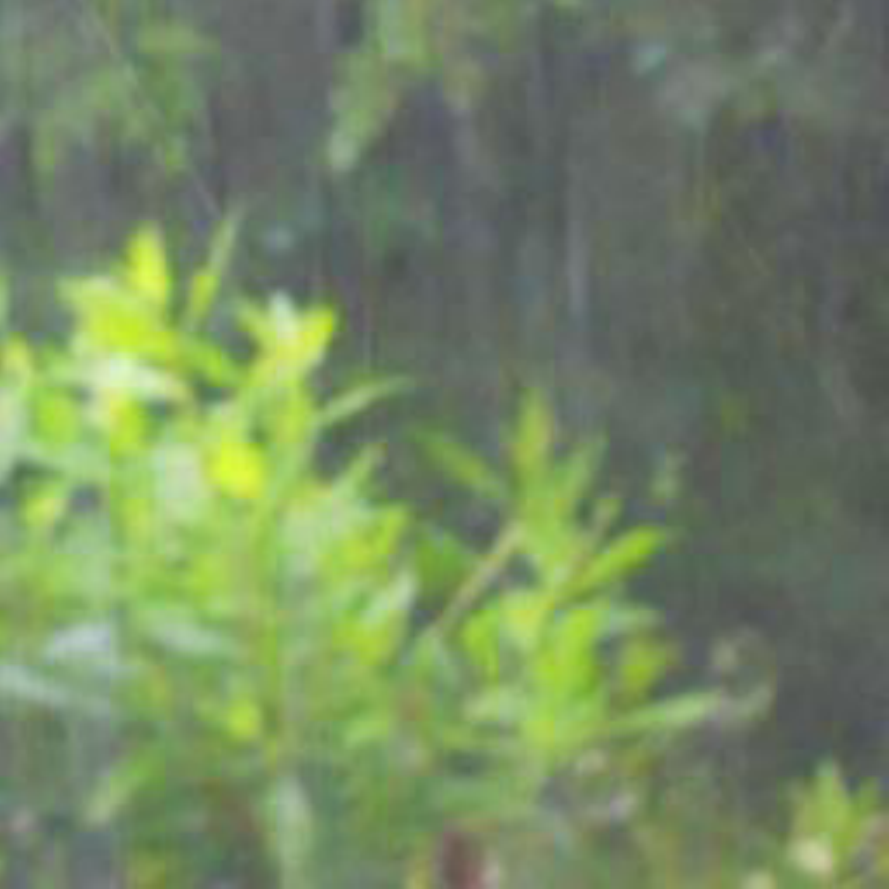}}
\centerline{(j)}
\end{minipage}
\hfill
\begin{minipage}{0.085\linewidth}
\centering{\includegraphics[width=1\linewidth]{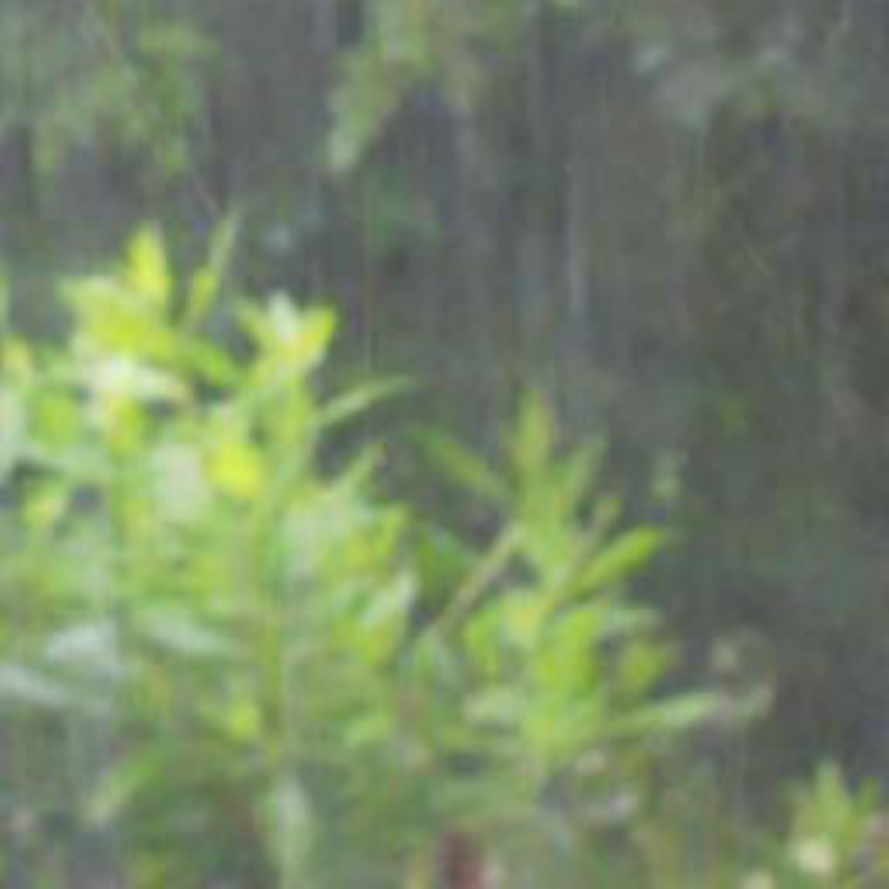}}
\centerline{(k)}
\end{minipage}
\end{center}
\caption{Rain removal results for real-world rainy images. (a) Input rainy image. (b)-(g) The results of \cite{Fu_2017_CVPR,Yang_2017_CVPR,Zhang_2018_CVPR,Li_2018_ECCV,Li_rt_2019_CVPR,Wang_ty_2019_CVPR}. (h)-(k) Our results with $\alpha$ equaling to $1.0$, $0.6$, $0.3$, $0.0$. Please zoom in to see clearly.}
\label{fig:real_word}
\end{figure*}

\begin{table*} [t]
\small
\newcommand{\tabincell}[2]{\begin{tabular}{@{}#1@{}}#2\end{tabular}}
\centering
\caption{Average running time on $512 \times 512$ image of different methods on our testing datasets.}
\begin{tabular}{c|c|c|c|c|c|c}
\hline
\cite{Fu_2017_CVPR} &  \cite{Yang_2017_CVPR} & \cite{Zhang_2018_CVPR} & \cite{Li_2018_ECCV}  &\cite{Li_rt_2019_CVPR} &\cite{Wang_ty_2019_CVPR} & Ours \\
\hline
0.09s & 1.40s & 0.06s & 0.50s & 0.45s & 0.66s & 0.05s      \\
\hline
\end{tabular}
\label{tab:time}
\end{table*}

\begin{table}[]
\centering
\caption{PSNR/SSIM of selected and our methods with different $\alpha$. Please pay attention to the role of different $\alpha$.}
\begin{tabular}{c|c|c}
\hline
 Baseline & Rain-I & Rain-II \\ \hline
 Metric &   PSNR/SSIM        &   PSNR/SSIM        \\ \hline
 \cite{Fu_2017_CVPR} &    29.22/0.867      &    29.86/0.901         \\
 \cite{Yang_2017_CVPR} &   27.22/0.832        &  29.25/0.886          \\
 \cite{Zhang_2018_CVPR} &   25.93/0.865         & 25.03/0.871       \\
 \cite{Li_2018_ECCV} &    27.38/0.881          &    27.56/0.899       \\
\cite{Li_rt_2019_CVPR} &    17.96/0.675          &    17.99/0.605       \\
 \cite{Wang_ty_2019_CVPR} &    28.43/0.848          &    30.53/0.905   \\ \hline
 Ours ($\alpha=1.0$)&    28.90/0.853        &    30.45/0.925      \\
 Ours ($\alpha=0.6$)&    30.13/0.887       &     31.96/0.940      \\
 Ours ($\alpha=0.3$)&    31.03/0.903       &    33.26/0.951       \\
 Ours ($\alpha=0.0$)&    \textbf{31.65}/\textbf{0.905} & \textbf{33.33}/\textbf{0.952}      \\ \hline
\end{tabular}
\label{tab:psnrssim_rain}
\end{table}

\subsection{Implementation Details}

In training, we crop image patches of $512 \times 512$ as training samples.
We adopt Adam \cite{Kingma_2015_ICLR} to train our network. The learning rate is initialized as $0.001$ and then decreased at each epoch by multiplying $0.1$.
Our network is implemented with PyTorch and tested on an NVIDIA 1080Ti GPU.
The batch sizes for training LocNet and the rest part are $4$ and $2$, respectively.

\section{Experiments} \label{sec:exp}

To assess the performances of our method quantitatively, we utilize PSNR and SSIM \cite{Wang_2004_TIP}
as evaluation metrics.
For real-world images, we only evaluate the visual performance.
We compare with different variants of our model as the ablation studies and compare with very recent state-of-the-art methods \cite{Fu_2017_CVPR,Yang_2017_CVPR,Zhang_2018_CVPR,Li_2018_ECCV,Li_rt_2019_CVPR,Wang_ty_2019_CVPR}.

\subsection{Datasets}

\noindent \textbf{Synthetic training and testing dataset}
We utilize the dataset by Li et al. \cite{Li_2018_arxiv} as our training dataset.
For our LocNet, we utilize the dataset of \cite{Yang_2017_CVPR}, which includes $2000$ pairs of samples.
We validate the proposed method on two testing datasets. Firstly, we conduct a testing dataset with $300$ images by randomly selecting $100$ testing samples from the testing datasets of \cite{Fu_2017_CVPR,Li_2018_ECCV,Zhang_2018_CVPR}, respectively, which is referred to as Rain-I dataset.
Secondly, we also synthesize another dataset Rain-II with $400$ images suffering apparent haze-like effect \footnote{http://www.photoshopessentials.com/photo-effects/rain/}.
Additionally, a different dataset with very strong rain streaks was proposed in \cite{Yang_2017_CVPR}. Considering that the cases in this dataset are not very correlated to the proposed method, we only show visual results in Figure \ref{fig:synthetic_1} and leave the objective results in the supplementary materials.

\noindent \textbf{Real-world dataset}
Real-world rainy images selected from the Internet and other works \cite{Fu_2017_CVPR,Li_2018_ECCV,Yang_2017_CVPR,Zhang_2018_CVPR,Li_rt_2019_CVPR,Wang_ty_2019_CVPR} are used to evaluate the methods.
Our real-world images include images with light and heavy rain and various contents, including people, landscape, city scenes, and so on.

\subsection{Studies of the Model-based Estimation}

In the proposed model, we jointly train the two-branch model to predict the coefficients $\mathbf{R}$ and $\mathbf{T}$. The supervision on the final target image $\bB$, and the rain model-based loss functions $\mathcal{L}_1$ and $\mathcal{L}_2$ are applied to restrict the prediction of $\mathbf{R}$ and $\mathbf{T}$. To study the intermediate estimation (for $\mathbf{R}$ and $\mathbf{T}$) of the proposed model, we visualize two real examples in Figure \ref{fig:betas}. As shown in Figure \ref{fig:betas}, the estimated $\mathbf{R}$ and $\mathbf{T}$ can clearly reflect the degeneration caused by the strong rain streaks and the haze-like mist, respectively. In the visualization of $\mathbf{I}-\mathbf{R}$, we can see that most of the strong rain streaks are removed. After processing with $\mathbf{T}$, the mist in the images is further removed. Furthermore, we can observe that the operation with $\mathbf{T}$ can compensate for the underestimate of $\mathbf{R}$. As the supervision is only imposed on the final deraining result $\mathbf{B}$, the model-based estimation may not exactly work as the original physical meaning, but the process with $\mathbf{R}$ and $\mathbf{T}$ can lead to desired results jointly.

\subsection{Quantitative Evaluation on Synthetic Datasets}

Table \ref{tab:psnrssim_rain} shows the PSNR/SSIM values of different methods on the
testing datasets Rain-I and Rain-II.
The results of the proposed method (with different settings for $\alpha$) are superior or comparative to other state-of-the-art methods.
When $\alpha=1$ (without RefNet), our method has comparable PSNR/SSIM with the other methods. After RefNet is added, our PSNR/SSIM surpass them.
We observe that in many ground truth images, slight haze-like effect exists. The models with high $\alpha$ value tend to remove the haze-like mist (in ground truth), leading to brighter color hue but lower objective index, e.g., PSNR and SSIM.
We visualize the results of different methods and our method with different $\alpha$ in Figure \ref{fig:synthetic_1}. The results obtained by the proposed method with different $\alpha$ values are better than other methods.

In Table \ref{tab:time}, we show the averaged running time consumed by selected methods on our testing datasets. Our method can obtain better results with less running time.

\subsection{Qualitative Evaluation on Real-World Images}

In this section, we show the visual results on real-world images in Figure \ref{fig:real_word}.
We can see that our method outperforms other state-of-the-art methods.
Rain streaks and haze-like effect are both removed.
Figure \ref{fig:real_word}(h) is the result of our two-branch unit without $\mathcal{R}(\cdot)$ (by setting $\alpha=1$), in which colors become brighter than rainy images.
This is the result of removing the haze-like effect. The color of watermark letters in the second line images is pale, which is kept in our result in Figure \ref{fig:real_word}(h). This can imply that our method does not introduce abnormal hue in the deraining results. The results of tuning the degree of removing haze-like effect (by setting different $\alpha$) are shown in Figure \ref{fig:real_word}(i)(j)(k). We can see that the degree of removing haze-like effect can be controlled by $\alpha$ flexibly.
Our results are closer to reality, and the image details are also better. The work \cite{Li_rt_2019_CVPR} can also remove haze-like effect for some images, but it causes apparent blocking effect ((f) in the first and third lines of Figure \ref{fig:real_word}), and this method can make image very dark and produce unnatural hue ((g) in the second line of Figure \ref{fig:synthetic_1}). Other selected methods cannot remove haze-like effect well, and some apparent rain streaks with blurry edges remain in the final results for some images.

\begin{table}[]
\newcommand{\tabincell}[2]{\begin{tabular}{@{}#1@{}}#2\end{tabular}}
\centering
\caption{PSNR/SSIM of the variants of our AMPE-Net}
\begin{tabular}{ccccc|c|c}
\hline
$\mathcal{H}$ & $\mathcal{F}$ & $\mathcal{G}$ & $\mathcal{L}_{1}$ & $\mathcal{L}_{2}$ & Rain-I & Rain-II \\ \hline
$\surd$ &  & $\surd$ & $\surd$ & $\surd$ & \tabincell{c}{26.42/0.794} & \tabincell{c}{28.58/0.888} \\ \hline
 & $\surd$ & $\surd$ & $\surd$ & $\surd$ & \tabincell{c}{27.80/0.830} & \tabincell{c}{29.67/0.914} \\ \hline
$\surd$ & $\surd$ & $\surd$ &  & $\surd$ & \tabincell{c}{28.63/0.849} & \tabincell{c}{30.37/0.923} \\ \hline
$\surd$ & $\surd$ & $\surd$ & $\surd$ & $\surd$ & \tabincell{c}{\textbf{28.90}/\textbf{0.853}} & \tabincell{c}{\textbf{30.45}/\textbf{0.925}} \\ \hline
\end{tabular}
\label{tab:psnrssim_ablation}
\end{table}

\begin{figure}[t]
\begin{center}
\begin{minipage}{0.155\linewidth}
\centering{\includegraphics[width=1\linewidth]{images/gt-5.pdf}}
\end{minipage}
\hfill
\begin{minipage}{0.155\linewidth}
\centering{\includegraphics[width=1\linewidth]{images/Synthesized_rain-5.pdf}}
\end{minipage}
\hfill
\begin{minipage}{0.155\linewidth}
\centering{\includegraphics[width=1\linewidth]{images/srain-5_lambda_0_3_Derain.pdf}}
\end{minipage}
\hfill
\begin{minipage}{0.155\linewidth}
\centering{\includegraphics[width=1\linewidth]{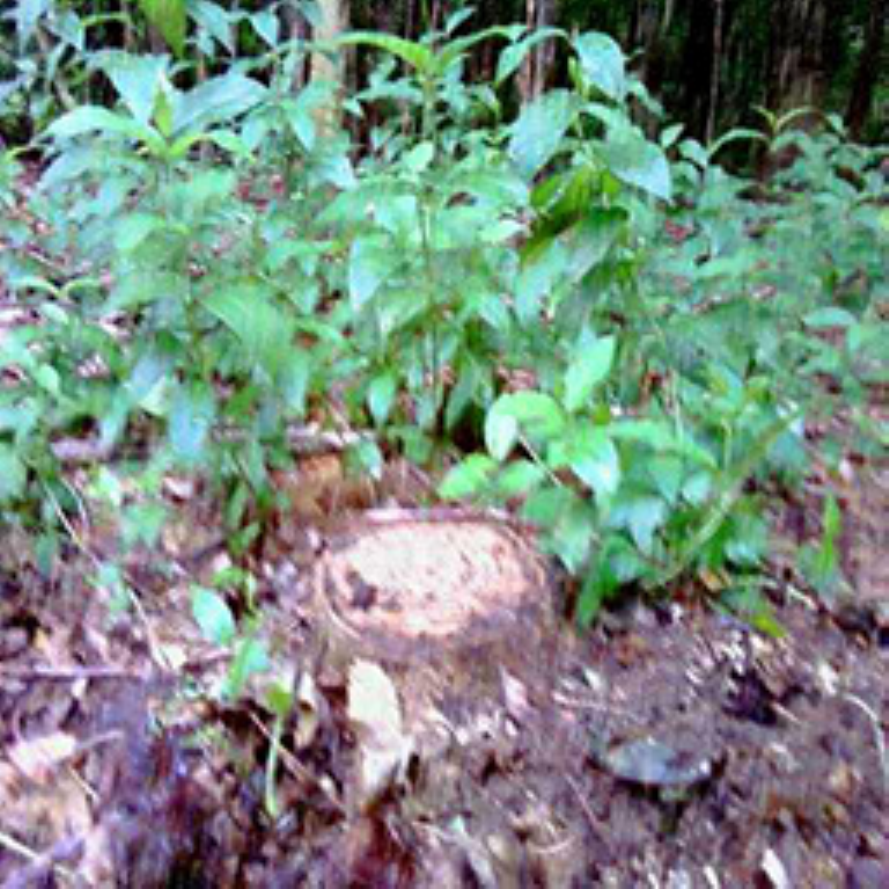}}
\end{minipage}
\hfill
\begin{minipage}{0.155\linewidth}
\centering{\includegraphics[width=1\linewidth]{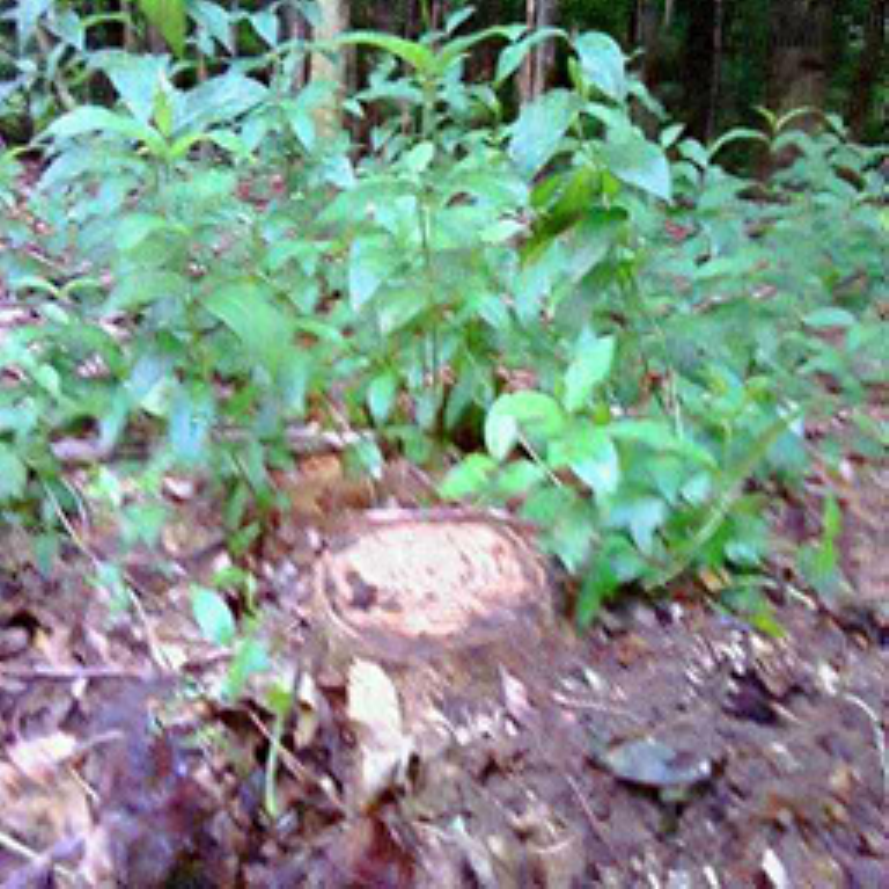}}
\end{minipage}
\hfill
\begin{minipage}{0.155\linewidth}
\centering{\includegraphics[width=1\linewidth]{images/Synthesize_Derain-5_one_loss.pdf}}
\end{minipage}
\vfill
\begin{minipage}{0.155\linewidth}
\centering{\includegraphics[width=1\linewidth]{images/Gt-33_new.pdf}}
\centerline{(a)}
\end{minipage}
\hfill
\begin{minipage}{0.155\linewidth}
\centering{\includegraphics[width=1\linewidth]{images/Synthetic_rain-33_new.pdf}}
\centerline{(b)}
\end{minipage}
\hfill
\begin{minipage}{0.155\linewidth}
\centering{\includegraphics[width=1\linewidth]{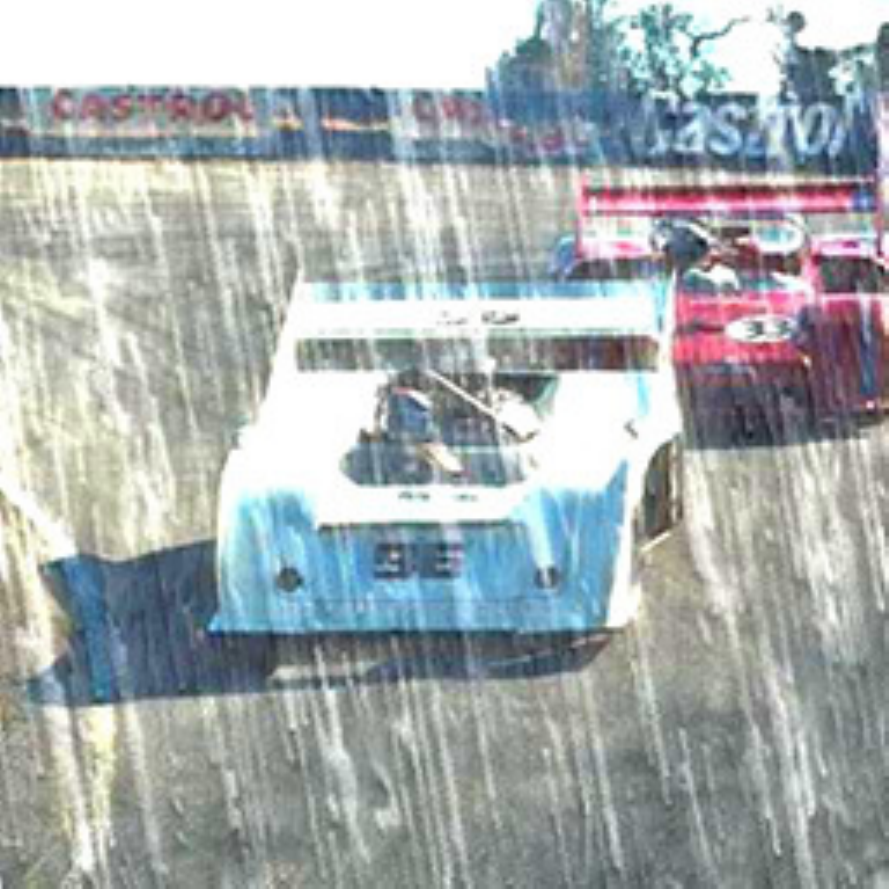}}
\centerline{(c)}
\end{minipage}
\hfill
\begin{minipage}{0.155\linewidth}
\centering{\includegraphics[width=1\linewidth]{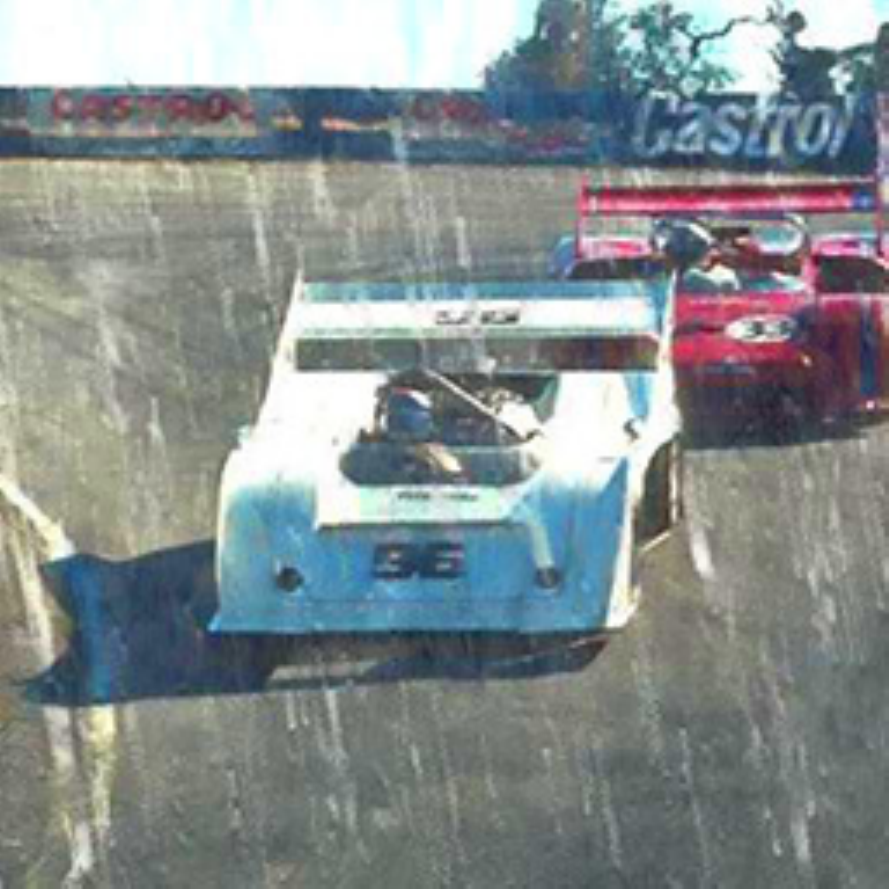}}
\centerline{(d)}
\end{minipage}
\hfill
\begin{minipage}{0.155\linewidth}
\centering{\includegraphics[width=1\linewidth]{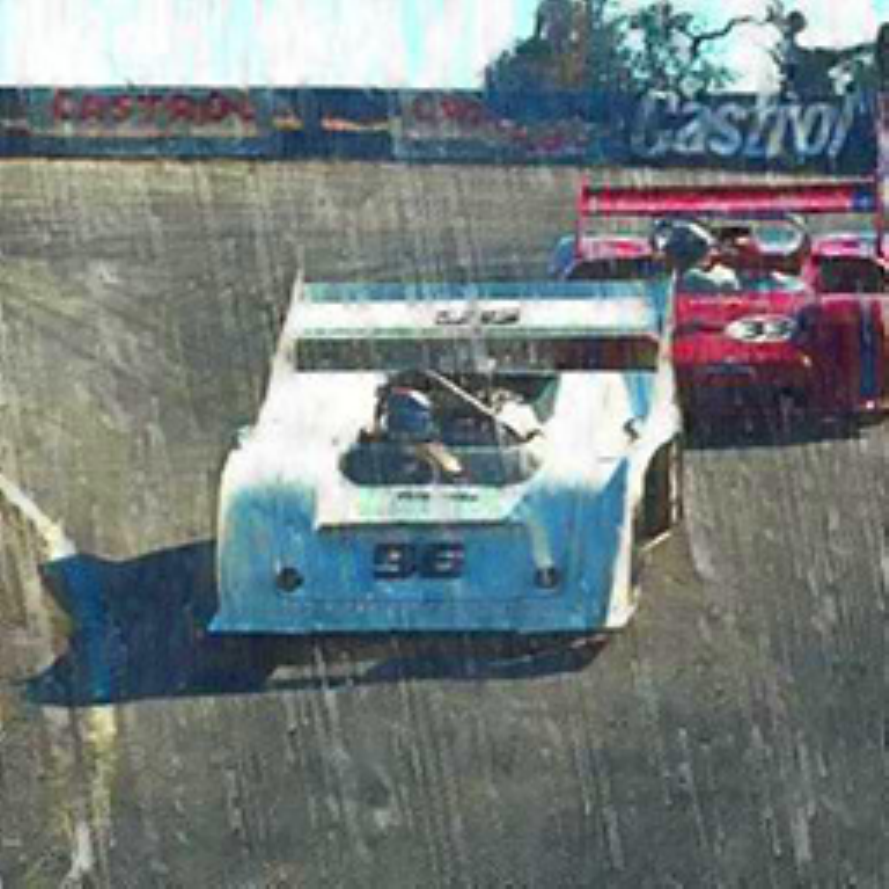}}
\centerline{(e)}
\end{minipage}
\hfill
\begin{minipage}{0.155\linewidth}
\centering{\includegraphics[width=1\linewidth]{images/Yang_test33_derain_I.pdf}}
\centerline{(f)}
\end{minipage}
\end{center}
\caption{Visual results of ablation studies on synthetic rainy images. (a) Ground truth. (b) Synthetic rainy image. (c-f) Results of $\mathcal{H}+\mathcal{G}$, $\mathcal{F}+\mathcal{G}$, $\mathcal{H}+\mathcal{F}+\mathcal{G}$ (w/o $\mathcal{L}_{2}$), $\mathcal{H}+\mathcal{F}+\mathcal{G}$ (our whole model).
}
\label{fig:synthetic_ablation}
\end{figure}

\begin{figure}[t!]
\begin{center}
\begin{minipage}{0.19\linewidth}
\centering{\includegraphics[width=1\linewidth]{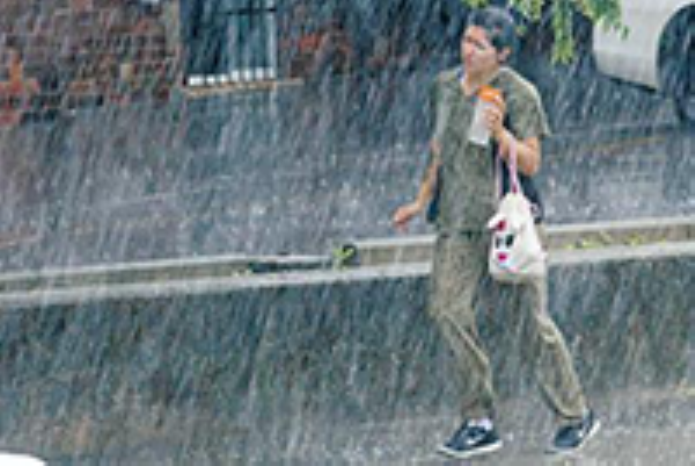}}
\end{minipage}
\hfill
\begin{minipage}{0.19\linewidth}
\centering{\includegraphics[width=1\linewidth]{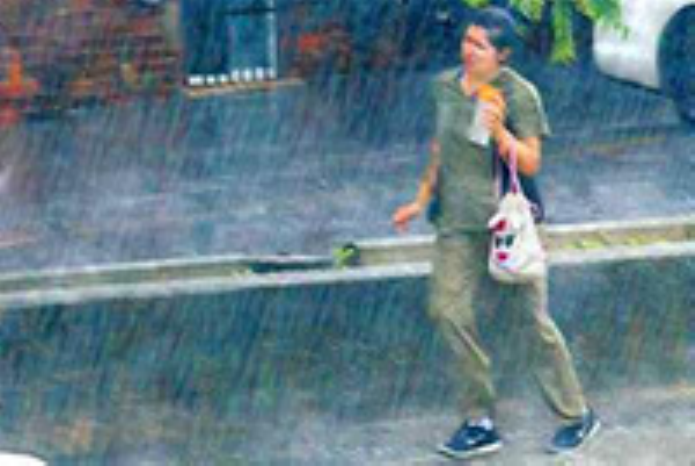}}
\end{minipage}
\hfill
\begin{minipage}{0.19\linewidth}
\centering{\includegraphics[width=1\linewidth]{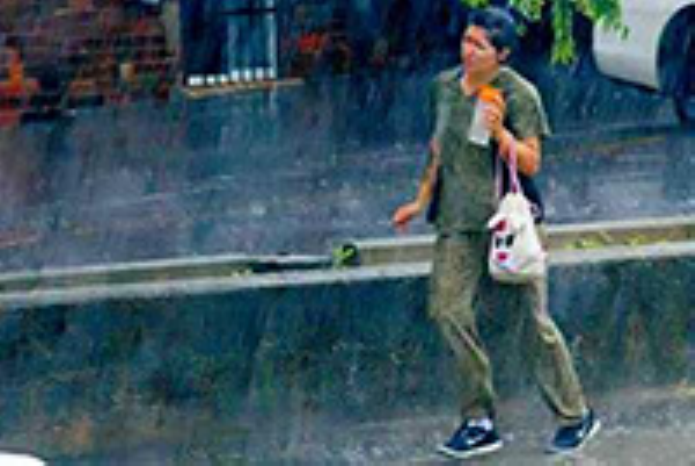}}
\end{minipage}
\hfill
\begin{minipage}{0.19\linewidth}
\centering{\includegraphics[width=1\linewidth]{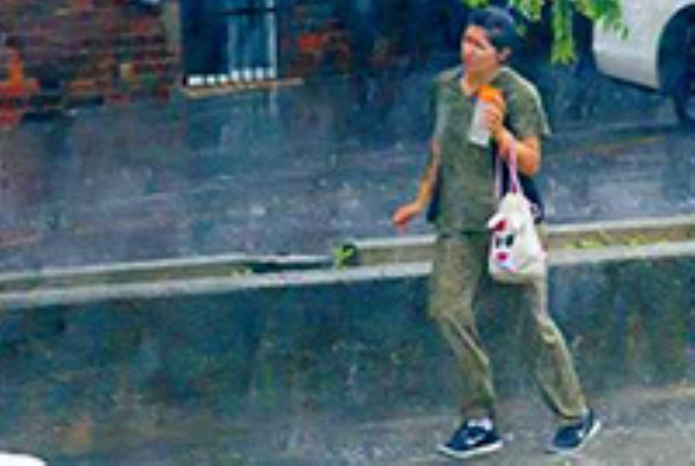}}
\end{minipage}
\hfill
\begin{minipage}{0.19\linewidth}
\centering{\includegraphics[width=1\linewidth]{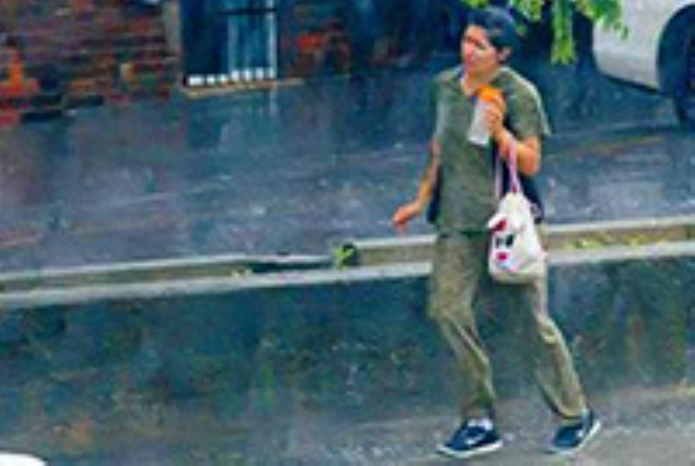}}
\end{minipage}
\vfill
\begin{minipage}{0.19\linewidth}
\centering{\includegraphics[width=1\linewidth]{images/rain-81.pdf}}
\centerline{(a)}
\end{minipage}
\hfill
\begin{minipage}{0.19\linewidth}
\centering{\includegraphics[width=1\linewidth]{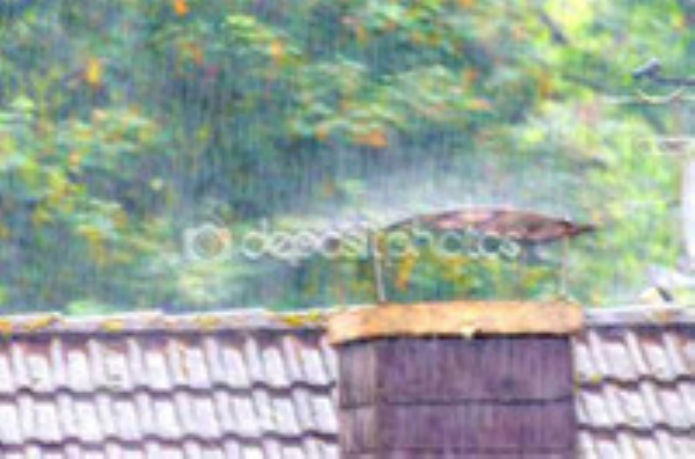}}
\centerline{(b)}
\end{minipage}
\hfill
\begin{minipage}{0.19\linewidth}
\centering{\includegraphics[width=1\linewidth]{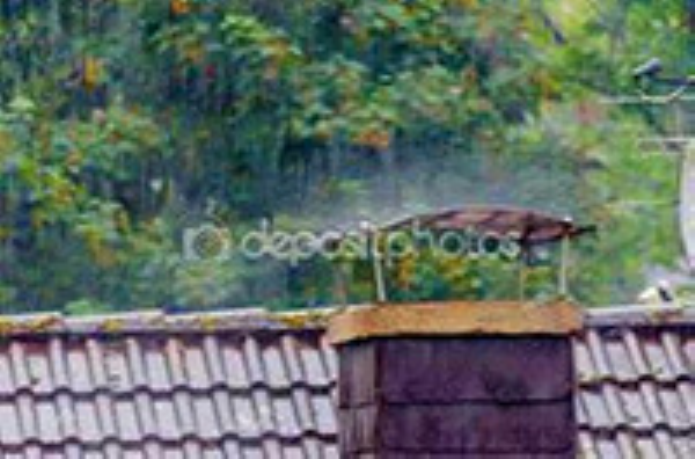}}
\centerline{(c)}
\end{minipage}
\hfill
\begin{minipage}{0.19\linewidth}
\centering{\includegraphics[width=1\linewidth]{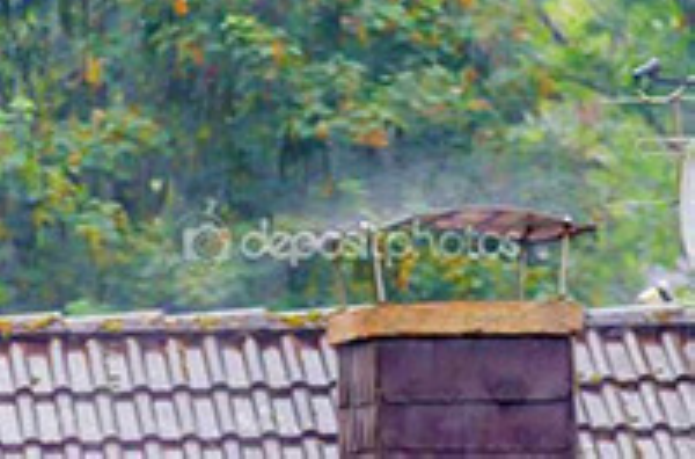}}
\centerline{(d)}
\end{minipage}
\hfill
\begin{minipage}{0.19\linewidth}
\centering{\includegraphics[width=1\linewidth]{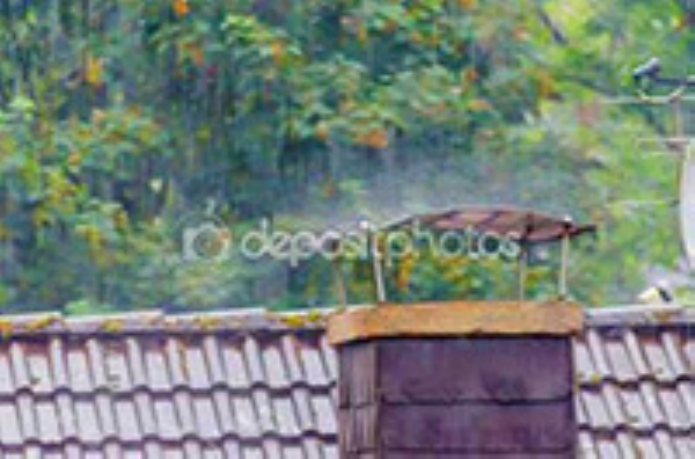}}
\centerline{(e)}
\end{minipage}
\end{center}
\caption{Visual results of ablation studies on real-world rainy images. (a) Rainy images. (b-e) Results of $\mathcal{H}+\mathcal{G}$, $\mathcal{F}+\mathcal{G}$, $\mathcal{H}+\mathcal{F}+\mathcal{G}$ (w/o $\mathcal{L}_{2}$), $\mathcal{H}+\mathcal{F}+\mathcal{G}$ (our whole model).}
\label{fig:parctical_ablation}
\end{figure}

\begin{figure}[t!]
\begin{center}
\begin{minipage}{0.19\linewidth}
\centering{\includegraphics[width=1\linewidth]{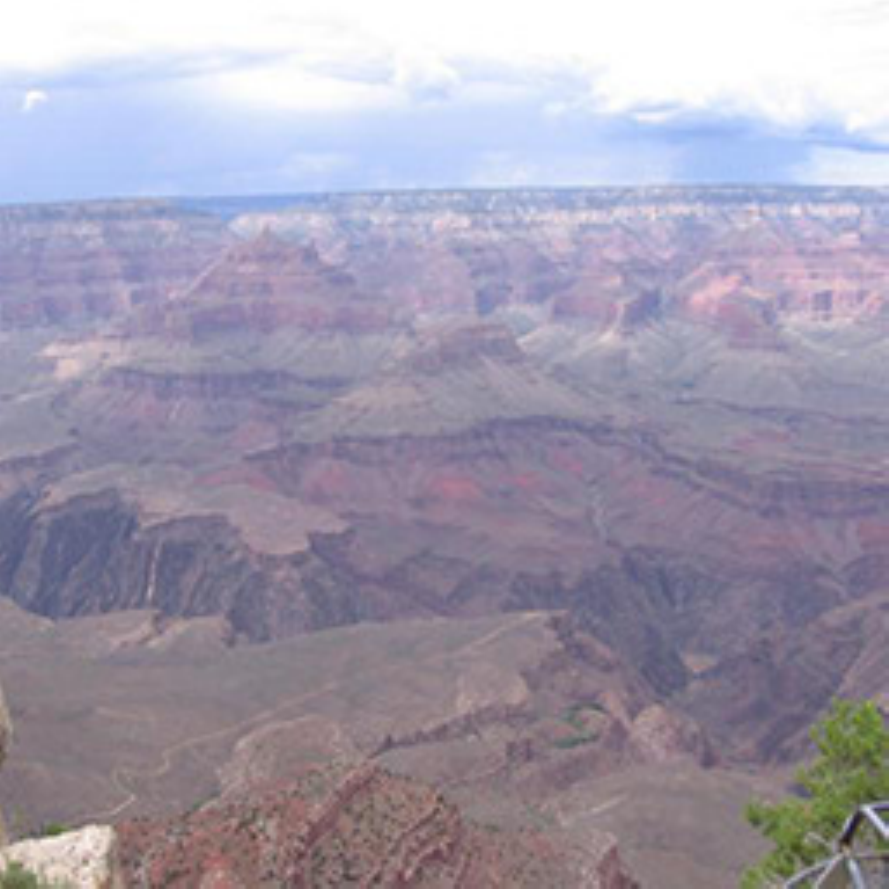}}
\end{minipage}
\hfill
\begin{minipage}{0.19\linewidth}
\centering{\includegraphics[width=1\linewidth]{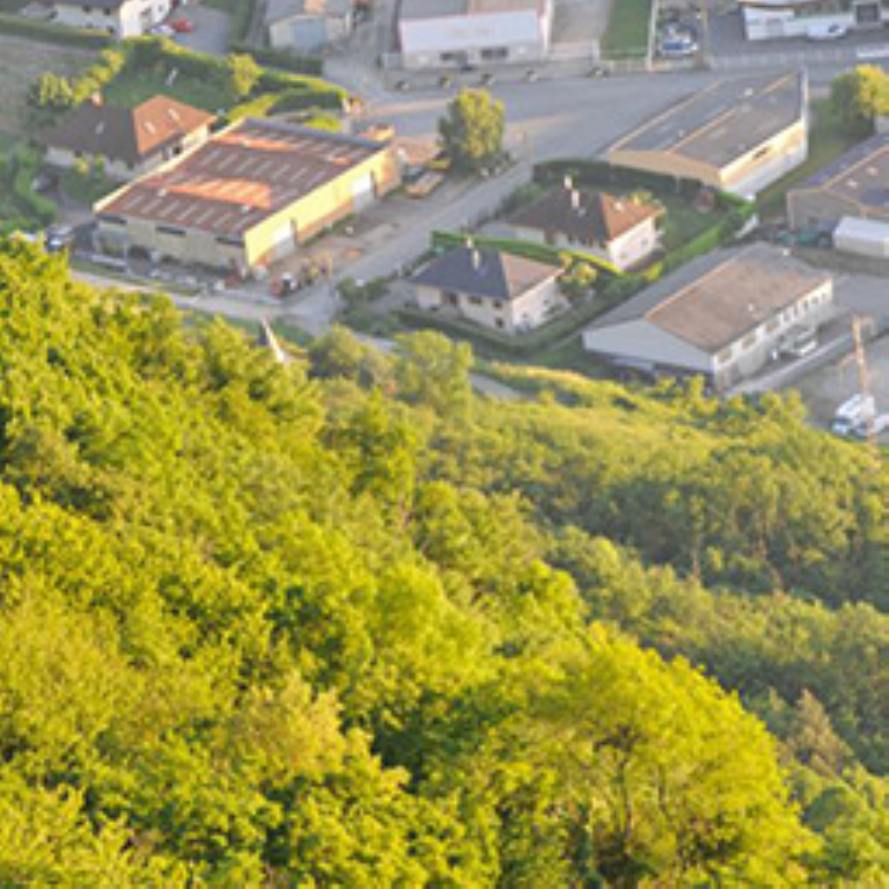}}
\end{minipage}
\hfill
\begin{minipage}{0.19\linewidth}
\centering{\includegraphics[width=1\linewidth]{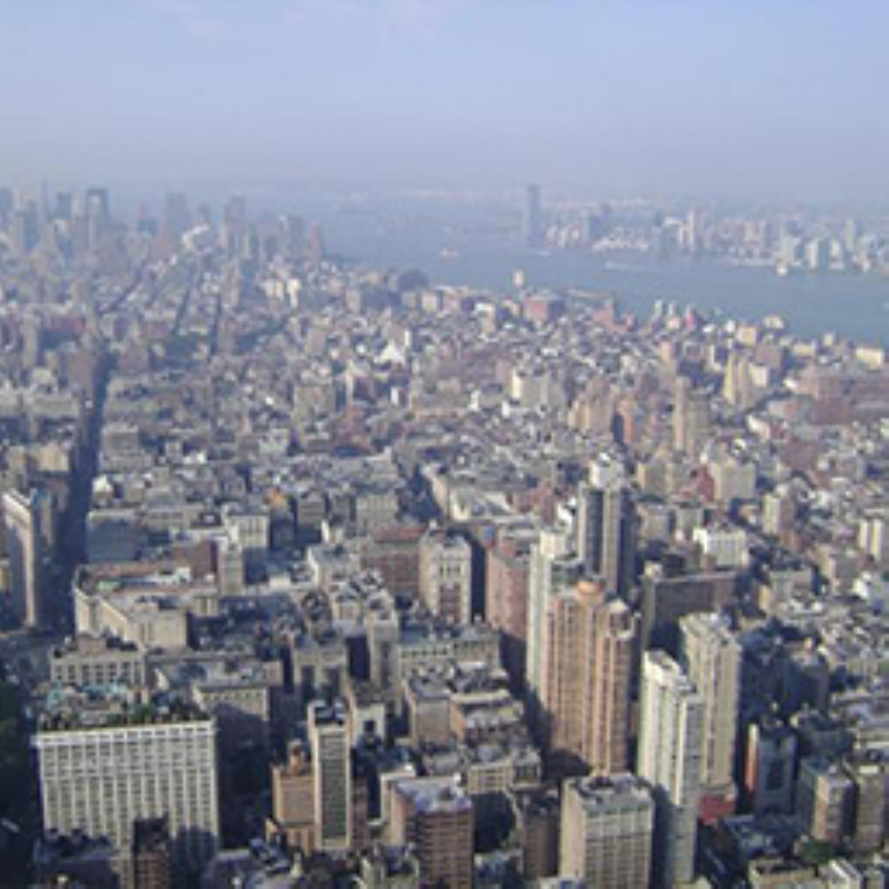}}
\end{minipage}
\hfill
\begin{minipage}{0.19\linewidth}
\centering{\includegraphics[width=1\linewidth]{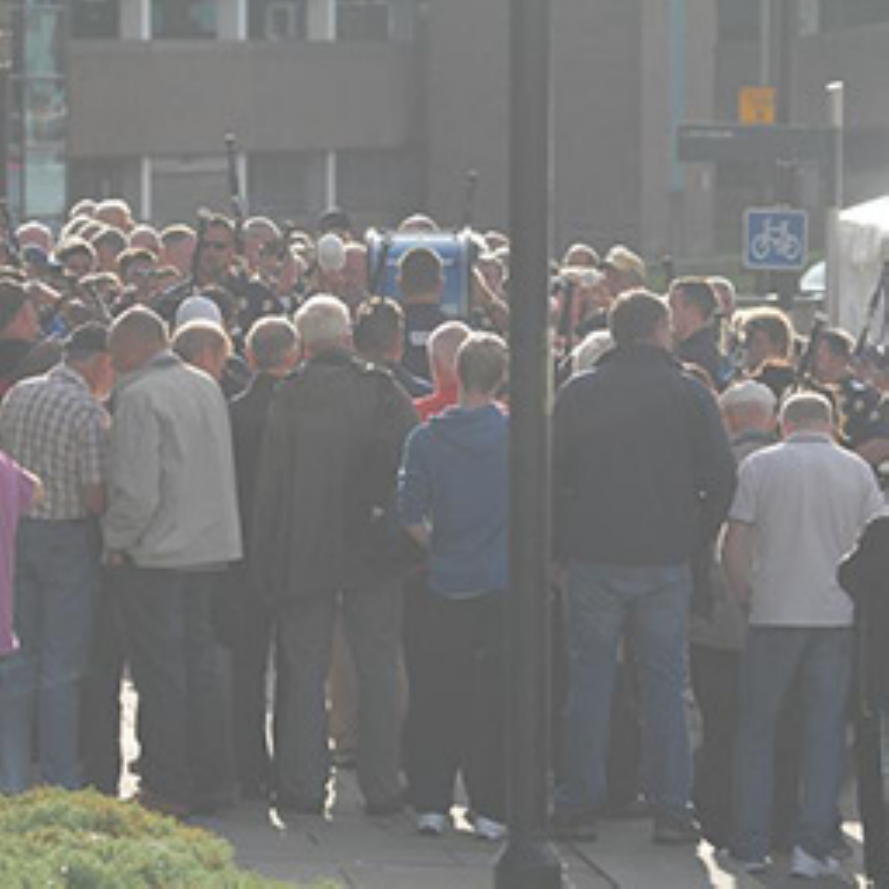}}
\end{minipage}
\hfill
\begin{minipage}{0.19\linewidth}
\centering{\includegraphics[width=1\linewidth]{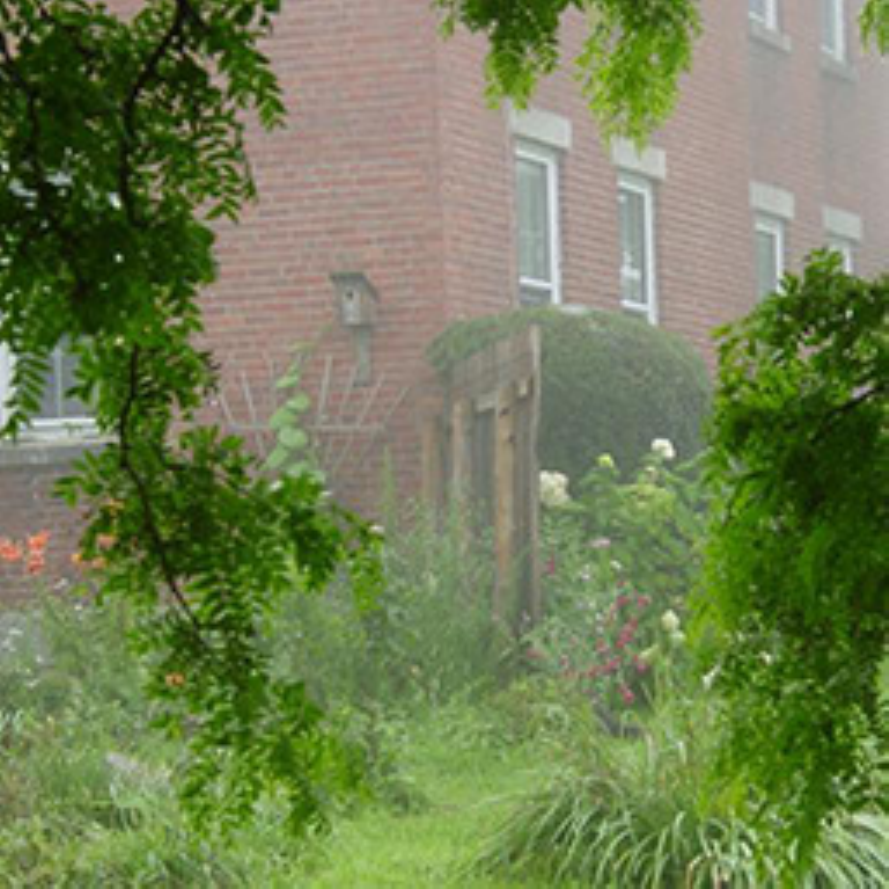}}
\end{minipage}
\vfill
\begin{minipage}{0.19\linewidth}
\centering{\includegraphics[width=1\linewidth]{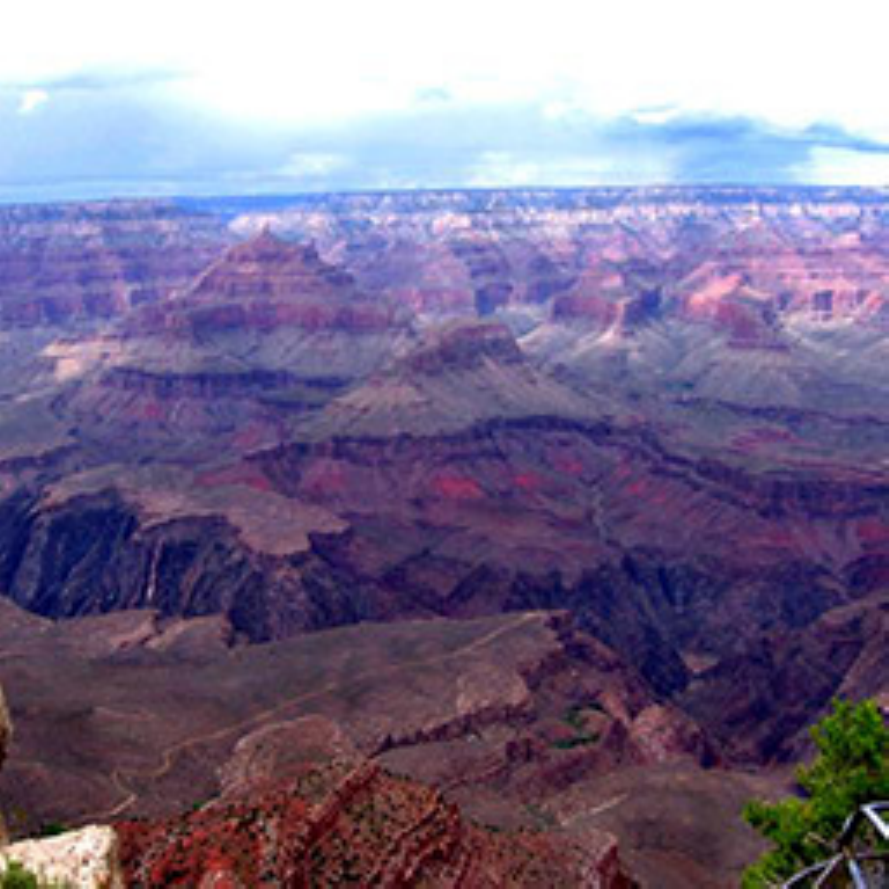}}
\end{minipage}
\hfill
\begin{minipage}{0.19\linewidth}
\centering{\includegraphics[width=1\linewidth]{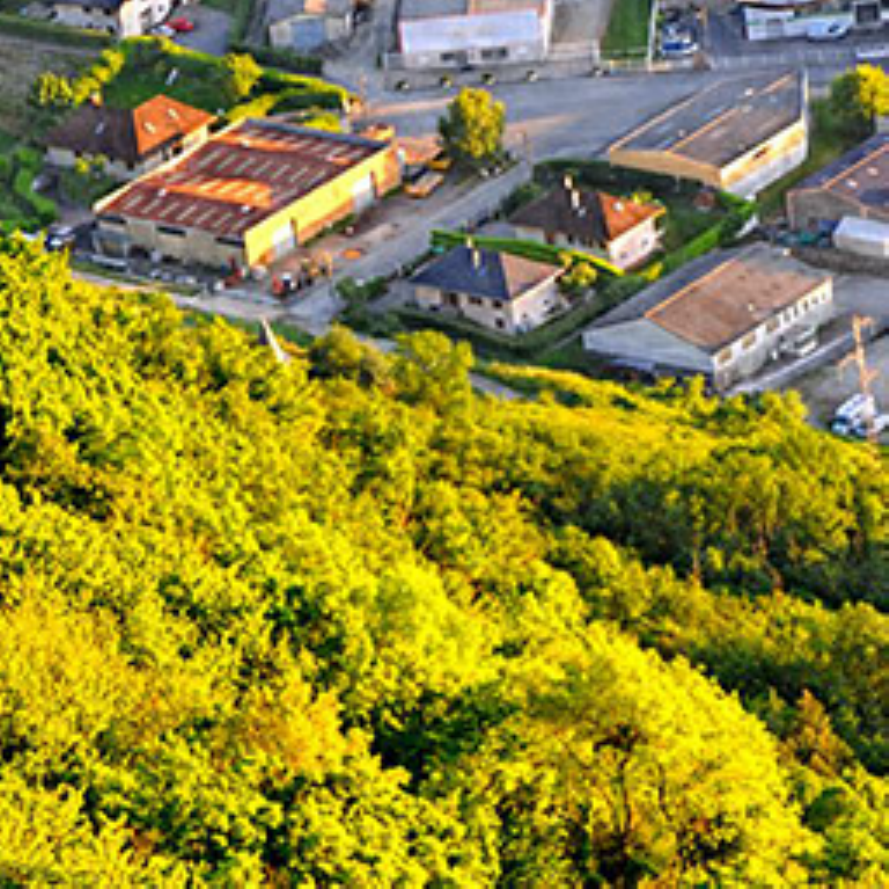}}
\end{minipage}
\hfill
\begin{minipage}{0.19\linewidth}
\centering{\includegraphics[width=1\linewidth]{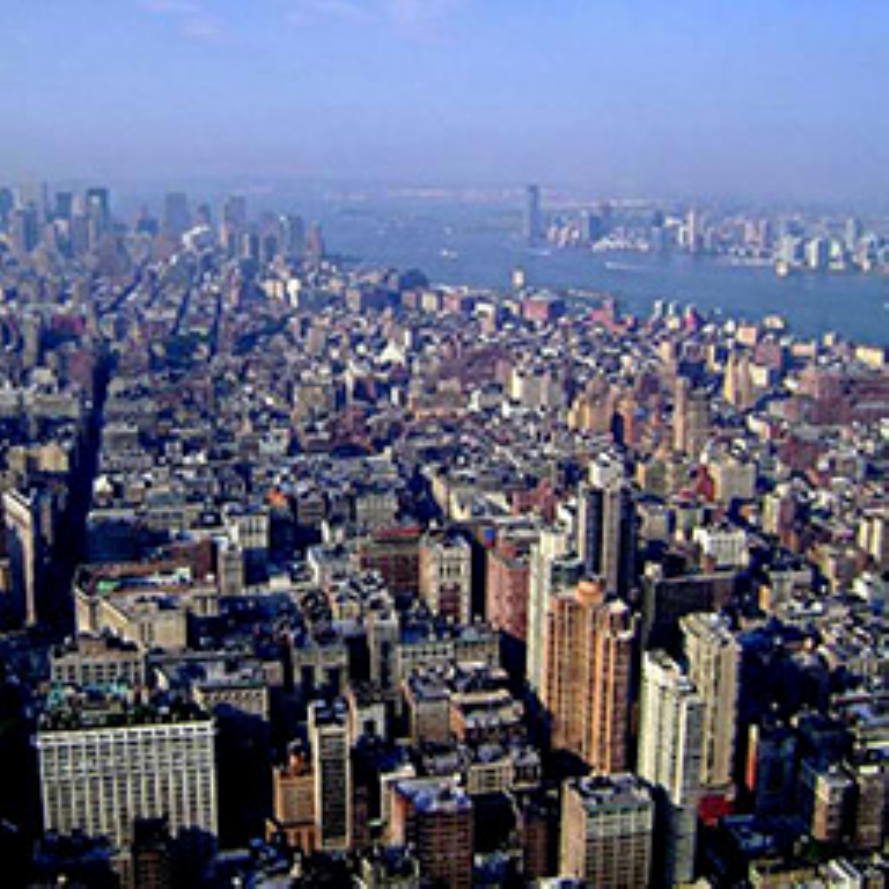}}
\end{minipage}
\hfill
\begin{minipage}{0.19\linewidth}
\centering{\includegraphics[width=1\linewidth]{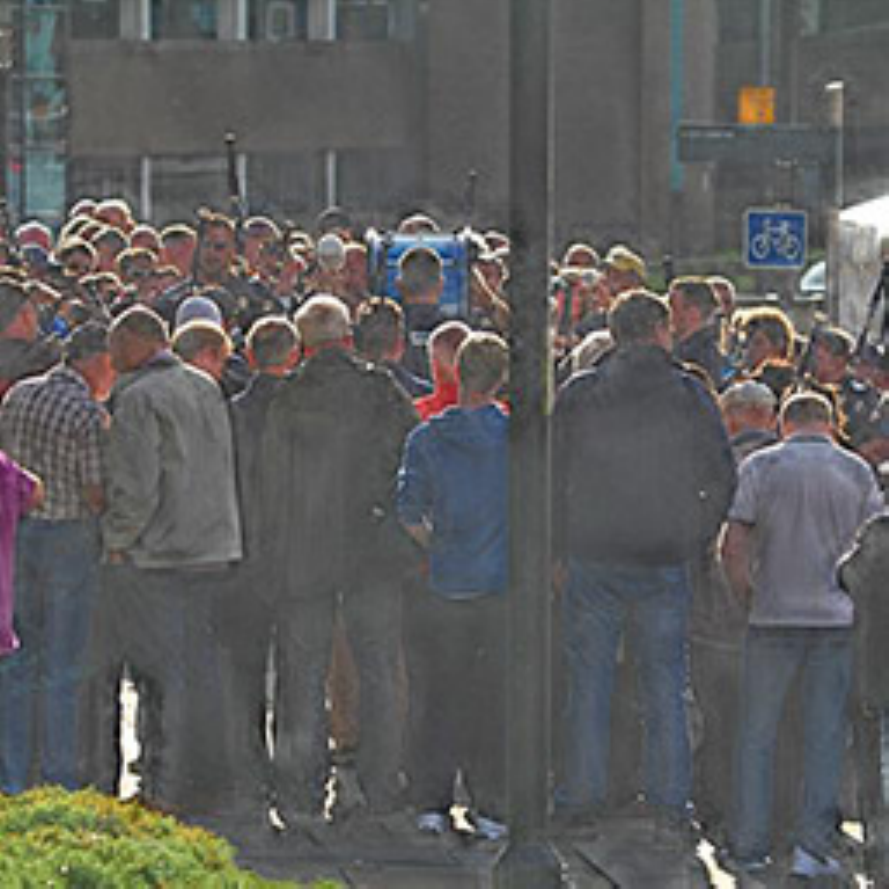}}
\end{minipage}
\hfill
\begin{minipage}{0.19\linewidth}
\centering{\includegraphics[width=1\linewidth]{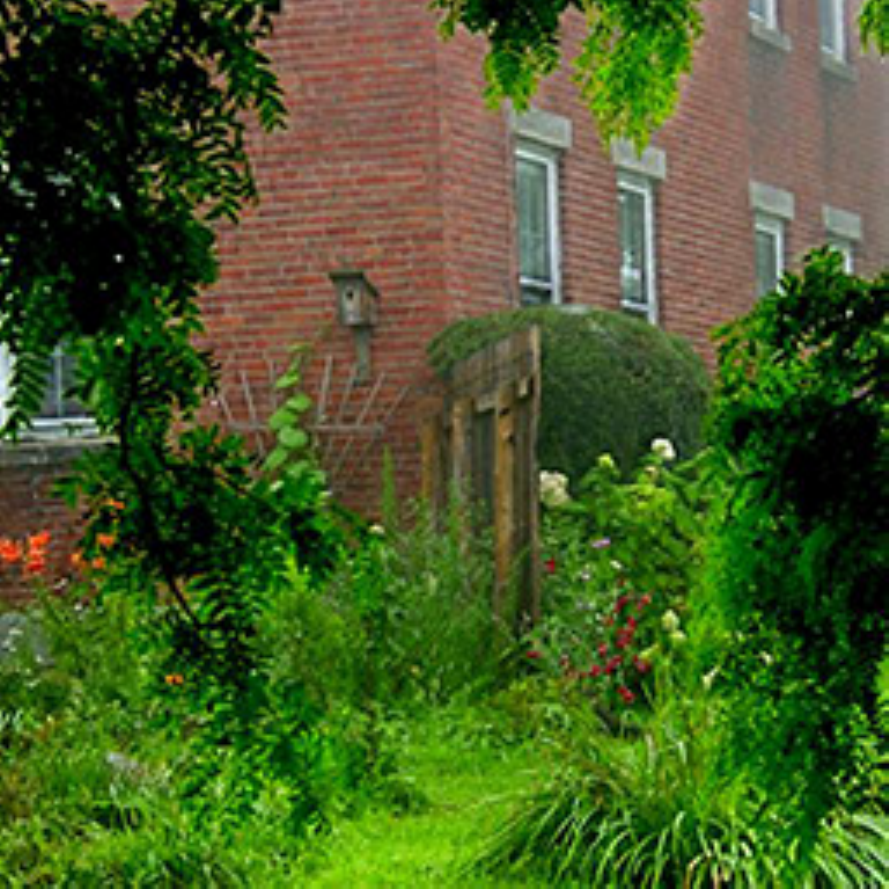}}
\end{minipage}
\end{center}
\caption{Dehazing results for some real-world hazy images: the first row is hazy images and the second row is the results.}
\label{fig:dehaze}
\end{figure}

\subsection{Ablation Studies}\label{sec:ablation}

To verify the roles of different parts in our AMPE-Net, we do some ablation experiments. Table \ref{tab:psnrssim_ablation} shows the  PSNR and SSIM of different variants of our network. Some visual results on synthetic and real-world images are shown in Figure \ref{fig:synthetic_ablation} and \ref{fig:parctical_ablation} respectively. In our ablation study, we do not include $\mathcal{R}(\cdot)$ whose role has been shown above. All the experiments in this subsection are conducted with $\alpha=1$. We can see that the guide of $\mathcal{H}(\cdot)$ (LocNet) is important, and it boosts performances. When removing  $\mathcal{F}(\cdot)$ (EstNet-T), our model degrades into $\mathbf{I}=\mathbf{B}+\mathbf{R}$, the PSNR/SSIM decrease the most seriously. Moreover, the performance of removing haze-like effect is also lower than other cases which proves the role of our rain model further (Figure \ref{fig:parctical_ablation}). The loss $\mathcal{L}_{2}$ also contributes to the performance improvement slightly.

\subsection{Potentials of our Model and Network}
Our model can be extended to other weather conditions, such as haze and snow. We conduct simple experiments to show the potential.
For haze, we randomly select $5000$ training samples from \cite{Li_2019_TIP} to dehaze with our model and networks (note that the LocNet will not be used in haze condition). We show some dehazing results for
real-world hazy images in Figure \ref{fig:dehaze}.

\section{Conclusion}
In this paper, we utilized a new model to describe rainy images more completely. To remove the rain effect more completely,
we proposed a two-branch network to learn the parameters in our rain model jointly.
Two invertible loss functions are utilized to optimize the two-branch unit alternatively to fit our model better. To control the strength for removing haze-like effect, an average weighted combination and an SPP structure were utilized to refine our rain-removed results. Besides, a location map of rain was also learned to guide the training of our network.
Compared with several state-of-the-art deep learning works,
our method outperforms these methods objectively and subjectively,
and our work can handle more kinds of rainy images, including removing haze-like effect to recover the original color of degraded images.


\input{egpaper_for_review_final.bbl}
\end{document}

%% file: dg_def.tex
\def\0{{\bf 0}}
\def\1{{\bf 1}}

\def\bB{{\bf B}}

\def\bI{{\bf I}}

\def\bR{{\bf R}}






%% file: egpaper_for_review_final.bbl
\begin{thebibliography}{10}\itemsep=-1pt

\bibitem{Fu_2017_TIP}
X.~Fu, J.~Huang, X.~Ding, Y.~Liao, and J.~Paisley.
\newblock Clearing the skies: a deep network architecture for single-image rain
  removal.
\newblock {\em IEEE Transactions on Image Processing}, 26(6):2944--2956, July
  2017.

\bibitem{Fu_2017_CVPR}
X.~Fu, J.~Huang, D.~Zeng, Y.~Huang, X.~Ding, and J.~Paisley.
\newblock Removing rain from single images via a deep detail network.
\newblock In {\em IEEE Conference on Computer Vision and Pattern Recognition
  (CVPR-2017)}, pages 1715--1723, Honolulu, HI, USA, July 2017. IEEE.

\bibitem{Garg_2004_CVPR}
K.~Garg and S.~K. Nayar.
\newblock Detection and removal of rain from videos.
\newblock In {\em IEEE Conference on Computer Vision and Pattern Recognition
  (CVPR 2004)}, volume~1, pages 528--535, Washington DC, USA, Jun. 2004. IEEE.

\bibitem{gong2017motion}
D.~Gong, J.~Yang, L.~Liu, Y.~Zhang, I.~Reid, C.~Shen, A.~van~den Hengel, and
  Q.~Shi.
\newblock From motion blur to motion flow: a deep learning solution for
  removing heterogeneous motion blur.
\newblock In {\em IEEE Conference on Computer Vision and Pattern Recognition
  (CVPR)}, 2017.

\bibitem{gong2018learning}
D.~Gong, Z.~Zhang, Q.~Shi, A.~v.~d. Hengel, C.~Shen, and Y.~Zhang.
\newblock Learning an optimizer for image deconvolution.
\newblock {\em arXiv preprint arXiv:1804.03368}, 2018.

\bibitem{He_2015_CVPR}
K.~He, X.~Zhang, S.~Ren, and J.~Sun.
\newblock Deep residual learning for image recognition.
\newblock In {\em IEEE Conference on Computer Vision and Pattern Recognition
  (CVPR-2017)}. IEEE, July 2015.

\bibitem{He_2015_arxiv}
K.~He, X.~Zhang, S.~Ren, and J.~Sun.
\newblock Spatial pyramid pooling in deep convolutional networks for visual
  recognition.
\newblock {\em arXiv:1406.4729}, 2015.

\bibitem{Huang_2017_CVPR}
G.~Huang, Z.~Liu, and L.~Maaten.
\newblock Densely connected convolutional networks.
\newblock In {\em IEEE Conference on Computer Vision and Pattern Recognition
  (CVPR-2017)}, pages 4700--4708, Honolulu, HI, USA, July 2017. IEEE.

\bibitem{Kang_2012_TIP}
L.~W. Kang, C.~W. Lin, and Y.~H. Fu.
\newblock Automatic single-image-based rain streaks removal via image
  decomposition.
\newblock {\em IEEE Transactions on Image Processing}, 21(4):1742--1755, Apr.
  2012.

\bibitem{Kingma_2015_ICLR}
D.~P. Kingma and J.~Ba.
\newblock Adam: A method for stochastic optimization.
\newblock In {\em the 3rd International Conference for Learning
  Representations(ICLR-2015)}, San Diego, 2015. IEEE.

\bibitem{ledig2017photo}
C.~Ledig, L.~Theis, F.~Husz{\'a}r, J.~Caballero, A.~Cunningham, A.~Acosta,
  A.~Aitken, A.~Tejani, J.~Totz, Z.~Wang, et~al.
\newblock Photo-realistic single image super-resolution using a generative
  adversarial network.
\newblock In {\em Proceedings of the IEEE conference on computer vision and
  pattern recognition}, pages 4681--4690, 2017.

\bibitem{Li_2019_TIP}
B.~Li, W.~Ren, D.~Fu, D.~Tao, D.~Feng, W.~Zeng, and Z.~Wang.
\newblock Benchmarking single-image dehazing and beyond.
\newblock {\em IEEE Transactions on Image Processing}, 28(1):492--505, 2019.

\bibitem{Li_2018_MM}
G.~Li, X.~He, W.~Zhang, H.~Chang, L.~Dong, and L.~Lin.
\newblock Non-locally enhanced encoder-decoder network for single image
  de-raining.
\newblock In {\em ACM Multimedia (MM-2018)}, Seoul, Republic of Korea, Oct.
  2018. ACM.

\bibitem{Li_rt_2019_CVPR}
R.~Li, F.~Cheong, F.~Cheong, and T.~Tan.
\newblock Heavy rain image restoration: integration physics model and
  conditional adversarial learning.
\newblock In {\em IEEE Conference on Computer Vision and Pattern Recognition
  (CVPR 2019)}, Long Beach CA, USA, Jun. 2019. IEEE.

\bibitem{Li_2017_arxiv}
R.~Li, L.~F. Cheong, and R.~T. Tan.
\newblock Single image deraing using scale-aware multi-stage recurrent network.
\newblock {\em arXiv:1712.06830}, 2017.

\bibitem{Li_sy_2019_CVPR}
S.~Li, I.~Araujo, W.~Ren, Z.~Wang, and E.~Tokuda.
\newblock Single image deraining: a comprehensive benchmark analysis.
\newblock In {\em IEEE Conference on Computer Vision and Pattern Recognition
  (CVPR 2019)}, Long Beach CA, USA, Jun. 2019. IEEE.

\bibitem{Li_2018_arxiv}
S.~Li, W.~Ren, J.~Zhang, J.~Yu, and X.~Guo.
\newblock Fast single image rain removal via a deep decomposition-composition
  network.
\newblock {\em arXiv:1804.02688}, 2018.

\bibitem{Li_2018_ECCV}
X.~Li, J.~Wu, Z.~Lin, H.~Liu, and H.~Zha.
\newblock Recurrent squeeze-and-excitation context aggregation net for single
  image deraining.
\newblock In {\em European Conference on Computer Vision (ECCV-2018)}, Munich,
  Germany, Sep. 2018. IEEE.

\bibitem{Mairal_2010_JMLR}
J.~Mairal, F.~Bach, J.~Ponce, and G.~Sapiro.
\newblock Online learning for matrix factorization and sparse coding.
\newblock {\em Journal of Machine Learning Research}, pages 19--60, Mar. 2010.

\bibitem{Narasimhan_2002_IJCV}
S.~G. Narasimhan and S.~K. Nayar.
\newblock Vision and the atmosphere.
\newblock {\em International journal of computer vision}, (3):233--254, Dec.
  2002.

\bibitem{Pan_2018_arxiv}
J.~Pan, Y.~Liu, J.~Dong, J.~Zhang, J.~Ren, J.~Tang, Y.~Tai, and M.~Yang.
\newblock Physics-based generative adversarial models for image restoration and
  beyond.
\newblock {\em arXiv:1808.00605}, 2018.

\bibitem{Sifre_2014_Phd_thesis}
L.~Sifre.
\newblock Rigid-motion scattering for image classification.
\newblock {\em Ph.D thesis}, 2014.

\bibitem{Wang_ty_2019_CVPR}
T.~Wang, X.~Yang, K.~Xu, S.~Chen, Q.~Zhang, and R.~Lau.
\newblock Spatial attentive single-image deraining with a high quality real
  rain dataset.
\newblock In {\em IEEE Conference on Computer Vision and Pattern Recognition
  (CVPR 2019)}, Long Beach CA, USA, Jun. 2019. IEEE.

\bibitem{Wang_2017_TIP}
Y.~Wang, S.~Liu, C.~Chen, and B.~Zeng.
\newblock A hierarchical approach for rain or snow removing in a single color
  image.
\newblock {\em IEEE Transactions on Image Processing}, 26(8):3936--3950, August
  2017.

\bibitem{Wang_2016_ICIP}
Y.~L. Wang, C.~Chen, S.~Y. Zhu, and B.~Zeng.
\newblock A framework of single-image deraining method based on analysis of
  rain characteristics.
\newblock In {\em IEEE International Conference on Image Processing (ICIP
  2013)}, pages 4087 -- 4091, Phoenix, USA, Sep. 2016. IEEE.

\bibitem{Wang_2004_TIP}
Z.~Wang, A.~C. Bovik, H.~R. Sheikh, and E.~P. Simoncelli.
\newblock Image quality assessment: from error visibility to structural
  similarity.
\newblock {\em IEEE Transactions on Image Processing}, 13(4):600--612, April
  2004.

\bibitem{yan2019attention}
Q.~Yan, D.~Gong, Q.~Shi, A.~v.~d. Hengel, C.~Shen, I.~Reid, and Y.~Zhang.
\newblock Attention-guided network for ghost-free high dynamic range imaging.
\newblock {\em arXiv preprint arXiv:1904.10293}, 2019.

\bibitem{yang2018seeing}
J.~Yang, D.~Gong, L.~Liu, and Q.~Shi.
\newblock Seeing deeply and bidirectionally: A deep learning approach for
  single image reflection removal.
\newblock In {\em Proceedings of the European Conference on Computer Vision
  (ECCV)}, pages 654--669, 2018.

\bibitem{Yang_2017_CVPR}
W.~Yang, R.~Tan, J.~Feng, J.~Liu, Z.~Guo, and S.~Yan.
\newblock Deep joint rain detection and removal from a single image.
\newblock In {\em IEEE Conference on Computer Vision and Pattern Recognition
  (CVPR-2017)}, pages 1685--1694, Honolulu, HI, USA, July 2017. IEEE.

\bibitem{Zhang_2018_CVPR}
H.~Zhang and V.~Patel.
\newblock Density-aware single image de-raining using a multi-stream dense
  network.
\newblock In {\em IEEE Conference on Computer Vision and Pattern Recognition
  (CVPR-2018)}, pages 1685--1694, Salt Lake City, UT, July 2018. IEEE.

\end{thebibliography}
